\documentclass[twoside]{article}
\usepackage[utf8]{inputenc}
\usepackage{amsmath}
\usepackage{amsfonts}
\usepackage{amssymb}
\usepackage{graphicx}
\usepackage{color}
\usepackage[normalem]{ulem}
\usepackage{bbm, relsize}
\usepackage[left=2cm,right=2cm,top=2cm,bottom=2cm]{geometry}
\usepackage{bbm}

\usepackage{algorithm}
\usepackage{algpseudocode}
\usepackage{hyperref}

\def\hindu{\arabic}
\newenvironment{Proof}{\noindent{\sc Proof.}}{\qed}
\newtheorem{theorem}{Theorem}[section]
\newtheorem{lemma}{Lemma}[section]
\newtheorem{cor}{Corollary}[section]
\newtheorem{rem}{Remark}[section]
\newtheorem{definition}{Definition}[section]
\newtheorem{prop}{Proposition}[section]
\newtheorem{uda}{Example}[section]
\newcommand{\qed}{\hfill$\Box$\par\medskip}
\renewcommand{\theequation}{\hindu{section}.\hindu{equation}}
\def\bhag#1{\noindent
\setcounter{equation}{0}
\section{#1}
}

\def\RR{{\mathbb R}}

\def\ZZ{{\mathbb Z}}

\def\PPI{{{\rm I}\kern-1pt\Pi}}

\def\S{\mathbf{S}}
\def\TT{\mathbb T}

\def\b #1;{{\bf #1}}
\def\x{{\bf x}}
\def\k{{\bf k}}
\def\y{{\bf y}}

\def\z{{\bf z}}

\def\m{\mathfrak{m}}
\def\j{\mathbf{j}}

\def\O{{\cal O}}

\def\C{{\mathcal C}}

\def\A{{\cal A}}

\def\argmax{\mathop{\hbox{{\rm arg max}}}}

\def\be{\begin{equation}}
\def\ee{\end{equation}}
\def\bea{\begin{eqnarray}}
\def\eea{\end{eqnarray}}
\def\eref#1{(\ref{#1})}
\def\disp{\displaystyle}

\def\donchitre#1#2{\vskip 6.5cm\noindent
\parbox[t]{1in}{\special{eps:#1.eps x=6.5cm y=5.5cm}}
\hbox to 7cm{}\parbox[t]{0.0cm}{\special{eps:#2.eps x=6.5cm y=5.5cm}}}

\def\XX{{\mathbb X}}
\def\BB{{\mathbb B}}

\def\bs#1{{\boldsymbol{#1}}}

\def\dist{\mathsf{dist }}
\def\supp{\mathsf{supp\ }}

\title{Cautious Active Clustering }
\author{
 A.~Cloninger\thanks{Department of Mathematics and Halicioglu Data Science Institute, University of California San Diego, San Diego, CA 92093, U.S.A..  The research of this author is supported in part by the NSF DMS grants 2012266,  1819222, and Sage Foundation Grant 2196.
 \textsf{email:} acloninger@ucsd.edu},\ \  H.~N.~Mhaskar\thanks{
Institute of Mathematical Sciences, Claremont Graduate University, Claremont, CA 91711, U.S.A.. 
The research of this author is supported in part
NSF DMS grant 2012355.
\textsf{email:} hrushikesh.mhaskar@cgu.edu} 
  }
\date{}

\begin{document}
\maketitle
\begin{abstract}
We consider the problem of classification of points sampled from an unknown probability measure on a Euclidean space.
 We study the question of querying the class label at a very small number of judiciously chosen points so as to be able to attach the appropriate class  label to every point in the set.
 Our approach is to consider the unknown probability measure as a convex combination of the conditional probabilities for each class. 
Our technique involves the use of a highly localized kernel constructed from Hermite polynomials, in order to create a hierarchical estimate of the supports of the constituent probability measures. 
We do not need to make any assumptions on the nature of  any of the probability measures nor know in advance the number of classes involved.
We give theoretical guarantees measured by the $F$-score for our classification scheme.
Examples include classification in hyper-spectral images  and MNIST classification.
\end{abstract}

\bhag{Introduction}\label{intsect}

The main purpose of this paper is to study a classification problem in the theory of machine learning, which we have called cautious active learning, by modifying certain ideas originating in our previous work on blind source signal separation. 
In Section~\ref{bhag:machine}, we describe the classification problem in the theory of machine learning which we are interested in.
In Section~\ref{bhag:super}, we describe briefly our work on blind source signal separation that motivates our current paper, and provides a prototype for the results in this paper. 
Section~\ref{bhag:meas_sep} explains the difficulties involved in  adapting the approach in Section~\ref{bhag:super} and gives a preview of the kind of  results expected with our solution presented in this paper.
In Section~\ref{bhag:priorwork}, we discuss connections with a few other works related to the problem and our solution to the same.
The outline of the  paper is given in Section~\ref{bhag:outline}.
For the convenience of exposition, the notation used in this section is not the same as the one used in the rest of this paper after Section~\ref{bhag:notation}.

\subsection{Cautious active learning}\label{bhag:machine}
An important task of machine learning is to classify various objects into a finite number of classes.
Typically, this task is formulated as follows (e.g, \cite{chapelle2009semi, bartlett}). 
We are given data of the form $\{(\x_i,y_i)\}_{i=1}^M$ where $\x_i$'s are in some Euclidean space $\RR^q$, and $y_i\in \{1,\cdots,K\}$ for some integer $K\ge 1$. 
In supervised learning, we have to a build a model $P$ such that for any vector $\x\in\RR^q$ (or a compact subset thereof), $P(\x)$ gives reliably the class to which $\x$ belongs.
In semi-supervised learning, the labels $y_i$ are known only for a small number of $\x_i$'s, and the problem is to extend this labeling to the rest of the data set. 
It is assumed that the data set is known in advance; it is not expected to build a model for points not in the original data set.
In unsupervised learning, no information is known about the labels, and the best that can be done is to find the right clusters in the dataset.

Active learning is a relatively recent area of machine learning that combines aspects of all of three paradigms above. 
We do not know any labels to begin with, but are allowed to seek labels on judiciously chosen points $\x_i$, as few as needed to construct a model $P$ as in the case of supervised learning. 
Clearly, this must be done in the beginning using clustering as in unsupervised learning, based on some model.
We then ``purify'' this clustering using a small number of queries for the label. 
In the end, we have started in the unsupervised regime, and then collected a small number of labeled data as in the semi-supervised regime, and finally built a model as in the supervised regime.
However, in semi-supervised learning, we cannot control the set of points at which the label is known, and we do not expect a model for the points not in the original data set. 
In contrast, in active learning, we get to choose which points to query the label at, and a model is expected as the end-product.

It is customary to assume that the data is drawn from an unknown probability distribution. Obviously,
\be\label{intuitive_decomposition}
\mathsf{Prob}(\x,k)=\mathsf{Prob}(k|\x)\mathsf{Prob}(\x)=\mathsf{Prob}(\x|k)\mathsf{Prob}(k).
\ee 
The first equation leads us to discriminative models. 
The class $k$ of a given point $\x$ is 
$$
\argmax_{k=1,\cdots,K}\mathsf{Prob}(k|\x)\mathsf{Prob}(\x).
$$
In \cite{witnesspaper}, we have explored this approach in further detail, giving theoretically well founded criteria to determine how to estimate the class reliably.

In this paper, we take a closer look at the second equation in \eref{intuitive_decomposition}; i.e., use the fact that
\be\label{marginal_as_sum}
\mathsf{Prob}(\x)=\sum_{k=1}^K \mathsf{Prob}(\x|k)\mathsf{Prob}(k).
\ee
In measure theoretic notation, one can write
\be\label{prob_sum}
\mu^*=\sum_{k=1}^K \mu_k,
\ee
where $\mu^*$ is the marginal distribution from which the points $\x$ are chosen, and $\mu_k$ represents the $k$-th term in \eref{marginal_as_sum}; i.e., some positive measure. (It is understood that $\mu^*$ is a probability measure, and it is not claimed that each $\mu_k$ is a probability measure; indeed, it represents $k$-th summand in  \eref{marginal_as_sum}, which accounts for the proportion of samples in class $k$.)
The task is to separate the supports of the unknown component measures $\mu_k$ given random samples taken from the unknown probability distribution $\mu^*$.
The intuition is that once the supports of each $\mu_k$ is known, we just need one sample from each to complete the task of classification using only the smallest number of samples.

Because of overlapping class boundaries, it is not reasonable to assume that the classes; i.e., the supports of the constituent measures, are well separated. 
In this paper, we propose a hierarchical classification scheme, where  the minimal separation among the supports of $\mu_k$'s is decreased step by step. 
The accuracy of our hierarchical clustering schemes is proved using the classical $F$-score as the measurement of quality of clustering.  We focus on the transductive learning case, in which we are generalizing from the small labeled examples to the specific unlabeled data that is available \cite{yu2006active}.  

We note that in the absence of minimal separation among the supports of the measures $\mu_k$, the decomposition \eqref{prob_sum} is not uniquely defined; e.g., we may group the $K$ measures in many different ways, resulting in $<K$ components. 
In an unsupervised setting, this is only natural. 
For example, a data base of images can be classified at different levels as that of an animate or non-animate object, as that of a human, or animal, or movable or unmovable object, etc.; ultimately viewing each image as its own class.
Accordingly, our definitions and theorems will not assume a prior knowledge of the number of classes (equivalently, the measures $\mu_k$). 
The role of active learning is to reconcile this with a given classification problem with increasing confidence.
Thus, having separated the clusters at different levels of minimal separation, we seek labels from each of them, and combine or further subdivide them so as to achieve the known number of classes. 

\subsection{Motivation for our approach}\label{bhag:super}

 Let $q\ge 1$ be an integer, $\TT^q=\RR^q/(2\pi\ZZ^q)$. 
For $\x,\y\in \TT^q$, we define (in this section only) $|\x-\y|=\max_{1\le k\le q}|(x_k-y_k) \mbox{ mod } 2\pi|$.
One formulation of the problem of blind source signal separation is the following.
Let $\mu^*=\sum_{k=1}^K a_k\delta_{\x_k}$ be a (signed) measure supported at points $\x_k\in\TT^q$, where $\delta_\x$ is the Dirac delta measure supported at $\x$. 
, The goal is to recuperate the number $K$ of components, the point sources $\x_k$ and the (signed, complex) amplitudes $a_k$, given the Fourier moments  $\widehat{\mu^*}(\j)=\sum_{k=1}^K a_k\exp(-i\j\cdot \x_k)$ for $|\j|_\infty <N$ for some $N$. 
Our solution to this problem described in  \cite{bspaper, trigwave, loctrigwave} is the following.
We consider a filter $H :\RR\to [0,1]$ that is an infinitely differentiable,  even function, with $H(t)=0$ for $|t|\ge 1$.
For integer $n\ge 1$, we then consider an operator
$$
\mathcal{T}_n(f)(\x)=\hbar_n\sum_{\j\in\ZZ^q} H\left(\frac{|\j|}{n}\right)f(\j)\exp(i\j\cdot\x), \qquad \x\in\TT^q,
$$
where 
$$\hbar_n=\left(\sum_{\j\in\ZZ^q} H\left(\frac{|\j|}{n}\right)\right)^{-1}.
$$
With
$$
\Phi_n^T(\x-\y)=\sum_{\j\in\ZZ^q} H\left(\frac{|\j|}{n}\right)\exp(i\j\cdot(\x-\y)),
$$
it is not difficult to verify that
\be\label{sso_integral}
\mathcal{T}_n(f)(\x)=\frac{\hbar_n}{(2\pi)^q}\int_{\TT^q}\Phi_n^T(\x-\y)d\mu^*(\y).
\ee
The following theorem from \cite{bspaper} serves as a precursor of our research described in this paper, where we use the notation $\eta$ to denote the minimal separation among the points $\x_k$ (i.e., $\eta=\min_{1\le k<j\le K}|\x_k-\x_j|$), and $\mathfrak{m}$ to denote the minimum of the $|a_k|$'s.
Applications of theorems of this sort to direction finding in phased array antennas is discussed in \cite{loctrigwave}.
\begin{theorem}\label{theo:bstheo}
For sufficiently large $n$ (depending upon $\eta$), the set of $\x\in\TT^q$ at which $|\mathcal{T}_n(f)(\x)|\ge\mathfrak{m}/2$
 is a disjoint union of \textbf{exactly} $K$ sets $\mathcal{G}_k$, $1\le k\le K$, each containing exactly one point $\x_k$, and each with diameter $\le c/n$ for some positive constant $c$ with each of the following properties. (i) The minimal separation among the sets $\mathcal{G}_k$ is at least $c/n$, (ii) 
 If $\widehat{\x_k}$ is the highest peak of the power spectrum $|\mathcal{T}_n(f)|$ in $\mathcal{G}_k$, then (clearly) $|\widehat{\x_k}-\x_k|\le c/n$, and (iii) 
$
\left|\mathcal{T}_n(f)(\widehat{\x_k})-a_k\right| \le c_1/n.
$
\end{theorem} 
The key ingredient in the proof of Theorem~\ref{theo:bstheo} is the localization estimate
\be\label{trigkernloc}
|\Phi_n^T(\x-\y)|\le \frac{c(H,S)}{\max(1, (n|\x-\y|)^S)}, \qquad \x,\y\in\TT^q,
\ee
where $c(H,S)>0$ is a constant independent of $n,\x,\y$.
We note that $n$ is the degree (order) of the trigonometric polynomial $\Phi_n^T$. 
In contrast to the kernel estimators in statistics, the localization here is achieved by letting $n\to\infty$.
\subsection{Separation of measures}\label{bhag:meas_sep}
The basic idea in our paper is to use an analogous localized kernel $\Phi_n$ to be defined in \eref{summkerndef} below (cf. \cite{mohapatrapap, hermite_recovery}) based on Hermite polynomials. 
It is not difficult to verify using known results about these kernels that $\int_{\RR^q}\Phi_n(\x,\y)d\mu^*(\y)\to d\mu^*(\x)$ in a weak-star sense, and the rate of approximation is optimal in the case when $\mu^*$ is absolutely continuous with respect to the Lebesgue measure on $\RR^q$ with a smooth density.
Therefore, it is reasonable to expect that 
$$
\int_{\RR^q}\Phi_n(\x,\y)d\mu^*(\y)=\sum_{k=1}^K \int_{\RR^q}\Phi_n(\x,\y)d\mu_k(\y)
$$
will split into clusters of $\x$ belonging to the supports of the measures $\mu_k$.

This optimality of approximation of measures however \emph{requires}  that the kernel $\Phi_n$ is  not a positive kernel. 
When $\mu^*$ is discretely supported, then the localization properties of the kernel ensure that near any one point of the support, the contribution to the integral from other points is negligible. 
This is not the case when the measure is supported on a continuum. 
Therefore, the problem of finding the support of $\mu^*$ is different from the problem of finding $\mu^*$ itself.
In our paper, we are interested only finding the supports, not the measures themselves. 
So, we will use the kernel $\Phi_n^2$ instead.

Apart from this technicality, there are many inherent barriers which makes the problem in our setting similar, yet very different, from the problem of super-resolution as described.
We illustrate with two examples. 

\begin{uda}\label{uda:balllineseparation}
{\rm
We consider a mixture of two distributions, $2/3$-rd part a uniform distribution on a 2D ball, and $1/3$-rd part a uniform distribution on a 1D line, with minimal separation $\delta\in\{0,0.1,0.2\}$ between the distributions.
In Figure~\ref{fig:balloon}, we show the results of our method based on a total of $1000$ points, with the value of the parameter $n=7$.
Our method estimates the relative supports of the two distributions, and is able to maintain the separation between the two distributions even for small minimal separation, and at low degree given by $n^2$.  This is of note because there is no assumption on the dimension of the support of the distributions, and similarly no assumption on the nature of the constituent distributions.

We compare our support detection algorithm to a standard gaussian kernel density estimate with the same kernel bandwidth in Figure \ref{fig:balloon}.  As a comparison, we display an indicator function of whether the density estimate is greater than $\frac{1}{4}$ the maximum peak of the density estimate,
\be
\frac{1}{M}\sum_{i=1}^M \Phi_n^2(x,x_i)>0.25\cdot \max_x\left(\frac{1}{M}\sum_{i=1}^M \Phi_n^2(x,x_i)\right).
\ee
This is a proxy for the estimate of the support of the clusters.
It is clear from the figures that our $\Phi_n^2$ kernel defines a sharper boundary for the indicator function, detecting all points regions within the support of the density while still maintaining a separation between the clusters.  Similarly, the support estimate from our kernel provides a tighter estimate normal to the 1D line than a Guassian KDE, and captures a better estimate of the minimal separation. \qed

\begin{figure}[ht]
\centering
\begin{tabular}{ccccc}
\includegraphics[height=.12\textwidth]{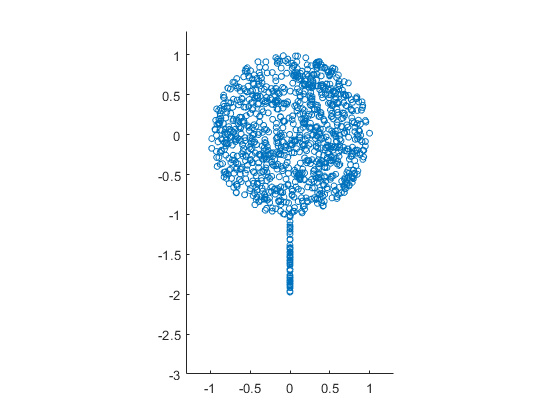} & 
\includegraphics[height=.12\textwidth]{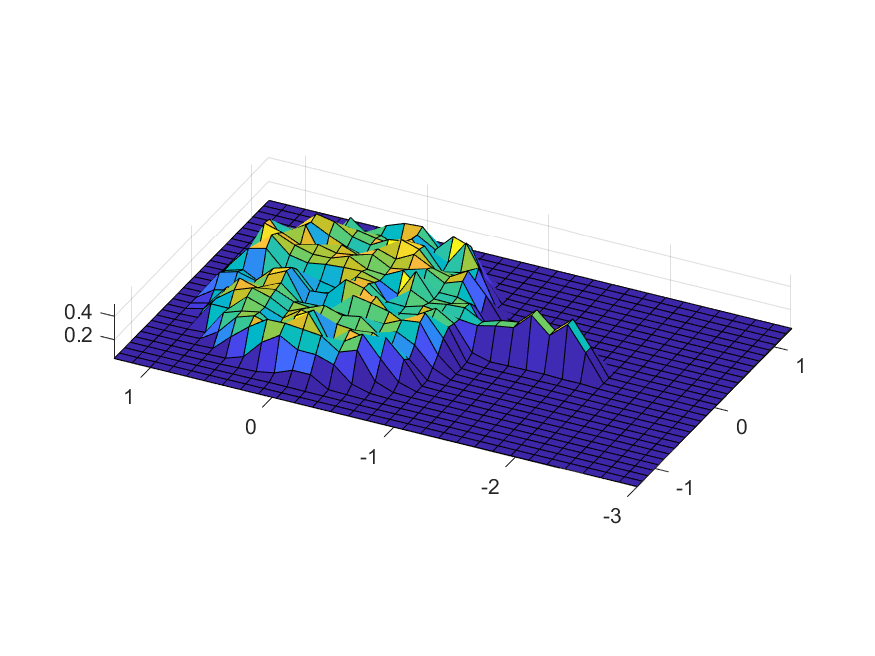} &
\includegraphics[height=.12\textwidth]{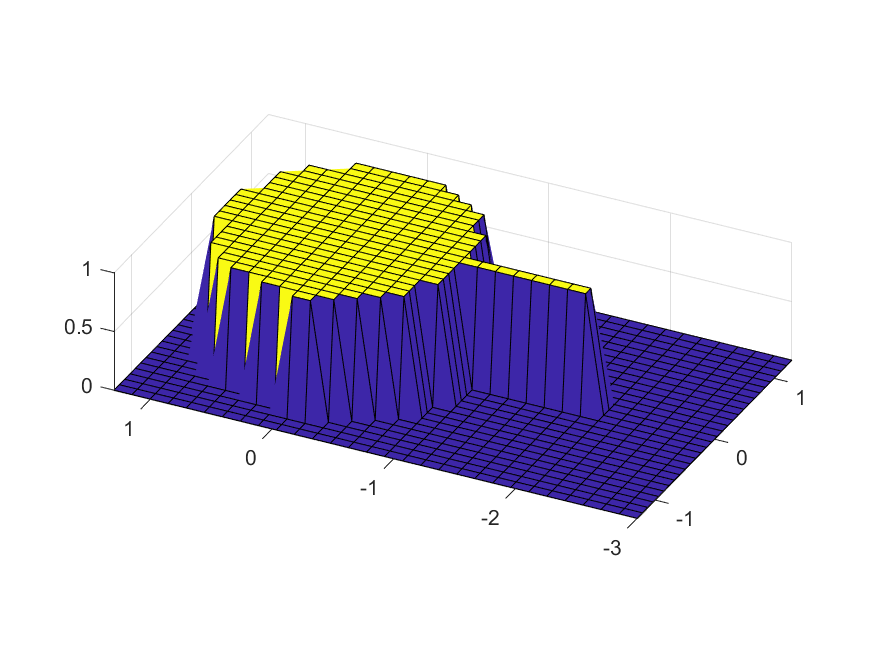} &
\includegraphics[height=.12\textwidth]{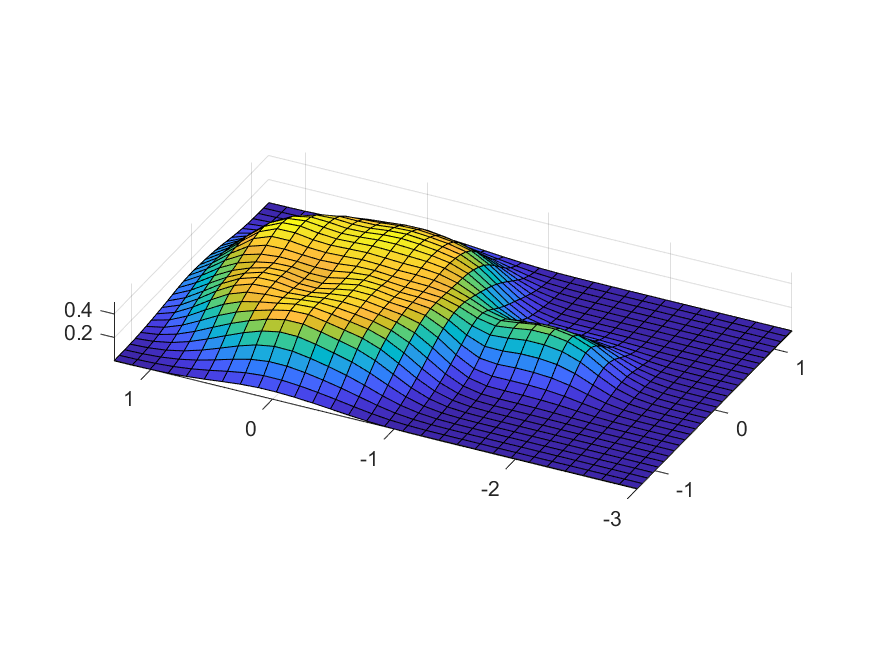} &
\includegraphics[height=.12\textwidth]{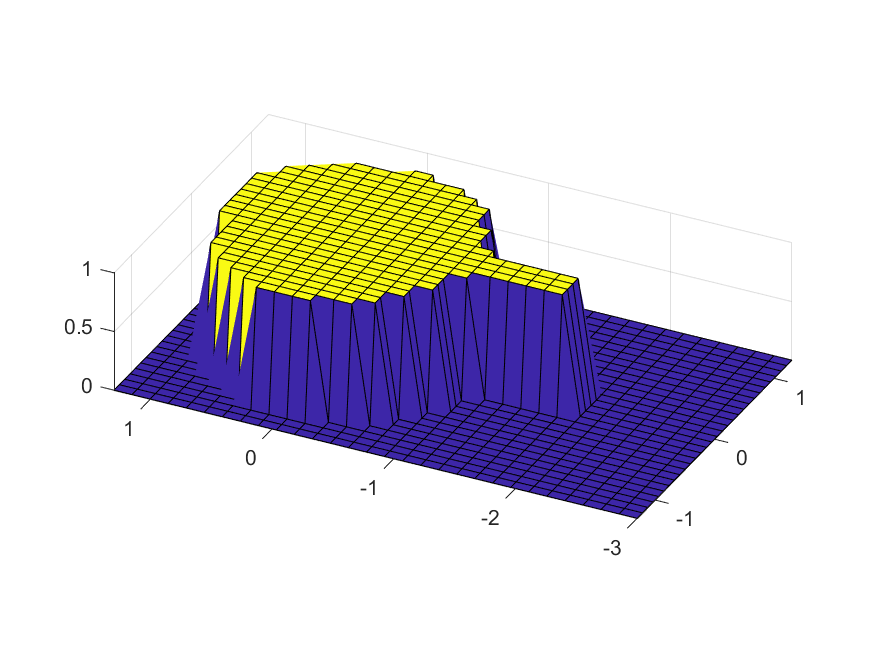} \\
\includegraphics[height=.12\textwidth]{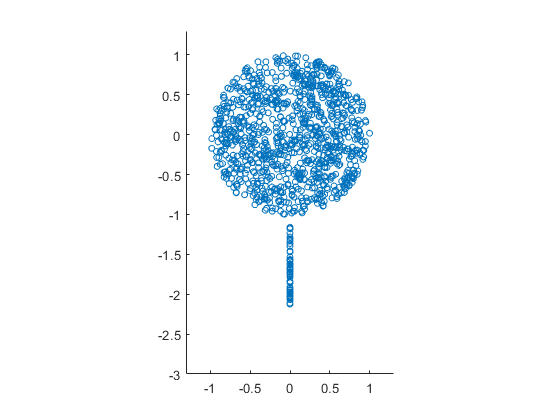} & 
\includegraphics[height=.12\textwidth]{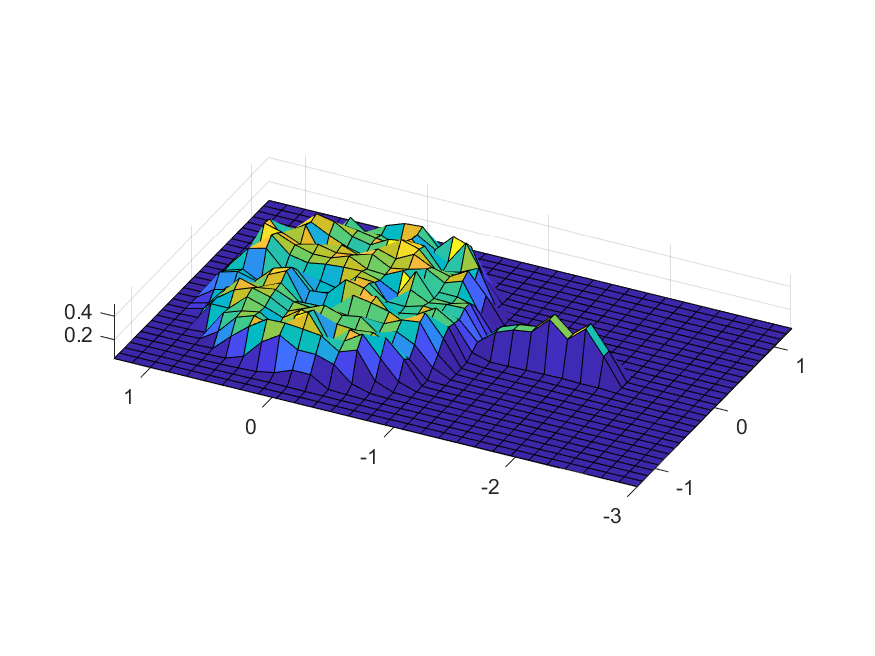} &
\includegraphics[height=.12\textwidth]{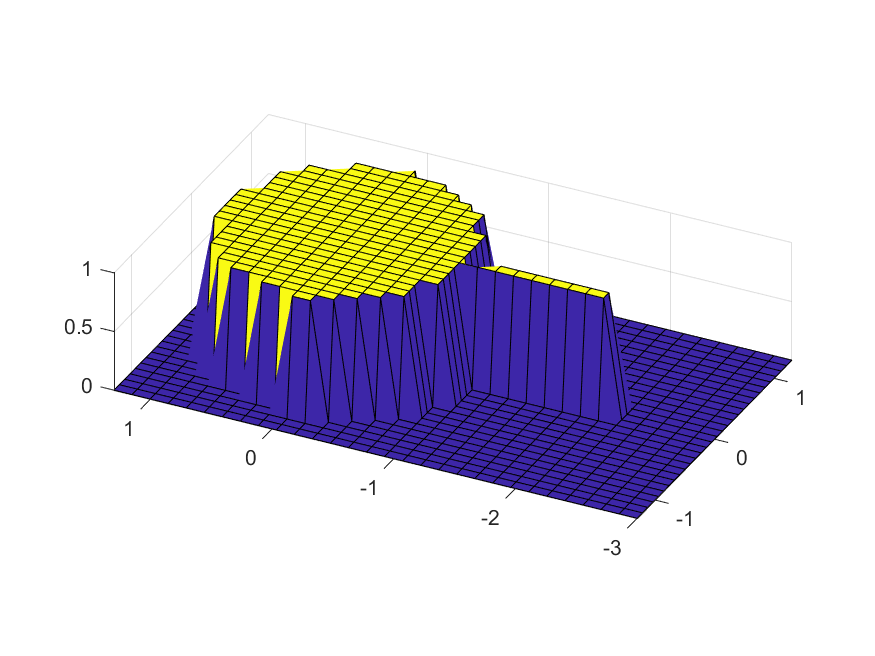} &
\includegraphics[height=.12\textwidth]{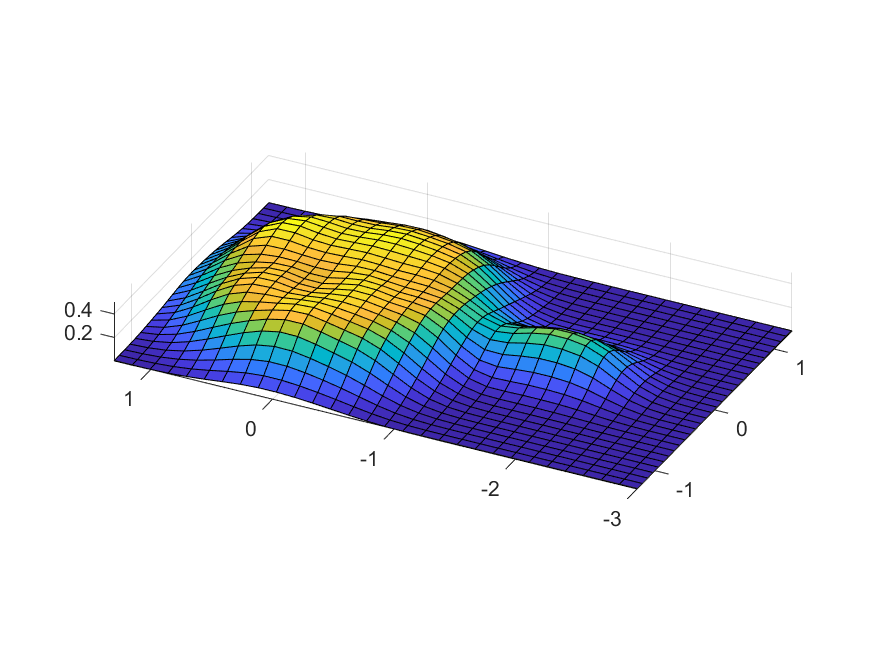} &
\includegraphics[height=.12\textwidth]{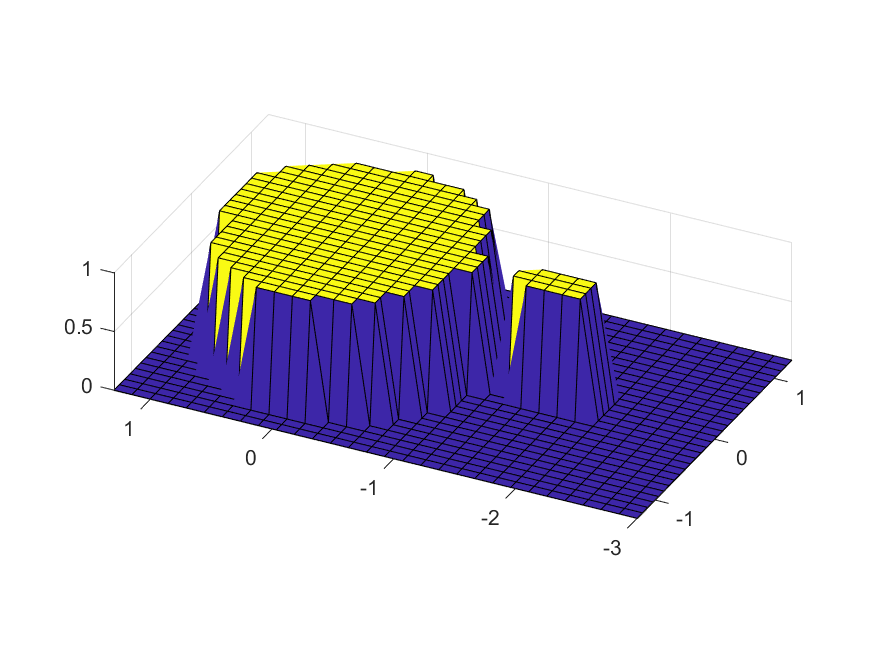} \\
\includegraphics[height=.12\textwidth]{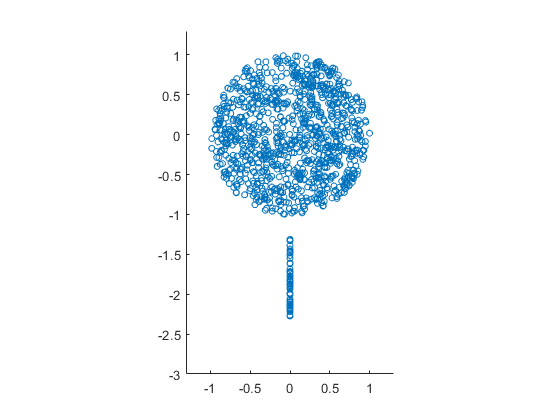} & 
\includegraphics[height=.12\textwidth]{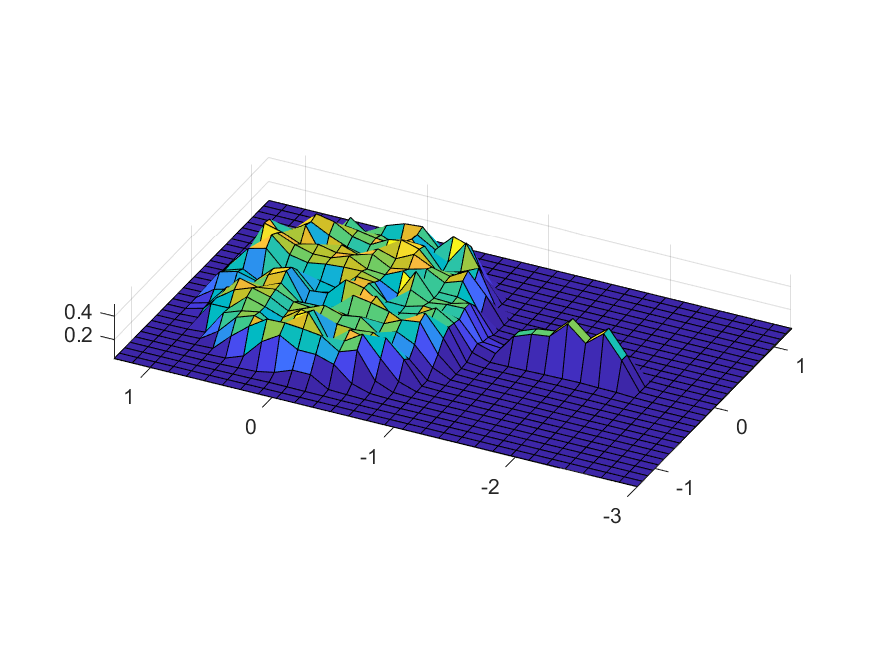} &
\includegraphics[height=.12\textwidth]{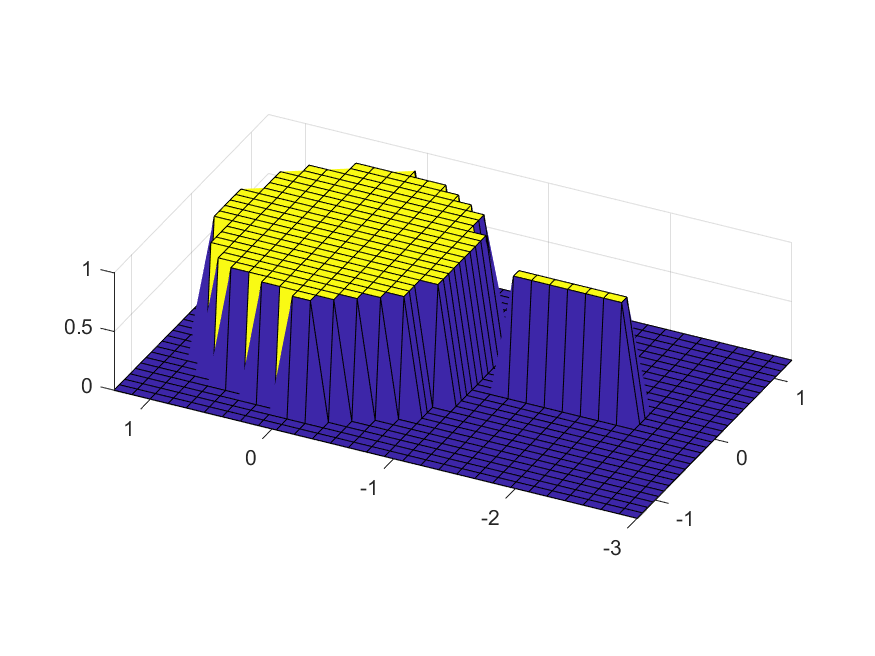} &
\includegraphics[height=.12\textwidth]{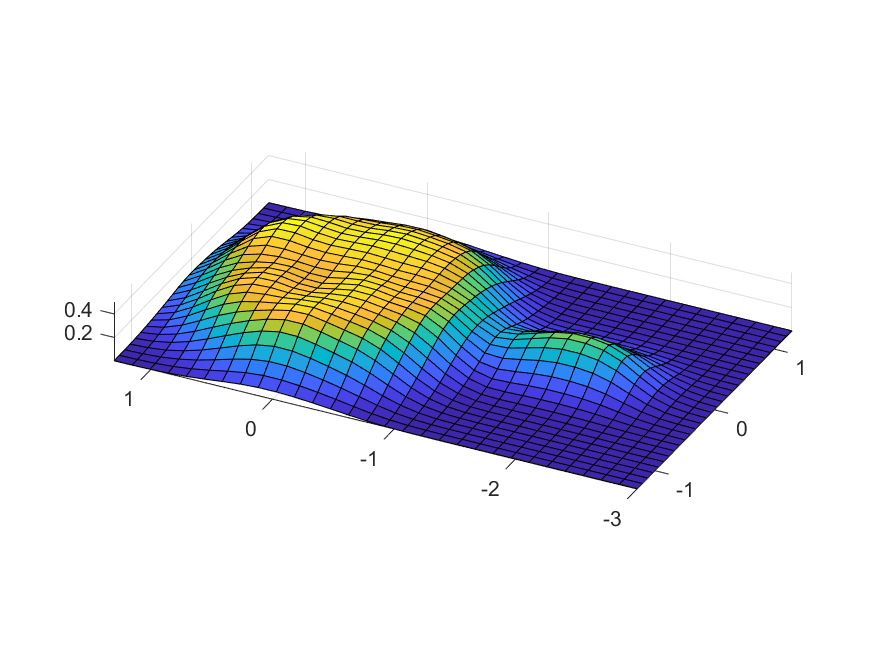} &
\includegraphics[height=.12\textwidth]{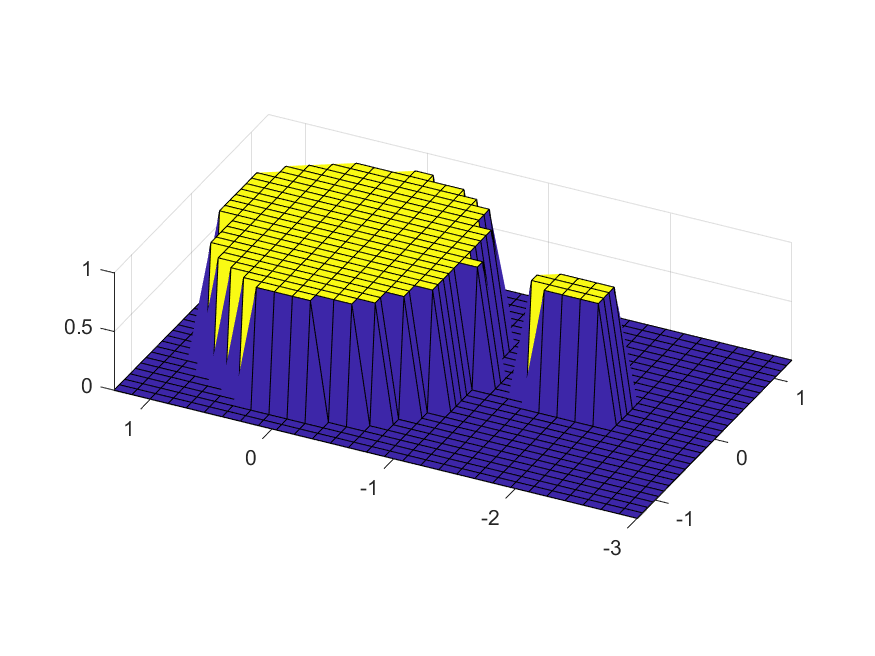} 
\end{tabular}
\caption{(Column 1) Data points, (Column 2) $\Phi_n^2$ density approximation for $n=7$, (Column 3) Indicator of whether $\mathbb{E}_{x_i}[\Phi_n^2(x,x_i)]>0.25\cdot \max(\mathbb{E}_{x_i}[\Phi_n^2(x,x_i)])$, (Column 4) Gaussian kernel density estimate, (Column 5) Indicator of whether $KDE(x)>0.25\cdot \max(KDE(x))$. (Top) No gap between different clusters, (Middle) Small gap between different clusters, (Bottom) Large gap between different clusters.  For all kernels, the bandwidth $\sigma=0.25$ was chosen to be half the median distance between points.}\label{fig:balloon}
\end{figure}
}
\end{uda}

\begin{uda}\label{uda:twomoon}
{\rm
We consider a mixture of two distributions in the so-called two moons data set.  These are point clouds concentrated near one-dimensional manifolds and the clouds are not linearly separable.  In Figure \ref{fig:twomoon} we show the results of our method with 1000 points, with the value of the parameter $n=6$.  
Our method estimates the relative supports of the two distributions, and demonstrates a minimal separation between the distributions.  On top of this, we show the selection of only two well chosen labled points leads to perfect classification of the entire data set using our algorithm.

\begin{figure}[ht]
\centering
\begin{tabular}{ccc}
\includegraphics[height=.15\textwidth]{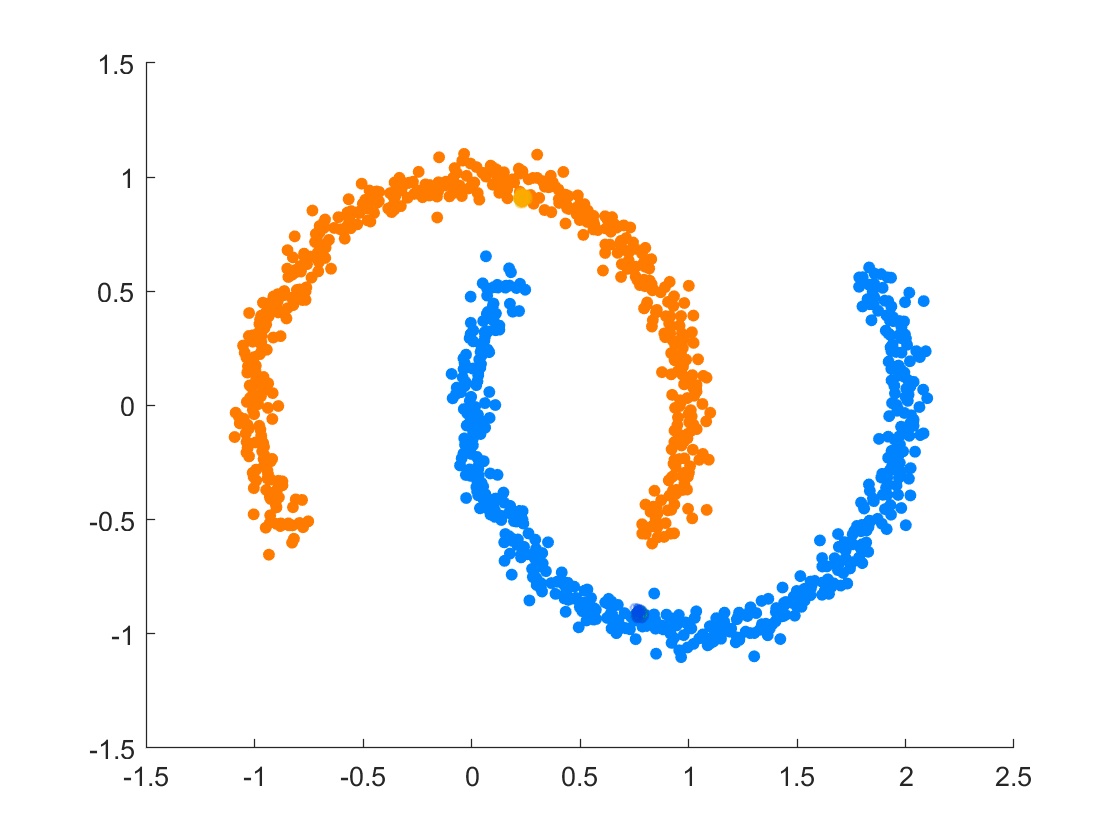} &
\includegraphics[height=.15\textwidth]{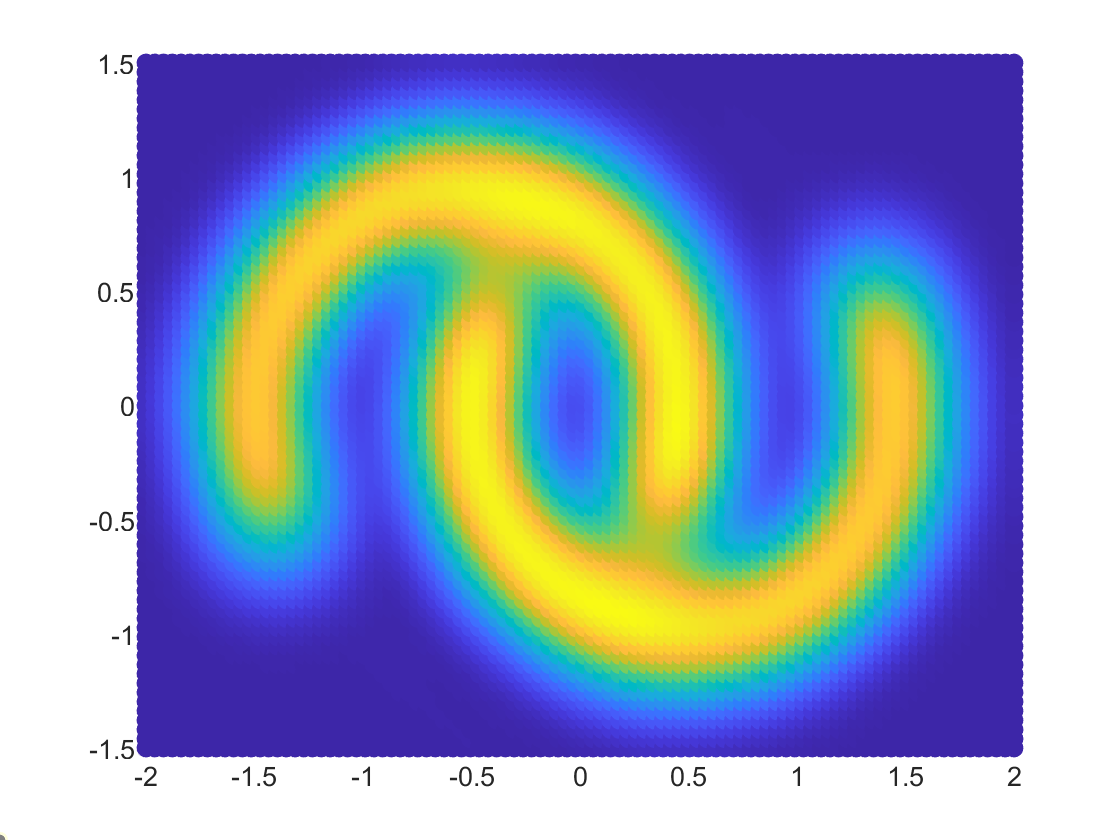} &
\includegraphics[height=.15\textwidth]{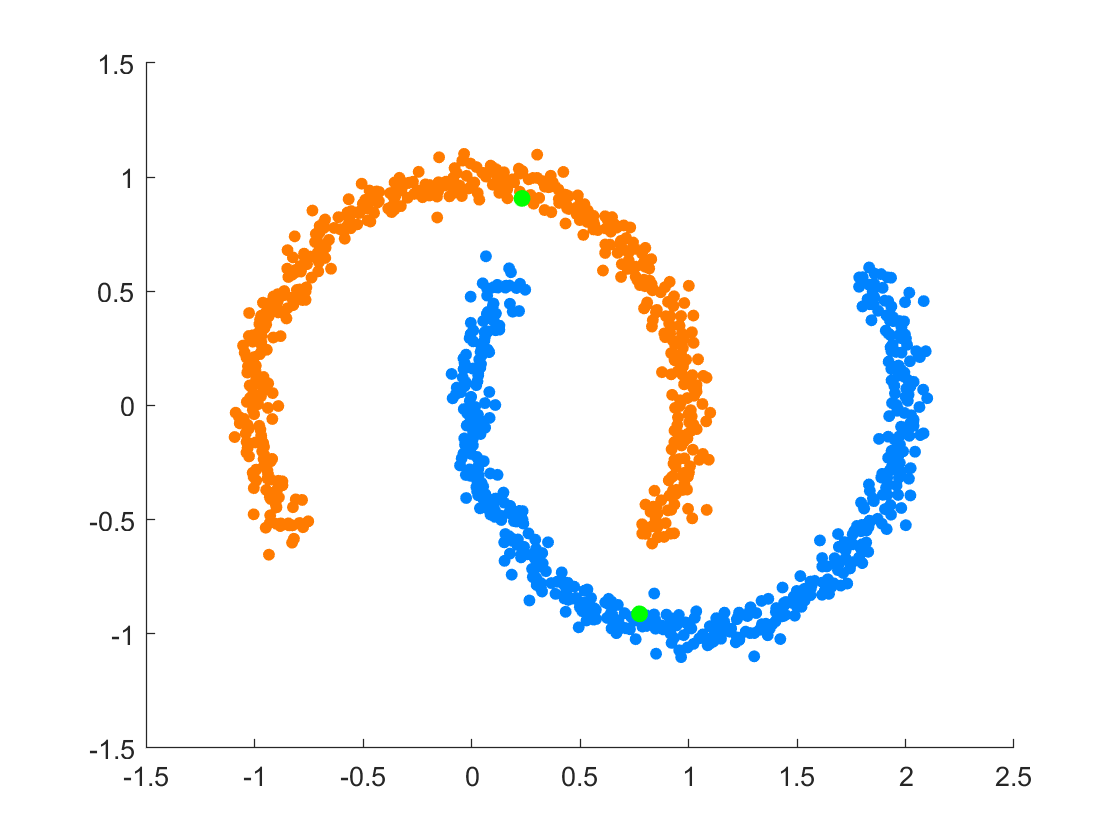} 
\end{tabular}
\caption{(Left) Data points, (Center) Density Approximation for $n=6$, (Right) Class prediction using our method with two labeled points.  Labeled points are highlighted in green.}\label{fig:twomoon}
\end{figure}
\qed}
\end{uda}

To summarize, we are interested in extending the theory summarized in Section~\ref{bhag:super} to overcome the following problems in particular.

\begin{enumerate}
\item Instead of having a linear combination of Dirac deltas, we have a linear combination of arbitrary probability measures, whose supports may be continua.
In turn, this requires the coefficients of these constituent distributions to be positive.
\item Instead of having values of $f(\j)$, we have random samples chosen from the distribution $\mu^*$. 
In some sense, this simplifies matters, since we could then discretize the integral in \eref{sso_integral} directly using the samples.
\item There is no minimal separation anymore.
We will replace it by a multiscale notion where we consider the supports of the constituent measures to be separated by $\eta$ for different values of $\eta$, with the remainder having a smaller and smaller probability.
\item The problem is inherently ill-posed; the number of constituent measures may be different at different levels of minimal separation.
\item There is no analogue of minimal magnitude $\mathfrak{m}$ here. 
Although one could pose the problem as the separation of a convex combination of probability measures, and assume a minimum on the coefficients involved, the probability measures themselves may be close to $0$ on continua.  
\end{enumerate}

\subsection{Relation to prior work}\label{bhag:priorwork}

A main difficulty in the theory of unsupervised learning is to define what one should understand by a cluster. 
For example, the correct number of clusters is sometimes defined in terms of graph cuts \cite{lafonncut, ling2019certifying}. 
This definition of clustering is not necessarily intuitive and leads to arbitrary bifurcations in the geometric structure of the data.
It is pointed out in \cite{dasgupta2005performance} that the notion of a cluster (in an unsupervised setting) needs be defined hierarchically.
We follow the philosophy in  \cite{dasgupta2005performance} by defining a hierarchical clustering that is tied to the order of our localized kernel $\Phi_n$, with the benefit that $\Phi_n$ provides a smooth decay with known decay rates.

Our paper also ties into the general field of active learning and machine teaching, which has grown rapidly in recent years with a large number of applications.  For the sake of relevance, we will focus on the subset of papers with mathematical guarantees for the proposed algorithm and that focus on assumptions on the data geometry \cite{maggioni2019learning, xiong2016active} rather than low-complexity classifiers \cite{hanneke2007bound, hanneke2015minimax}.
 Many of these results either establish lower bounds on the number of labels needed, or establish very conservative criteria of where to query labels in order to avoid sampling bias.
A general overview can be found in \cite{settles2009active,zhu2018overview, liu2017towards}.  

A number of works by Dasgupta and his collaborators \cite{dasgupta2006coarse, beygelzimer2009importance, dasgupta2008hierarchical, dasgupta2019teaching} have examined active learning over a class of hypotheses (i.e. classifiers) for minimax bounds on the generalization error, with probabilistic methods of choosing the points to sample.  
The errors are in terms of the VC-dimension of the hypothesis class.  
The closest connection to our work is the paper \cite{dasgupta2008hierarchical}, which assumes that there exists a hierarchical tree on the data structure and samples randomly from various bins.


Our paper also examines active learning problems in the context of hierarchical clusters. 
The major tool in this research is localized kernels \cite{mhasbk}, 
which have been applied in a variety of contexts 
\cite{loctrigwave, jaenpap, locsmooth, mauropap, hermite_recovery, treepap, warpaper}. 
Localized kernels also play a central role in the determination of components of a blind source signal, whether stationary or time-variant \cite{bspaper, chuimhasmaryke16, chuigaussian, hermite_recovery}. 
The super-resolution problem has  been considered in a hierarchical context \cite{li2020super}.
Our paper uses the super-resolution aspect of this theory with the harmonic analysis aspects in the context of active learning.

Our paper also connects to the analysis considered in two sample testing \cite{cheng2017two,witnesspaper},
where we used the theory of localized kernels and quadrature formulas on Eulidean spaces to determine where two probability distributions deviate from one another.  The theory in \cite{witnesspaper} is used in this paper to determine which distribution is dominant \textbf{at each point}, as well as a measure of uncertainty in the classification. 
The approach in \cite{witnesspaper} also helps to extend the theory of the witness function to multi-class classification.
Our paper builds on the witness function approach by constructing an indicator of where each cluster dominates while knowing only a few label samples.  We also use the witness function to determine the classification of uncertain points (see Section \ref{witnesssect}).

We  wish to note in particular \cite{maggioni2019learning}, which studies the active clustering framework for diffusion distance between points.  This establishes conditions under which the clusters are ``well-enough'' separated that each cluster will have a smaller in-radius for diffusion distance than the inter-cluster distance.  Unlike \cite{maggioni2019learning}, which constructs a single density estimate for the data, our algorithm constructs a hierarchical density estimate and, at each scale, throws away low-density points in order to guarantee well separated clusters.  We will compare to \cite{maggioni2019learning}, and to the hyperspectral imaging variant \cite{murphy2018unsupervised}, in Section \ref{bhag:app}.

\subsection{Outline}\label{bhag:outline}
We introduce the notation to be used in this paper in Section~\ref{bhag:notation}.
The main theorems are given in Section~\ref{bhag:main}, and proved in Section~\ref{bhag:proofs}.
We describe our algorithm to implement the main theorem in Section~\ref{bhag:alg} (together with Appendix~\ref{bhag:construction}) and illustrate the same using several examples in Section~\ref{bhag:app}.

\bhag{Notation and definitions}\label{bhag:notation}

In this paper, $q\ge 1$ is a fixed integer. 
For $\x=(x_1,\cdots,x_q)\in\RR^q$, we denote  by $|\cdot|_p$ the $\ell^p$ norm of $\x$.
For $\x\in\RR^q$, $r>0$, we denote
\be\label{balldef}
\BB(\x,r)=\{\y\in\RR^q : |\x-\y|_\infty \le r\}.
\ee
If $A, B\subseteq \RR^q$, $\x\in\RR^q$, $r>0$, then
\be\label{genballdef}
\dist(\x,A)=\inf_{\y\in A}|\x-\y|_\infty, \ \BB(A,r)=\{\x\in\RR^q : \dist(\x,A)\le r\},\ \dist(A,B)=\inf_{\x\in A}\dist(\x, B).
\ee

In Section~\ref{bhag:measuredef}, we describe the measure theoretic notions used in this paper. 
In Section~\ref{bhag:fscoredef}, we develop a measurement to test the accuracy our classification algorithms. 
The bare minimum definition of our localized kernel is developed in Section~\ref{bhag:hermitedef}; the details of the properties of this kernel will be developed in Section~\ref{bhag:proofs}.

\subsection{Measures}\label{bhag:measuredef}

The term measure will mean a positive, Borel measure on $\RR^q$. 
The support of a measure $\mu$, denoted by $\mathsf{supp}(\mu)$ is the set of all $\x\in\RR^q$ for which $\mu(\BB(\x,r))>0$ for all $r>0$. 
We will fix a probability measure $\mu^*$ on $\RR^q$.\\
We will use the following convention regarding generic positive constants.\\[2ex]

\noindent\textbf{Constant convention}\\

\textit{In this paper, the symbols $c, c_1,\cdots$ will denote generic constants depending only on the fixed quantities under consideration, such as $q$, $\mu^*$, and parameters $H$, $S$ to be introduced later. 
Their values may be different at different occurrences, even within a single formula.
There are occasions when we need to retain the values of some constants. 
Those constants whose value depend only on $q$ and $H$ will be denoted by $\kappa,\kappa_1,\cdots$, those whose value depends upon the measure as well will be denoted by $C, C^*, C_1,\cdots$.
}

\begin{definition}\label{def:mustardef}
Let $\mu^*$ be a probability measure on $\RR^q$. \\
{\rm (a)} The measure $\mu^*$ is called \textbf{detectable} if each of the following conditions is satisfied.
\begin{enumerate}
\item (\textbf{Compact support condition}) The support of $\mu^*$ is compact; in particular, 
$$
\mathsf{supp}(\mu^*) \subseteq \{\x\in\RR^q:|\x|_\infty \le C\}.
$$
\item (\textbf{Ball measure condition}) There exist $C_2$ and $ \alpha\ge 0$ such that 
\be\label{ballmeasurecond}
\mu^*(\BB(\x,r)) \le C_2 r^\alpha.
\ee
\item (\textbf{Density condition}) There exist $C_1>0$, $r_0>0$ such that (with $\alpha$ as in \eref{ballmeasurecond},
\be\label{densitycond}
\mu^*\left(\BB(\x,r)\right)\ge C_1r^\alpha, \qquad \x\in \supp(\mu^*), \ 0<r\le r_0.
\ee
\end{enumerate}
{\rm (b)}
The measure $\mu^*$ is said to have \textbf{fine structure} if it is detectable, and  there exists $\eta_0>0$ with the property that for every $\eta \in (0,\eta_0]$, there is in integer $K_\eta\ge 1$  and a partition $\S_{k,\eta}$, $k=1,\cdots, K_\eta+1$ of $\mathsf{supp}(\mu^*)$ such that each of the following conditions is satisfied.
\begin{enumerate}
\item (\textbf{Cluster minimal separation condition})
\be\label{minsepcondalt}
\mathsf{dist}(\S_{k,\eta}, \S_{j,\eta}) \ge 2\eta,\qquad k\not=j, \ k, j=1,\cdots, K_\eta.
\ee
\item (\textbf{Exhaustion condition})
$$
\lim_{\eta\downarrow 0} \mu^*(\S_{K_\eta+1,\eta}) =0.
$$
\end{enumerate}
{\rm (c)} The $\mu^*$ is said to have \textbf{fine structure in the classical sense} if it has a fine structure with $K_\eta=K$ for all $\eta\le \eta_0$, and there are measures $\mu_k$ such that $\mu^*=\sum_{k=1}^K\mu_k$ such that each $\S_{k,\eta}\subseteq \supp(\mu_k)$, $k=1,\cdots,K$.
\end{definition}

\begin{rem}\label{rem:oneormany}
{\rm Necessarily, a non-atomic, detectable probability measure $\mu^*$ has fine structure. 
The qualification of measures having fine structure in the classical sense is not an intrinsic property of such measures, but depends upon how the points in $\supp(\mu^*)$ are labeled.
In the \textbf{unsupervised setting} in the absence of minimal separation among various clusters, the notion of fine structure still enables us to construct reliable labels for various clusters by assigning  at each level $\eta$, the label $k$ with $\mathbf{S}_{k,\eta}$, $k=1,\cdots, K_\eta$. 
The exhaustion condition requires that the $\mu^*$-measure of the support of the ``left-over'' part of $\supp(\mu^*)$ tends to $0$ as $\eta\to 0$. 
\qed}
\end{rem}

\begin{uda}\label{uda:chui}
{\rm
Let $\mu^*=\sum_{k=1}^K a_k\delta_{\x_k}$, where $a_k$'s are positive, $\sum_k a_k =1$, and $\x_k\in\RR^q$. 
Then $\mu^*$ is compactly supported, and the ball measure condition is satisfied with $C_2=1$ and $\alpha=0$. 
The density condition is satisfied with $C_1=\min_k a_k$.
It easy to verify that $\mu^*$ has fine structure in the classical setting, with $\eta_0=(1/2)\min_{j\not=k}|\x_j-\x_k|_\infty$.  
\qed}
\end{uda} 

\begin{uda}\label{uda:infinitetrainint}
{\rm  With the setting as in Section~\ref{bhag:super},  we   let $q=1$, write $\TT=\TT^1$,  for $x\in\TT$, $r>0$.
For $x\in\TT$, $k\ge 1$, we let  $g_k(x)=(2/\pi)\chi_{\BB(\pi/2^k,\pi/{2^{k+1}}]}$, and define  $\mu^*$ by $\disp d\mu^*(x)=\left(\sum_{k=1}^\infty g_k(x)\right)dx$. 
If $x\in \supp(\mu^*)$, it is not hard to verify that for any $r\in (0,\pi/8]$, $cr\le \mu^*(\BB(x,r))\le 2r$ for a suitable constant $c>0$. 
So, $\mu^*$ is detectable.
If $L\ge 1$ is an integer, and $\eta=\pi/2^{L+2}$, we may write $K_\eta=L$, and consider the partition of the support of $\mu^*$ given by $\BB(\pi/2^k,\pi/{2^{k+1}}]$, $k=1,\cdots,L$, with $\mathbf{S}_{K_{\eta}+1}$ equal to the remainder of the support. 
Clearly $\mu^*$ has a fine structure. Since the number of elements in the partition depends upon $\eta$, this is the unsupervised setting.
On the other hand, we may view this also as a classical setting with $K=2$ as follows. Let
 $\mu_1$, $\mu_2$ be defined by  $\disp d\mu_1(x)=\left(\sum_{k=1}^\infty g_{2k}(x)\right)dx$, $\disp d\mu_2(x)=\left(\sum_{k=1}^\infty g_{2k-1}(x)\right)dx$. 
For integer $L\ge 2$, let  $\mu_{1,L}$, $\mu_{2,L}$ be defined by
$\disp d\mu_{1,L}(x)=\left(\sum_{k=1}^L g_{2k}(x)\right)dx$, $\disp d\mu_{2,L}(x)=\left(\sum_{k=1}^L g_{2k-1}(x)\right)dx$. 
Then $\dist( \supp(\mu_{1,L}),  \supp(\mu_{2,L}))=\pi/2^{2L+1}$, and
$$
\lim_{L\to\infty}\mu^*\left(\TT\setminus \left(\supp(\mu_{1,L})\cup \supp(\mu_{2,L})\right)\right)=0.
$$
Thus, only queries about labels on points can decide whether to interpret a measure with fine structure in the unsupervised or classical setting.
\qed
}
\end{uda}

\begin{uda}\label{uda:infinitetrainpts}
{\rm 
With the setting as in Example~\ref{uda:infinitetrainint}, let $\mu^*=\sum_{k=1}^\infty 2^{-k}\delta_{\pi/2^k}$. 
Then $\mu^*$ satisfies the compact support condition as well as the ball measure condition with $\alpha=0$, but not with any $\alpha>0$. The density condition is also not satisfied with $\alpha=0$. Thus, $\mu^*$ is not detectable.\qed
}
\end{uda}

\begin{uda}\label{uda:disjointmanifold}
{\rm
Let $K\ge 2$, $\XX_k$, $k=1,\cdots, K$  be mutually disjoint, compact, smooth, sub-manifolds (without boundary) of $\RR^q$  each having the dimension $d$. Let $\mu_k$ be the volume measure of $\XX_k$, and $\mu^*=\sum_{k=1}^K \mu_k$ normalized to be a probability measure.
For a large class of manifolds $\XX_k$, it is known that there exist constants $c_1, c_2>0$ such that for any point $\x\in\XX_k$ and $r>0$,  $c_1r^d\le \mu_k(\BB(\x,r))=\mu^*(\BB(\x,r)) \le c_2r^d$. 
Therefore, $\mu^*$ is detectable. It is trivial to see that $\mu^*$ has a fine structure in the classical setting.
 \qed
}
\end{uda}

\subsection{$F$-score}\label{bhag:fscoredef}
Our goal in this paper is to detect the support of $\mu^*$, and in the case when $\mu^*$ has a fine structure, to separate the components $\mathbf{S}_{k,\eta}$. 
If we attach the label $k$ with every data point in $\mathbf{S}_{k,\eta}$, then we need to discuss a measurement to assess the quality of our algorithms as classification tools. 
We recall the $F$-score described for a finite data set in \cite{diagraphcluster_ohiostate_2011}. If $\{C_1,\cdots,C_N\}$ are the obtained clusters from a certain clustering algorithm, and $\{L_1,\cdots,L_K\}$ is a partition of the data according to the (ground-truth) class labels (i.e., $L_k$ is the set of all points in the data set with the class label $k$), then one defines
$$
F_D(C_j)=2\max_{1\le k\le K}\frac{|C_j\cap L_k|}{|C_j|+|L_k|}, \qquad j=1,\cdots,N.
$$
The (micro--averaged) $F$-score is then defined by
\be\label{pre_fmeasuredef}
\mathcal{F}_D=\frac{\sum_j |C_j|F_D(C_j)}{\sum_j |C_j|}.
\ee

We interpret the cardinalities above as probabilities. 
In this paper, the total data is $\mathsf{supp}(\mu^*)$; the labeled sets (corresponding to $L_k$'s above) at separation level $\eta$ is  $\{\mathbf{S}_{k,\eta}\}_{k=1}^{K_\eta}$. Therefore, the $F$-score for the clusters $\{\mathcal{C}_j\}_{j=1}^N$ can be defined as follows: The analogue of $F_D(C_j)$ above is:
\be\label{f_for_one_cluster}
F_\eta(\mathcal{C}_j)=2\max_{1\le k\le K}\frac{\mu^*(\mathcal{C}_j\cap \mathbf{S}_{k,\eta})}{\mu^*(\mathcal{C}_j)+\mu^*(\mathbf{S}_{k,\eta})}.
\ee

Then
\be\label{fmeasuredef}
\mathcal{F}_\eta\left(\{\mathcal{C}_j\}_{j=1}^N\right)=\frac{\sum_{j=1}^N \mu^*(\mathcal{C}_j) F(\mathcal{C}_j)}{\mu^*\left(\bigcup_{j=1}^N\mathcal{C}_j\right)}.
\ee
Clearly, $\mathcal{F}_\eta\left(\{\mathcal{C}_j\}_{k=1}^N\right)\le 1$. If $N=K$, and $\mathcal{C}_k= \mathbf{S}_{k,\eta}$ for each $k$, then $\mathcal{F}_\eta\left(\{\mathcal{C}_j\}_{j=1}^N\right)= 1$. 
Thus, the closer the quantity $\mathcal{F}_\eta$ is to $1$, the better the quality of clustering with respect to the labels.

\subsection{Localized kernel}\label{bhag:hermitedef}

Our main tool in this paper are Hermite polynomials. In the univariate case, it is convenient to define the orthonormalized Hermite polynomial $h_k$ of degree $k$ recursively by
\bea\label{recurrence}
xh_{j-1}(x)&=&\sqrt{\frac{j}{2}}h_j(x) + \sqrt{\frac{j-1}{2}}h_{j-2}(x),\quad j=2,3,\cdots,\nonumber\\
&&h_0(x)=\pi^{-1/4},\ h_1(x)=\sqrt{2}\pi^{-1/4}x.
\eea
Writing $\psi_k(x)=h_k(x)\exp(-x^2/2)$, one has the orthogonality relation for $k, j\in \ZZ_+$,
\be\label{uniortho}
\int_\RR \psi_k(x)\psi_j(x)dx=\left\{\begin{array}{ll}
1, & \mbox{if $k=j$, }\\
0, & \mbox{if $k\not=j$.}
\end{array}\right.
\ee
In multivariate case, we adopt the notation $\x=(x_1,\cdots, x_q)$. The orthonormalized Hermite function is defined by
\be\label{multihermite}
\psi_\k(\x)=\prod_{j=1}^q\psi_{k_j}(x_j).
\ee
In general, when univariate notation is used in multivariate context, it is to be understood in the tensor product sense as above; e.g., $\k!=\prod_{j=1}^q k_j!$,
$\x^\k=\prod_{j=1}^qx_j^{k_j}$, etc.

Let $H: [0,\infty)\to [0,1]$ be a $C^\infty$ function, $H(t)=1$ if $t\in [0,1/2]$, $H(t)=0$ if $t\ge 1$. We define the \emph{localized kernel} by
\be\label{summkerndef}
\Phi_n(H;\x,\y)=\Phi_n(\x,\y)=\sum_{\k\in\ZZ_+^q}H\left(\frac{\sqrt{|\k|_1}}{n}\right)\psi_\k(\x)\psi_\k(\y).
\ee
The localization property is made precise in \eref{phin_loc1} below.

\bhag{Main theorems}\label{bhag:main}
In Section~\ref{bhag:super}, the quantity $\mathfrak{m}$ played several roles: the minimum value of the measure on arbitrarily small balls around points of its support and the threshold in Theorem~\ref{theo:bstheo}.
Here, the first role is played by the density condition.
We will take a multiscale approach by varying the minimal separation $\eta$ as defined in Definition~\ref{def:mustardef} and the threshold $\Theta$ to be used to determine sets of significant probabilities. 

The first theorem describes the determination of the support of $\mu^*$. For the sake of interpetability, we briefly discuss the important parameters associated with this theorem.  We fix $n$, which corresponds to the localization parameter for $\Phi_n$, and leads to an eventual clustering of all points as $n\rightarrow \infty$.  We set a threshold $\Theta$ to determine the level that a density estimate at $\x$ must attain for the fixed $n$ to be considered in the subsequent clustering.   
The parameter $\alpha$ corresponds to the effective dimension of the support of $\mu^*$, and determines how $n$ relates to the number of points $M$, the detectable minimal separation between clusters, and algorithmically determines how slowly we increase the localization of $\Phi_n$ and the number of points retained for that level of clustering.  These are the main parameters that appear in Algorithm \ref{alg:clustering}, though we use $\eta_n$ rather than $\alpha$ as a parameter by utilizing Eq. \eqref{add_n_cond_multi}.

\begin{theorem}\label{theo:main_single}
Let $\mu^*$ be detectable, $S>\alpha$,  $0<\Theta\le c_1$, $n\ge c_2$ be large enough, so that 
\be\label{ncond1}
\supp(\mu^*)\subseteq \BB(\bs 0,\kappa n). 
\ee
With $M\ge c_3n^{2\alpha}\log n$, let $\C=\{\x_1,\cdots,\x_M\}$ be independently sampled from the probability distribution $\mu^*$. We define
\be\label{gsetdef}
\mathcal{G}_n(\Theta, \C)=\left\{\x\in\RR^q : \sum_{j=1}^M \Phi_n(\x,\x_j)^2\ge \Theta \max_{1\le k\le M}\sum_{j=1}^M\Phi_n(\x_k,\x_j)^2\right\}.
\ee
Then with probability at least $1-c_4/M^{c_5}$,
\be\label{main_singlemeasure_det}
\supp(\mu^*)\subseteq \mathcal{G}_n(\Theta, \C) \subseteq \left\{\x\in \RR^q : \mathsf{dist} (\x,\supp(\mu^*))\le \frac{c_6}{\Theta^{1/(S-\alpha)}n}\right\}.
\ee
\end{theorem}

\begin{theorem}\label{theo:main_multi}
We assume the set-up as in Theorem~\ref{theo:main_single}. 
In addition, we assume that $\mu^*$ has a fine structure, and that
\be\label{add_n_cond_multi}
n\ge c(\eta\Theta)^{-1},\  n^\alpha \mu^*(\mathbf{S}_{K_\eta+1,\eta}) \le c_1\Theta.
\ee
Let
\be\label{partitiondef}
\mathcal{G}_{k,\eta, n}(\Theta,\C)=\mathcal{G}_n(\theta,\C) \cap \left\{\x\in\RR^q : \mathsf{dist}(\x, \mathbf{S}_{k,\eta})\le \frac{c_2}{n\Theta^{1/(S-\alpha)}}\right\}.
\ee
Then with probability exceeding $1-c_3M^{-c_4}$, the set $\mathcal{G}_n(\Theta,\C)$
 is a disjoint union of sets $\mathcal{G}_{k,\eta,n}(\Theta,\C)$, $k=1,\cdots, K_\eta$ such that
\be\label{separation}
\mathsf{dist}(\mathcal{G}_{k,\eta,n}(\Theta,\C), \mathcal{G}_{j,\eta, n}(\Theta,\C))\ge \eta, \qquad k\not=j, \ k, j=1,\cdots, K_\eta,
\ee
and for $k=1,\cdots, K_\eta$,
\be\label{comp_det}
\supp(\mu^*)\cap  \left\{\x\in\RR^q : \mathsf{dist}(\x, \mathbf{S}_{k,\eta})\le \frac{c_2}{n\Theta^{1/(S-\alpha)}}\right\}\subseteq \mathcal{G}_{k,\eta, n}(\Theta,\C) \subseteq  \left\{\x\in\RR^q : \mathsf{dist}(\x, \mathbf{S}_{k,\eta})\le \frac{c_2}{n\Theta^{1/(S-\alpha)}}\right\}.
\ee
\end{theorem}

Theorems \ref{theo:main_single} and \ref{theo:main_multi} combine to guarantee that, for a fixed $n$, the training data kept at a given threshold will be contained in a small tube around the true partitions $S_{k,\eta}$.  Similarly, points belonging to different clusters will have a minimal separation of at least $\eta$ (half the minimal separation of the true partitions).   This is a critical aspect to our algorithm, as we use a simple distance parameter to build a kNN network between training points, and use connected components to define the clusters.  Theorem \ref{theo:main_multi} guarantees that if we correctly choose this distance parameter, no connected component will span multiple true partitions.

\begin{uda}\label{uda:manifold_contd}
{\rm We continue the set-up as in Example~\ref{uda:disjointmanifold}. Then for $\eta\le \eta_0=\min_{1\le k\not= j\le K} \dist(\XX_k, \XX_j)$, the second condition in \eref{add_n_cond_multi} is satisfied trivially and both the conditions in \eref{ncond1} are satisfied for sufficiently large $n$. In particular, $K_\eta$ and the sets $\mathcal{G}_{k,\eta,n}(\Theta)$ do not depend upon $\eta$ if $\eta\le \eta_0$. The parameter $n$ controls how close one can get to the supports of the measures $\mu_k$.
\qed}
\end{uda}
\begin{uda}\label{uda:chui_contd}
{\rm
The same remarks as in Example~\ref{uda:manifold_contd} apply also in the set-up of Example~\ref{uda:chui}. In this case, the definition of $\mathcal{G}_{k,\eta,n}(\Theta)$ shows that the diameter of each of these sets is $\le 2\Theta/n$. In particular, if
 $$\hat{\x}_k=\argmax_{\x\in \mathcal{G}_{k,\eta,n}(\Theta)}\sum_{k=1}^K a_k\Phi_n(\x,\x_k)
 $$
satisfies $|\hat{\x}_k-\x_k|_\infty \le 2\Theta/n$. We note finally that the sum expression in the above expression can be computed using the Hermite moments of $\mu^*$; the precise location of $\x_k$'s or the values of $a_k$ need not be known. More impressively, the value of $K$ is found automatically rather than being required at the outset. This is consistent with the results in \cite{hermite_recovery}. \qed 
}
\end{uda}

\begin{rem}\label{rem:whyhermite}
{\rm
In a broad sense, we may view $(1/M)\sum_{j=1}^M \Phi_n(\x,\x_j)^2$ as a probability density estimator with kernel $\Phi_n(\x,\y)^2$ that localizes as $n\to \infty$. 
However, there are a number of important differences.
First, the measure $\mu^*$ is allowed to be singular with respect to the Lebesgue measure on $\RR^q$, so that there is no density to estimate. 
Second, we have demonstrated in \cite{witnesspaper} that in the case when $\mu^*$ is  absolutely continuous, the use of the special kernel $\Phi_n$ leads to a better estimation of the density than the customary Gaussian kernel, with sharper boundaries and faster empirical convergence in terms of the number of points.
Indeed, in applying this theorem, we may use both the bandwidth parameter as in the Gaussian kernel as well as the degree parameter $n$ to have a superior localization property. 
Third, in this paper we are not interested in approximating the measures themselves, just in finding their support. 
For the purpose of approximation, the oscillatory behavior of $\Phi_n$ leads to provably optimal approximation results. 
However, this very ability may lead to certain points in the support having zero values for the estimator.
Therefore, to find the support accurately, we need to use a positive kernel. 
We use $\Phi_n(\x,\y)^2$ because of the ease of evaluating certain integrals.
\qed
}
\end{rem}

\begin{theorem}\label{theo:fidelity}
We assume the set-up as in Theorem~\ref{theo:main_multi}. 
In addition, we assume that
\be\label{min_measure_cond}
\lim_{\eta\downarrow 0}\frac{\mu^*(\mathbf{S}_{K_\eta+1,\eta})}{\min_{1\le k\le K_\eta}\mu^*(\mathbf{S}_{k,\eta})} =0.
\ee
With probability $\ge 1-cM^{-c_1}$, the clusters $\mathcal{G}_{k,\eta, n}(\Theta, \C)$  satisfy
\be\label{fidelityest}
\lim_{\eta\to 0} \mathcal{F}_\eta\left(\{\mathcal{G}_{k,\eta, n}(\Theta, \C)\}_{k=1}^{K_\eta}\right)=1.
\ee
\end{theorem}

\begin{rem}\label{rem:classical}
{\rm 
In the classical setting (Remark~\ref{rem:oneormany}), $K_\eta=K$ for all $\eta$, and $\mathbf{S}_{k,\eta}\subseteq \supp(\mu_k)$. 
The exhaustion condition implies that $\mu^*(\supp(\mu_k)\setminus  \mathbf{S}_{k,\eta})\to 0$ as $\eta\to 0$.
In particular, the condition \eqref{min_measure_cond} holds trivially. 
Moreover, in the definition \eqref{f_for_one_cluster}, we may replace $\mathbf{S}_{k,\eta}$ by $\supp(\mu_k)$.
Thus,
Theorem \ref{theo:fidelity} guarantees that with high probability, the clusters of the high-density regions from each $n$ will grow to contain the entire supp$(\mu_k)$ for all $k$ and attain a perfect $F$-score (Eq. \eqref{fmeasuredef}).
\qed
}
\end{rem}

\begin{rem}\label{rem:fidelity1}
{\rm
In Theorem~\ref{theo:fidelity}, it is understood implicitly that the quantities $M$ and $\eta$ (and also $\Theta$) change with $n$ so as to satisfy the various conditions of Theorem~\ref{theo:main_multi}.
}
\end{rem}
\begin{rem}\label{rem:fidelity2}
{\rm
The statement of Theorem~\ref{theo:fidelity} is valid in a deterministic sense if the set-up of Theorem~\ref{theo:basic_multi} is assumed. In this case, we have, instead of \eref{fidelityest},
\be\label{fidelityest_det}
\lim_{n\to\infty} \mathcal{F}\left(\{\mathcal{S}_{k,\eta, n}(\theta)\}_{k=1}^{K_\eta}\right)=1.
\ee
}
\end{rem}

\bhag{Algorithmic considerations}\label{bhag:alg}
In order to apply the theory in Section \ref{bhag:main} to classification problems in practice, one needs to develop several further details. The theorems do not give a clear algorithm to find the clusters $\mathcal{G}_{k,\eta,n}$, and the choice of the parameters $n$, $\eta$, $\Theta$ need to be fixed experimentally in each application. 
We develop these details in this section.

We will describe the algorithm assuming that the data distribution $\mu^*$ has a fine structure in the classical sense defined in Definition~\ref{def:mustardef}(c); i.e., we assume that  there is a fixed set of labels involved, that does not change as the various parameters $n$, $\eta$, $\Theta$ change.  
We also note that although the theory in Section~\ref{bhag:main} (and Section~\ref{bhag:proofs}) allows one to decide what label if  any should be given to any point in the Euclidean space, it is convenient to assume in this section that all the points at which we wish to assign labels are already collected in a data set $\C$, analogous to the semi-supervised setting.

In Section \ref{conncompsect}
we discuss how to decide which points lie in a single cluster $\mathcal{G}_{k,\eta,n}$, and which of these one should query a label for.  
The theory suggests that we then assign the same label to every point in $\mathcal{G}_{k,\eta,n}$. 
In Section \ref{witnesssect}, we explain how to extend the known labels to the remaining points in the data set using the witness function approach in \cite{witnesspaper}.
To take advantage of the multiscale nature of the theory, we describe in Section \ref{interlayersect} how to transfer sampled labels at a coarse level to inform the clustering and label propagation at finer and finer levels. 
This discussion is summarized in an outline form in Algorithm \ref{alg:clustering}.   Finally, in Section \ref{compcomplex} we discuss the computational complexity of the algorithm.

\subsection{Connecting points in $\mathcal{G}_{k,\eta,n}$}\label{conncompsect}

In the classical setting, we may assume  that there exists some unknown label function $f:\RR^q \rightarrow \RR$.  Further, we assume it corresponds to a consistent clustering scheme such that, for small enough $\eta$ and  $\x\in \S_{k,\eta}$, we have that $f(\x) = k$ for $1\le k\le K$.    
The problem of active learning boils down to learning an estimate $\widehat{f}(\x)$ of $f(\x)$ for all $\x\in \bigcup_{k=1}^{K} \S_{k,\eta}$ given only a small set of points $\A\subset \C$ at which $f$ is actually known.  
The key difference between this and semi-supervised learning is that, in our case, $\A$ can be chosen in a data-dependent fashion prior to querying the function.   We note that the choice of label function $f$ may be any layer of some hierarchical tree of labels \cite{dekel2004large}, but we assume that  $f$ is fixed at the start of the algorithm.

 Theorem \ref{theo:main_multi} guarantees the existence of sets $\mathcal{G}_{k,\eta, n}(\Theta,\C)$ that satisfy a minimal separation condition \eqref{separation}.  However, in order to use this result to propagate learned cluster labels effectively, it is important to determine which data points $\x\in \C$ are in a particular cluster, $\x\in\C \cap \mathcal{G}_{k,\eta, n}(\Theta,\C)$.  This is necessary  both to:
\begin{enumerate}
\item decide the set $\A$ of points $x\in C$ we wish to query for a label $f$, and
\item propagate the labels from $\A$ to the rest of $\mathcal{G}_n(\Theta,\C)$ in a way that guarantees the estimate $\widehat{f}(\x)$ agrees with $f(\x)$ itself on points $\x\in\mathcal{G}_n(\Theta,\C)$.
\end{enumerate}

The insight for constructing the clusters $\mathcal{G}_{k,\eta,n}$ comes from three observations: (1) $\mu^*$ has a fine structure of minimal separation of $2\eta$ for some $\eta$, (2)  the theoretical guarantee that the clusters $\mathcal{G}_{k,\eta,n}$ satisfy \eqref{separation} for a finite $n$, and (3) we can decrease $\eta$ by  increasing $n$ as in \eqref{add_n_cond_multi}. 
For this reason, we will fix $\Theta$ and consider a minimal separation $\eta_n$ that depends on $n$ in a way that satisfies \eqref{add_n_cond_multi}.
Given this separation, we construct a nearest neighbor graph $G$ on $\mathcal{G}_n(\Theta,\C)$ with edge set $E=\{(\x,\y)\}\in \mathcal{G}_n(\Theta,\C) : |\x-\y|_2<\eta_n/2\}$.  In this way, we are guaranteed that each connected component $C_{\ell,\eta_n,n} \subset \C $ actually satisfies $C_{\ell,\eta_n,n}\subset \mathcal{G}_{k,\eta_n,n}(\Theta,\C)$ for some $k$.  
This implies that if we query and obtain a label $f(\x)$ at some point $\x\in C_{\ell,\eta_n,n}$, then  we are guaranteed that all other points in $C_{\ell,\eta_n,n}$ also have label $f(\x)$.  
Note that the number of connected components, which we'll call $\widetilde K_n$, satisfies $\widetilde K_n \ge K$.  This is because a single class can consist of multiple connected components of $G$ at separation $\eta_n$.  

The only other problem to address in this framework is to select the points $\A$ at which to query a label.  While theoretically any point in the connected component would be sufficient, the most reliable point to choose is the mode of the cluster, i.e.,
\be\label{peaksample}
\x^* = \arg\max_{\x\in C_{\ell,\eta_n,n}} \sum_{j=1}^M \Phi_n^2(x,x_j).
\ee
The argument for this choice is a heuristic one; if there do exist points in $C_{\ell,\eta_n,n}$ with the incorrect label (i.e., some cluster couldn't be fully resolved at level $\eta_n$) then they are more likely to lie at the boundary of the connected component, and thus have a lower empirical density.

\subsection{Classification of the remaining points}\label{witnesssect}
For any $\eta$ there may exist low density points that lie outside $\mathcal{G}_n(\Theta,\C)$ that may not be classified at level $n$.  While this is not an issue in the continuum limit due to Theorem \ref{theo:fidelity},  we are constrained in most applications to finite budget of labels to be sampled.  
This implies that
we must use a finite $n$, as we cannot realistically split $\C$ into $M$ different clusters and sample each point's label separately; defeating thereby the purpose of the exercise.  Because of this, we must have a trade-off between scaling $n$ until we've classified all points accurately, and stopping at a finite $n$ to make our best predictions of labels for points in $S_{K + 1}$. 

We propose to classify these additional points through the witness function approach developed in \cite{witnesspaper}.  To summarize in this context, we define each class estimate to be $\widehat{S}_{k,\eta_n} = \{x_i \in \C:\widehat f(x_i) = k\}$.  Then we construct a witness function for each class
\bea\label{witness}
\widehat F_k(\x) = \frac{1}{|\widehat{S}_{k,\eta_n}|} \sum_{\x_j\in \widehat{S}_{k,\eta_n}} \Phi_n(\x, \x_j), & \textnormal{ for } \x \in \C\setminus \mathcal{G}_n(\Theta, \C).
\eea
The proposed algorithm  assigns a label  to $\x$ given by
\be 
\widehat{f}(\x) = \arg \max_{1\le k\le K} \widehat{F_k}(\x),
\ee
and determine the certainty of classification through a permutation test as in \cite{witnesspaper}.  
Note that we will refer to the points assigned in $\mathcal{G}_{n}(\Theta,\C)$ and labeled according to Section \ref{conncompsect} as \textbf{\textit{confident points}}, and the points assigned by the witness function, $\C\setminus \mathcal{G}_{n}(\Theta,\C)$, as \textbf{\textit{uncertain points}}.  These uncertain points fall outside our guarantees in Theorem \ref{theo:main_multi}, other than the fact that the set becomes empty as $n\rightarrow \infty$.

\begin{algorithm}[!h]
\caption{Cautious Active Clustering}\label{alg:clustering}
\hspace*{\algorithmicindent} \textbf{Input:} $\C\subset\mathbb{R}^q$, $nmax$, $\Theta$, $\tau$ \\
\hspace*{\algorithmicindent} \textbf{Output:} $\widehat{f}$, $\mathcal{G}_{nmax}(\Theta,\C)$ 
\begin{algorithmic}
\State $\mathcal{A} \gets \emptyset$ \Comment{Set of points at which label is queried, together with the corresponding labels.}
\While {$n\le nmax$} 
    \State $\C_{n,\Theta} \gets \mathcal{G}_n(\Theta,\C) \cap \C$ using \eqref{gsetdef}.
    \State $E \gets \{(x_i, x_j): x_i, x_j\in \C_{n,\Theta} \textnormal{ and } |x_i - x_j|_2<\eta_n/2\}$ for $\eta_n$ as a function of $n$ as in \eqref{add_n_cond_multi}
    \State Construct graph $G = (\C_{n,\Theta}, E)$ 
    \State $\{C_{n,\ell}\}_{\ell=1}^{\widetilde K_n} \gets $connectedComponents$(G)$ (See Section \ref{conncompsect}) 
    \State Set $\mbox{flag}(\ell)=0$ for all $\ell$ \Comment{When we are done with this $n$, $\mbox{flag}(\ell)=1$ for all $\ell$.} 	
    \For {$\ell \le \widetilde K_n$}
    		\If {$C_{n,\ell} \cap \mathcal{A} = \emptyset$} 
    			\State $x_i \gets \arg\max\limits_{x_i\in C_{n,\ell}} \sum\limits_{j=1}^M \Phi_n(x_i, x_j)^2$ \Comment{Point at which label is sought}
    			\State $\mathcal{A} \gets \mathcal{A} \cup (x_i, f(x_i))$ \Comment{Update $\mathcal{A}$; this set does not lose points.}
    			\State $\widehat f(x_j) \gets f(x_i)$ $\forall x_j\in C_{n,\ell}$ \Comment{Extend label to the whole component}
    			\State $\mbox{flag}(\ell)=1$ \Comment{Done for this value of $\ell$.}
    		\Else
    			\If {$\forall x_i \in C_{n,\ell} \cap \mathcal{A},$ $f(x_i) = c_\ell$}
    				\State $\widehat f(x_j) \gets c_\ell$ $\forall x_j\in C_{n,\ell}$ \Comment{Extend label to the whole component}
    				\State $\mbox{flag}(\ell)=1$ \Comment{Done for this value of $\ell$.}

    			\EndIf
    		\EndIf
    	\EndFor
    	\If  {$\mbox{flag}(\ell)=1$ for all $\ell$,} 
    	 \State $n\gets n+\mbox{step}$ \Comment{Done for this pass, go to next level}\\
    	 \Comment{ to ensure that we captured all points that could be captured.}
        \Else 
         \State Increase threshold $\Theta \gets \tau \Theta$,  (See Section \ref{interlayersect}).  \Comment{Prune the graph.} 
          
    	 \EndIf
\EndWhile

\State $\C_{\widetilde K_{nmax} + 1} \gets \C\setminus \mathcal{G}_{nmax}(\Theta_{nmax},\C) $ using \eqref{gsetdef}\Comment{Uncertain points}
\State $\widehat{S}_{k,\eta_{nmax}} \gets \{x_i:\widehat f(x_i) = k\}$ 
\State $\widehat f(x_j) \gets \arg\max_k \frac{1}{|\widehat{S}_{k,\eta_{nmax}}|} \sum_{x_i\in \widehat{S}_{k,\eta_{nmax}}} \Phi_{nmax}(x_j, x_i)$ for $x_j\in \C_{\widetilde K_{nmax} + 1}$ \\  \Comment{Extend labels to uncertain points using witness function (See Section \ref{witnesssect}).}

\end{algorithmic}

\end{algorithm}

\subsection{Learning across layers}\label{interlayersect}
As described to this point, the algorithm for learning $\widehat{f}$ is computed independently at each $n$.  
However, this is not efficient from a label sampling perspective, as there may be significant information already learned at $n_0$ for $n>n_0$.  
We consider an increasing hierarchy of the parameter $n$, $\{n_i\}_{i=1}^{\infty}$.
This similarly determines a hierarchy of decreasing $\eta$, $\{\eta_i\}_{i=1}^\infty$ such that $\eta_j < \eta_i$ if $j>i$.  

Let $\mathcal{A}_i\subset \bigcup_{\ell=1}^{\widetilde K_{n_i}} C_{\ell,\eta_{i}, n_i}$ be the small collection of points at which $f$ was sampled. 
By definition of $\mu^*$ being detectable, $\mathcal{A}_i \subset \bigcup_{\ell=1}^{\widetilde K_{n_j}} C_{\ell,\eta_j,n_j}$ as well for $j>i$. 
This means that many of the connected components $\{ C_{\ell,\eta_j,n_j} \}_{\ell=1}^{\widetilde K_{n_j}}$ already have a member $\x\in C_{\ell,\eta_j,n_j}$ such that $\x\in \mathcal{A}_i$.  Thus, we must only sample the $\widetilde K_{n_j}- \widetilde K_{n_i}$ clusters that do not already contain a sample.

We also wish to comment on the stability of the connected component separation across levels.  While we are 
examining a minimal separation of $\eta_j$, this is not a known value a priori.  Even when estimated, it is possible that two clusters 
have separation just greater than $\eta_j$, and that removing low-density points with threshold $\Theta$ does not help increase the separation of clusters in $\mathcal{G}_{n_j}(\Theta, \C)$ sufficiently.
Fortunately, this can be easily detected in the situation that subsets were disconnected at $n_i$ for $j>i$.  In this situation, $\mathcal{A}_i$ will contain two points $\x, \y$ with different labels from level $n_i$ such that $\x, \y\in C_{\ell,\eta_j,n_j}$.  When this occurs, it is a simple fix to slightly increase $\Theta$ until $\x$ and $\y$ fall in different clusters.   This can be done with a parameter $\tau>1$ that is described in Algorithm \ref{alg:clustering}, basically increasing the thresholding of low density points before redefining the clusters.
This disagreement can thus be easily fixed at level $n_j$ and allows for a more robust clustering that must remain consistent across levels.  
Similarly, one could decrease the estimate of $\eta_j$ and rerun the algorithm.

As a final note, it's possible to increase $\Theta$ or decrease $\eta$ only for points in $C_{\ell,\eta_j,n_j}$ rather than on all $\C$.  This will lead to a different $\Theta,\eta$ in different regions of space, but the set of points and neighborhoods will be a proper subset of $\mathcal{G}_n(\Theta,\C)$.  


\subsection{Computational Complexity}\label{compcomplex}
The computational complexity of this algorithm depends mostly on construction of the kernel $\Phi_n$ at multiple $n$.  
The recurrence formula \eqref{recurrence} enables us to compute $\psi_n$ for any $n$, keeping in storage only the values of $\psi_{n-2}$ and $\psi_{n-1}$.   This means the computation depends only on the largest $n$, and depends quadratically in the number of points $M$, as do most kernel methods.  This leads to a complexity of 
$O(n^2M^2q)$.  This can be precomputed, and is followed by Equation \eqref{reduceproj}, which is complexity $O(n^4M^2)$ per projection, resulting in a final computation for the kernel with $O(qn^6M^2)$ flops.
We note that $\Phi_n$ is a $q$-variate weighted polynomial of total degree $n^2$ in each variable. 
Therefore, one expects a complexity of $\O(n^{2q}M^2)$ in the computation, as it would be if a monomial or tensor product Chebyshev polynomial basis was used to construct the kernel.
However, the special property of Hermite polynomials given in Equation \eqref{reduceproj} enables us to evaluate the kernel with complexity linear in $q$.
In practice and in Section \ref{bhag:app}, the $n$ can remain quite small while still resulting in significantly improved performance.  
 As with most kernel methods, this can be improved below a quadratic dependence on $M$ using nearest neighbor or $\varepsilon-$ball algorithms to estimate where contribution will be negligible.  
 Beyond this computation, the most expensive aspect is the connected components computation after thresholding $\mathcal{C}_{n,\Theta}$, which is $O(M + M^2)$ for most connected component algorithms because the number of edges is proportional to $M^2$.  All other aspects depend linearly only on the number of connected components $\widetilde K_n$, which is at most $M$ and in practical terms significantly smaller.

\bhag{Applications}\label{bhag:app}
In this section, we consider a number of applications to both synthetic and real data sets.  
For the synthetic data, we consider problems that either do not have a minimal separation, or has a very small separation relative to the inter-cluster radius (Section \ref{synthetic}).  
This is a particularly difficult set of examples for clustering and active learning problems because many algorithms, such as $k$-means, expect clusters to be somewhat isotropic (i.e., similar variance in all directions).  We also consider the latent space of a variational autoencoder that embedded the MNIST data set into a 2D latent space (Section \ref{mnistsect}).  This problem again poses difficulty for traditional clustering algorithms, as we have purposefully chosen a latent space dimension that leads to no minimal separation between some label clusters, and even partial overlap of different labels.  Even in this setting, we demonstrate strong classification accuracy based on a small number of samples.

As our main set of applications, we consider our active clustering framework on hyperspectral image pixel classification (Section \ref{hsi}).  Traditionally, this is an application that requires non-Euclidean clustering methods, and a very large number of pixel labels.  Similarly, there is rarely a minimal separation between clusters, made worse by the fact that pixels can even be a mix of multiple labels.  We compare our algorithms to the current state-of-the-art active clustering algorithm on HSI, the LAND algorithm \cite{maggioni2019learning}, which uses a Gaussian kernel density estimate and diffusion geometry to define the cluster centers and boundaries.

\subsection{Synthetic examples without minimal separation}\label{synthetic}
\begin{figure}
\centering
\begin{tabular}{ccc}
\includegraphics[width=.25\textwidth]{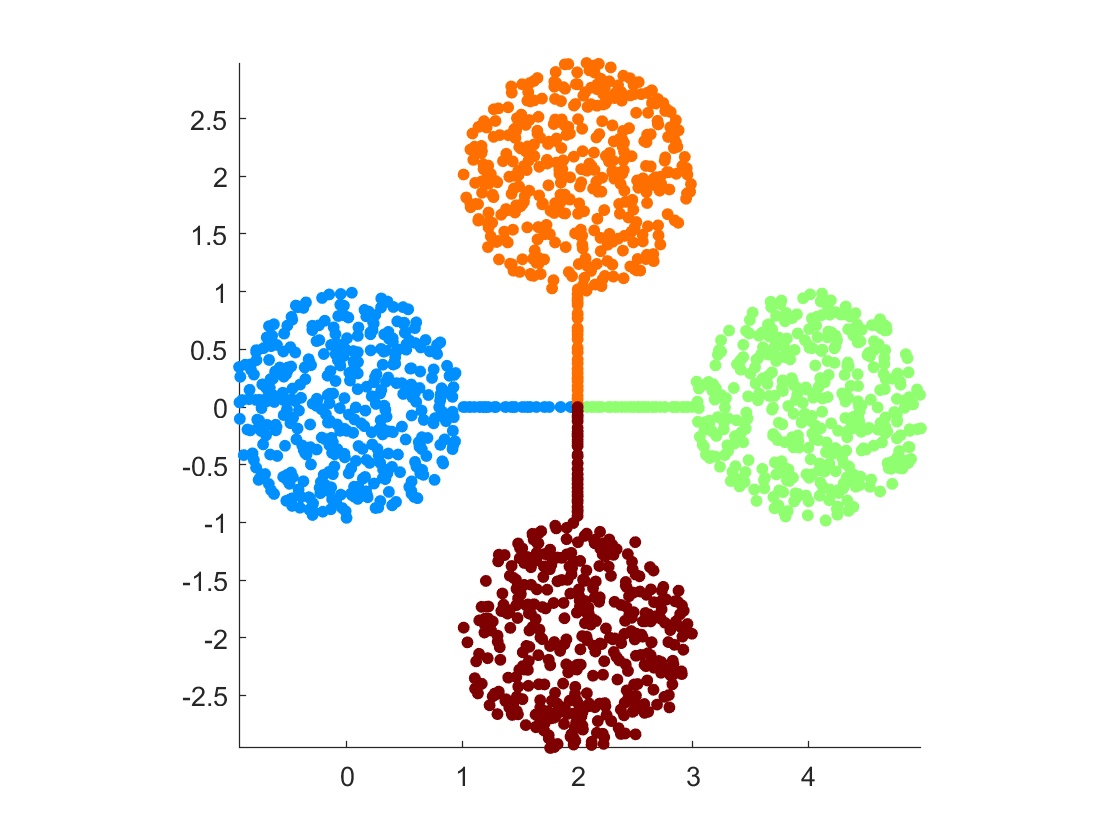} & 
\includegraphics[width=.25\textwidth]{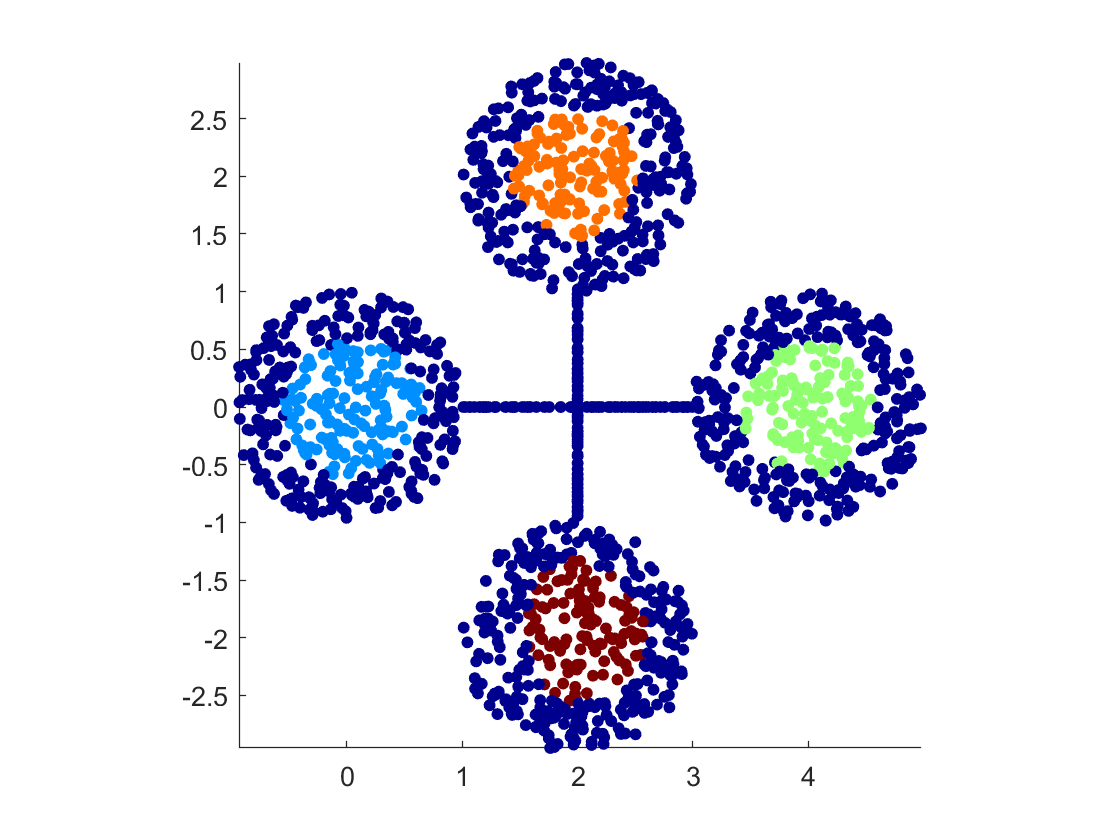} &
\includegraphics[width=.25\textwidth]{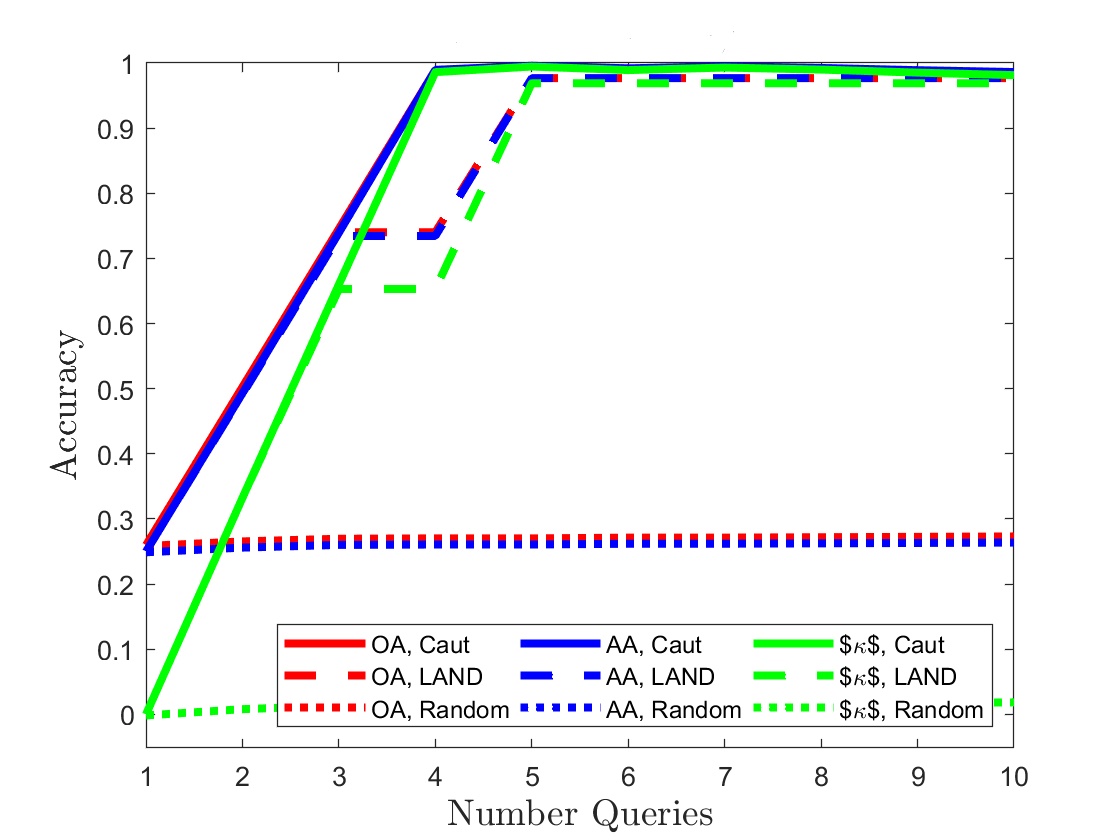} 
\end{tabular}
\caption{(Left) Clustered data with no minimal separation. Color corresponds to label. (Center) Example of cautious clustering approach with 4 labels queried and $n=4$.   Dark blue labels are uncertain points, i.e., points below the density threshold. (Right) Different measures of error after witness function propagation from confident points (our algorithm cautious active clustering), and comparison to LAND \cite{maggioni2019learning}.  Note that LAND performs similarly by the time 5 labels have been queried, and plateaus at a similar overall accuracy to our algorithm.}\label{fig:bottleneck}
\end{figure}

We examine the problem of learning with few labels on synthetic data that violates traditional clustering assumptions.  
In the first example in Figure \ref{fig:bottleneck}, we use the data that does not have a minimal separation between clusters.  This is a setting in which filtering by density significantly benefits the clustering algorithm, as the clusters in Figure \ref{fig:bottleneck} have long tails of low density.

\begin{figure}[!h]
\centering
\begin{tabular}{cc}
\includegraphics[width=.25\textwidth]{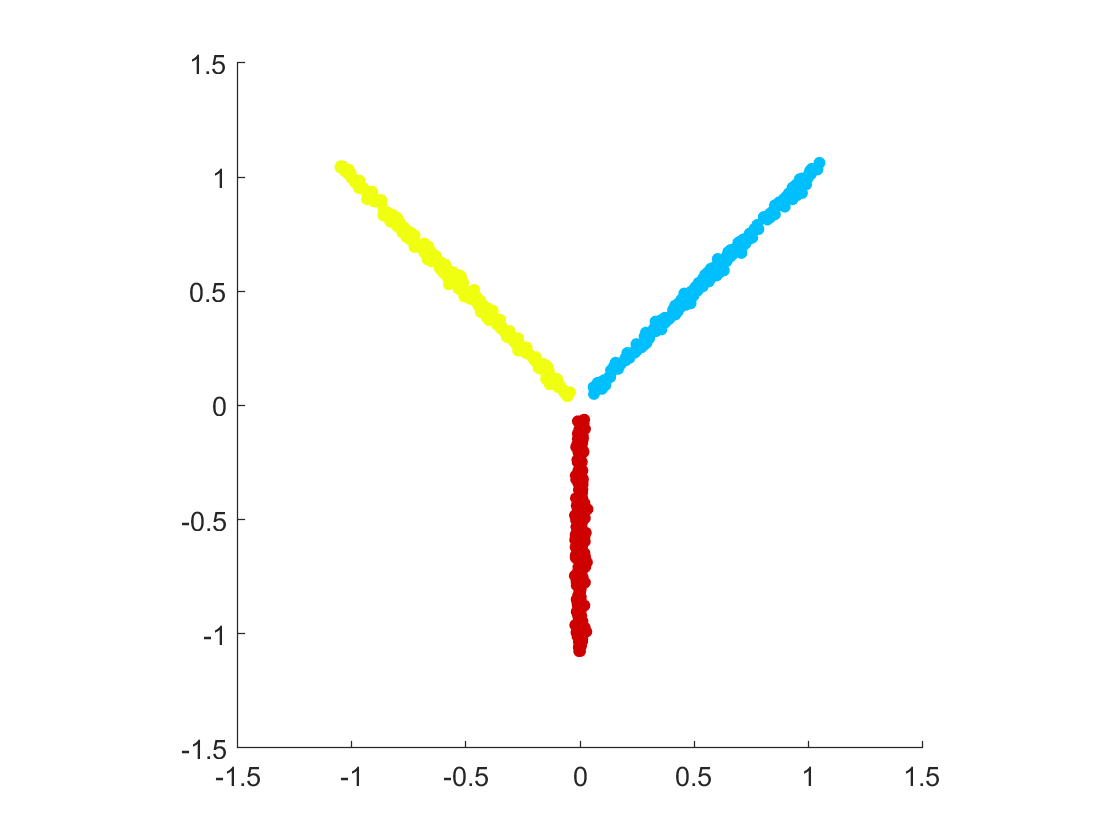} & 
\includegraphics[width=.25\textwidth]{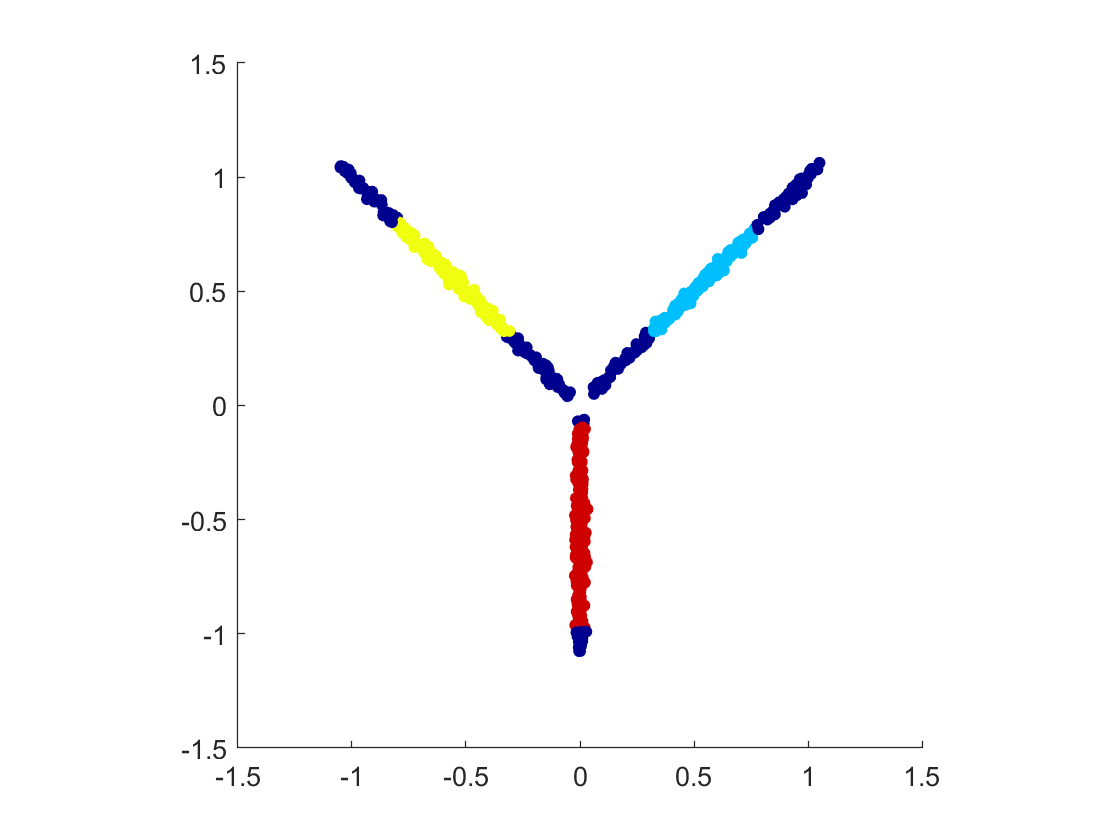} \\
\includegraphics[width=.25\textwidth]{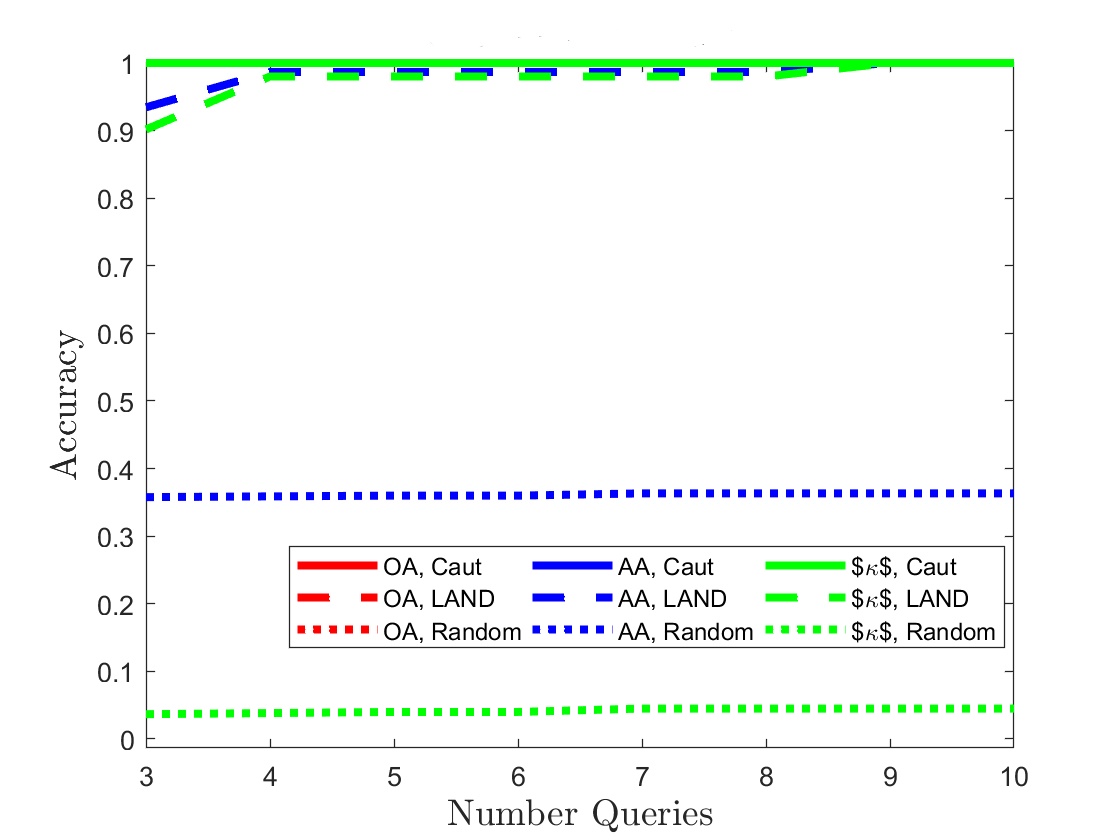} & 
\includegraphics[width=.25\textwidth]{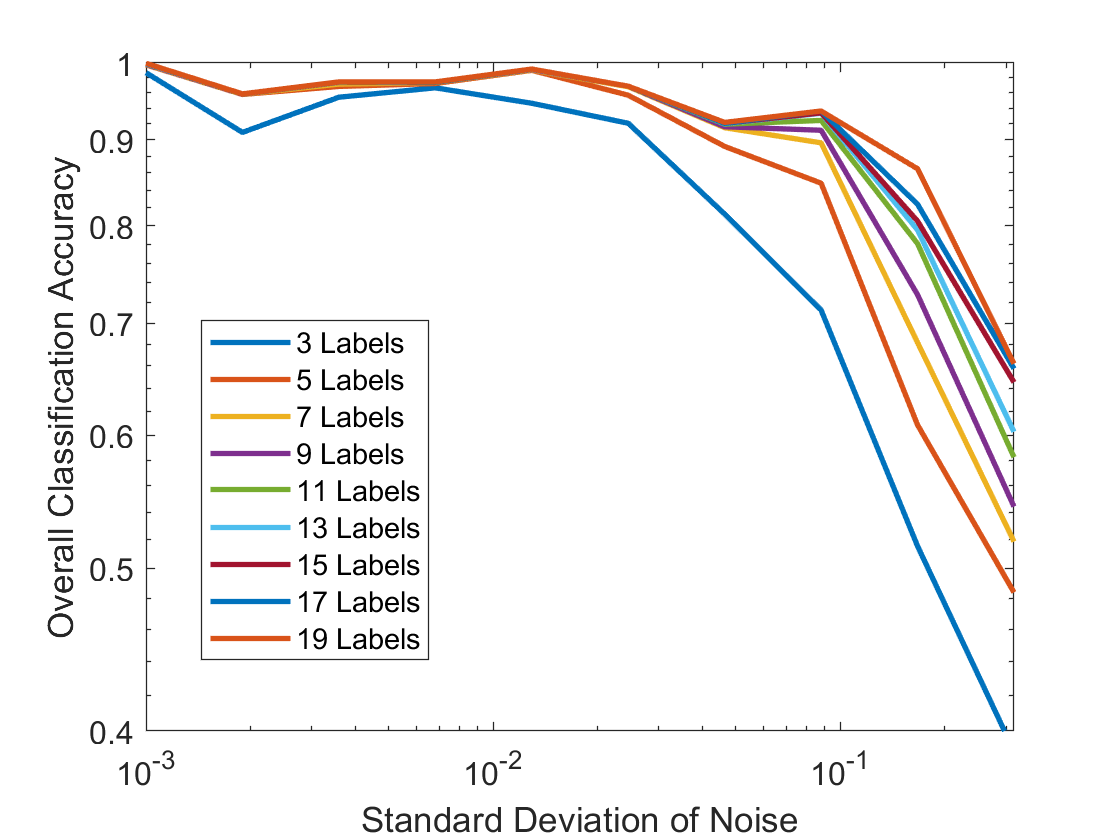} 
\end{tabular}
\caption{(Top Left) Clustered data with small minimal separation and no density peaks. Color corresponds to label. (Top Right) Example of cautious clustering approach  with 3 labels queried and $n=4$.  Dark blue labels are uncertain points, i.e., points below the density threshold. (Bottom Left) Different measures of error for our cautious active clustering algorithm, and comparison to LAND \cite{maggioni2019learning}.  Note that LAND requires 9 labels to attain perfect classification, as opposed to 3 for our algorithm.  (Bottom Right) Accuracy of our algorithm on same data with varying levels of Gaussian noise added to the points.  Different curves correspond to the allowed budget of sampled points, and curves are averaged over 25 realizations.  }\label{fig:Ydata}
\end{figure}

In a second example in Figure \ref{fig:Ydata}, we consider data that has a very small minimal separation, but the density remains relatively constant between the middle of the clusters and their tails.  In this setting, it is critical to have a highly localized kernel for density estimation and defining similarity between points.  Because the origin is close to all three of the clusters, using a kernel with poor localization would lead to points near the origin having a higher estimated density than any of the points sampled from the actual distributions.
This would lead to sampling points far away from the centers of the clusters. 

\begin{figure}
\centering
\begin{tabular}{ccc}
\includegraphics[width=.25\textwidth]{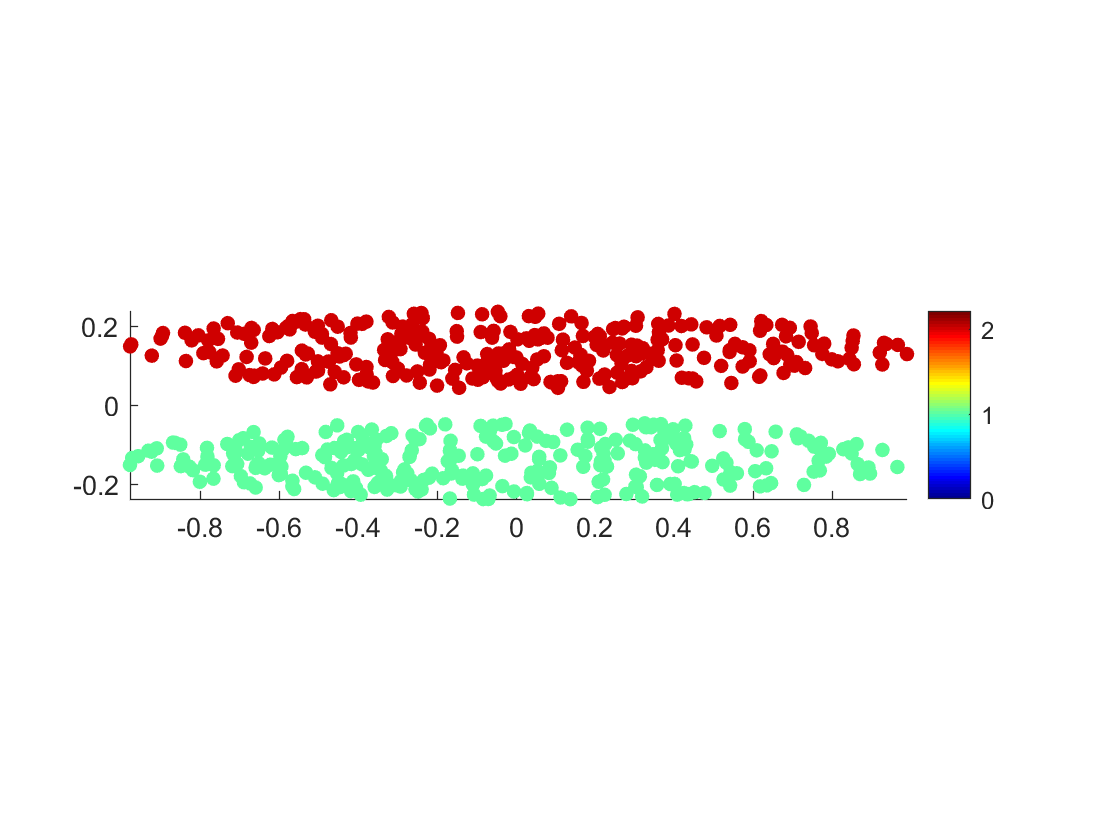} & 
\includegraphics[width=.25\textwidth]{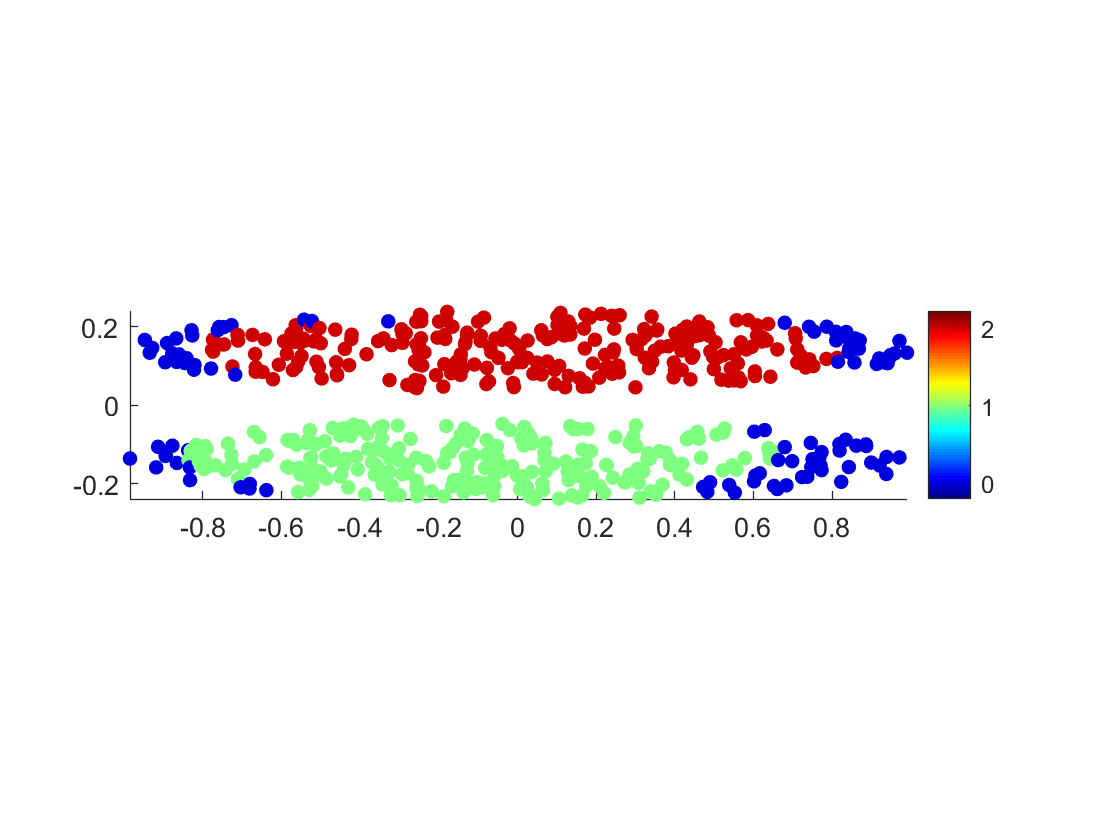} & 
\includegraphics[width=.25\textwidth]{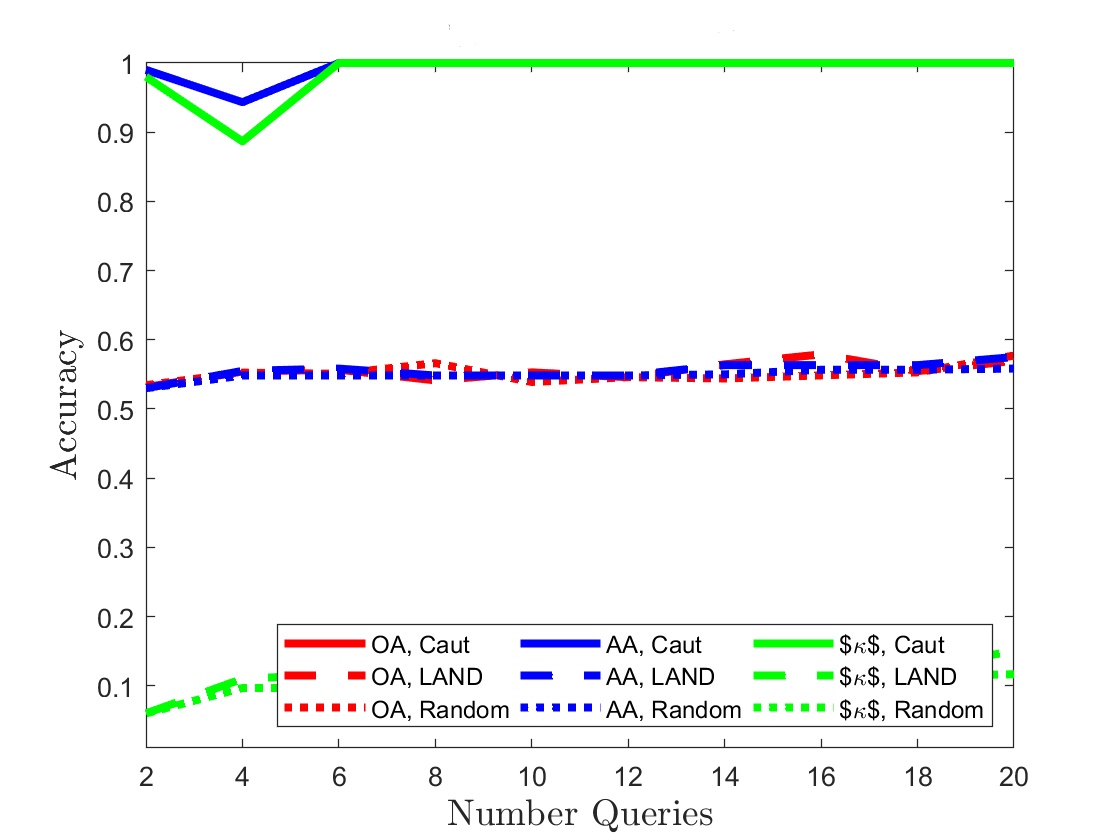}
\end{tabular}
\caption{(Left) Clustered data with small minimal separation compared to inner radius. Color corresponds to label. (Center) Example of cautious clustering approach  with 2 labels queried and $n=4$.  Dark blue labels are uncertain points, i.e., points below the density threshold. (Right) Different measures of error after witness function propagation from confident points (our algorithm cautious active clustering), and comparison to LAND \cite{maggioni2019learning}.  Note that LAND performs significantly worse than our algorithm in this case, mostly due to the small minimal separation between clusters that leads to false edges between the two clusters without first performing thresholding.}\label{fig:gauss}
\end{figure}

In a final example in Figure \ref{fig:gauss}, we consider data that has a very small minimal separation compared to their internal maximum radius.  In this setting, it is important to have a flexible method for connecting points within cluster, like connected components, that allows for connecting far apart points as long as there exists a path between them.  Without this, one requires a large number of points with queried labels spaced throughout the cluster in order to propagate the labels effectively.  

\subsection{MNIST generative models}\label{mnistsect}
The next set of experiments revolve around estimating regions of space corresponding to given classes, and determining which regions of the latent space correspond to which class labels.  This problem has been of great interest in recent years with the growth of generative networks, namely various variants of generative adversarial networks (GANs) \cite{goodfellow2014generative} and variational autoencoders (VAEs) \cite{kingma2013auto}.  Each has a low-dimensional latent space in which new points are sampled, and mapped to $\RR^q$ through a neural network.  While GANs have been more popular in literature in recent years, we focus on VAEs in this paper because it is possible to query the locations of training points in the latent space.  A good tutorial on VAEs can be found in \cite{doersch2016tutorial}.

We examine this problem with the well known MNIST data set \cite{lecun2010mnist}. This is a set of handwritten digits $0\cdots 9$, each scanned as a $28\times 28$ pixel image.  There are $50000$ images in the training data set, and $10000$ in the test data.

In order to select the latent space for this data set, we construct a three layer VAE with encoder $E(\x)$ with architecture $784-500-500-2$  and a decoder/generator $G(\z)$ with architecture $2-500-500-784$, and for clarity consider the latent space to be the 2D middle layer.  We have purposely chosen a 2D latent space because this leads to varying levels of separation between label clusters, including overlapping clusters of commonly confused digits (e.g., mixing $4$'s and $9$'s, and mixing $3$'s, $5$'s, and $8$'s).  
For this reason, we look at both a fine grained and coarse grained label set.  In fine grained, we have the traditional 10 labels, in coarse grained we create 7 labels by merging the $4$'s and $9$'s and merging $3$'s, $5$'s, and $8$'s.  This creates an example of the hierarchical label structure as described in the paper, as well as clusters with no minimal separation.

\begin{figure}
\centering
\begin{tabular}{ccc}
\includegraphics[width=.3\textwidth]{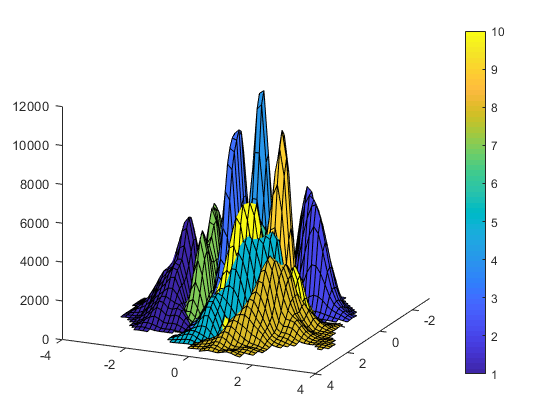} & 
\includegraphics[width=.3\textwidth]{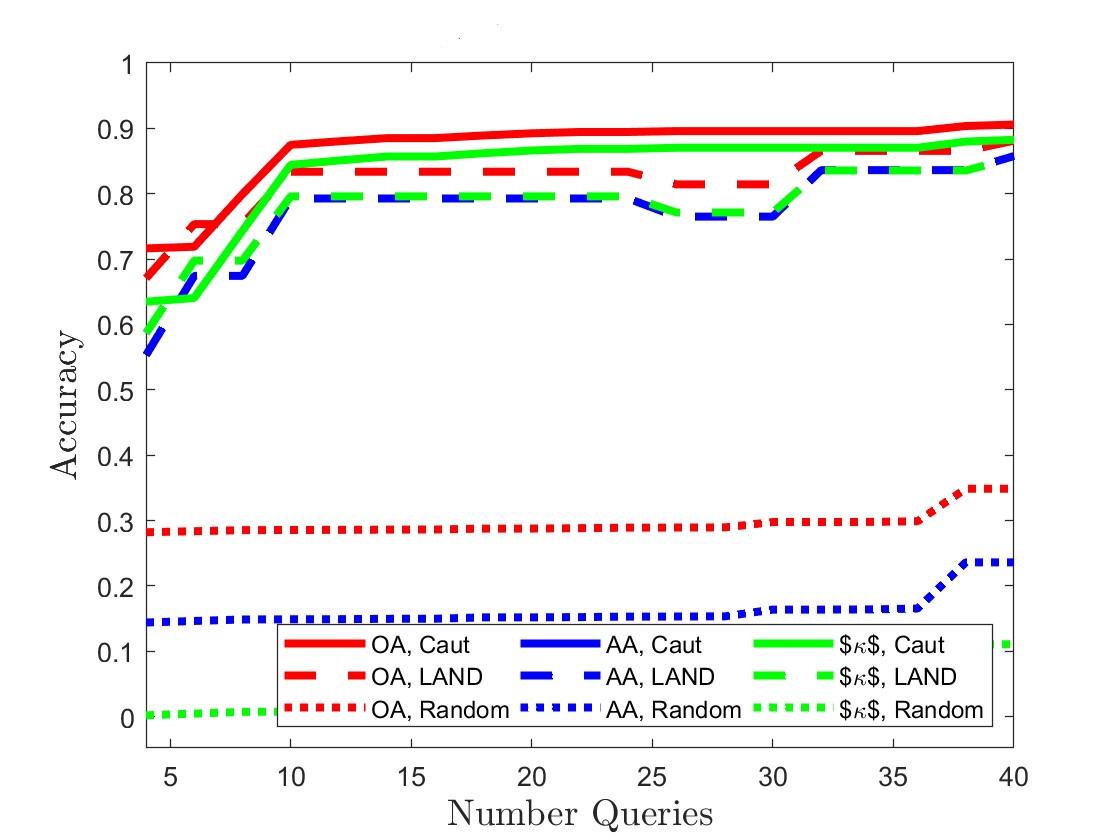} & 
\includegraphics[width=.3\textwidth]{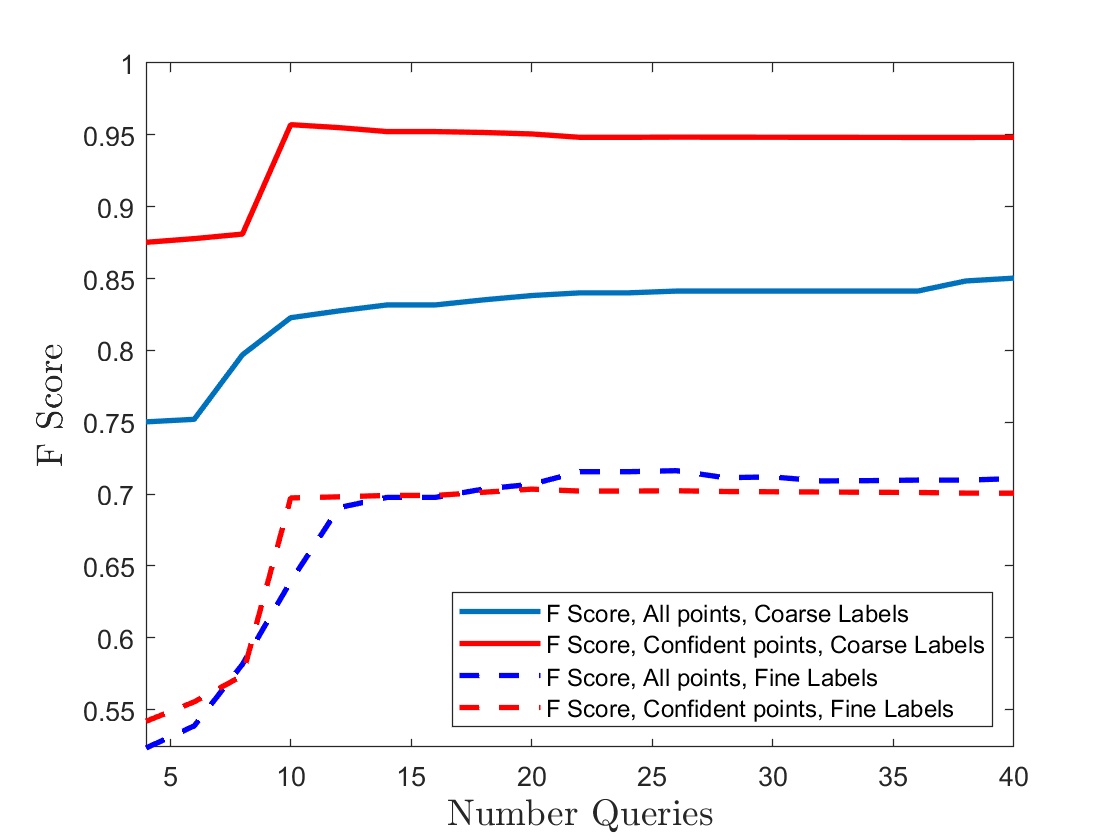}
\end{tabular}
\caption{(Left) Density estimate of 2-dimensional VAE latent space, colored by each MNIST label.  (Middle) Comparison of cautious clustering approach using coarse MNIST labels (merging overlapping clusters as described in Section \ref{mnistsect}).  (Right) $F$-score as a function of number of queries for cautious clustering approach for both coarse MNIST labels and for fine MNIST labels.}\label{fig:mnist_coarse}
\end{figure}

\subsection{Hyperspectral imagery}\label{hsi}
Hyperspectral imagery (HSI) is an imaging modality that captures radiation reflected from a surface across a number of different wavelengths (also called bands).  
This results in each pixel in an image being represented by its energy in $q$ different bands, where $q$ is sometimes in the hundreds, as the bands can range from long wavelength infrared ($\approx 2500nm$ wavelength) to ultraviolet ($\approx 400nm$ wavelength).
Each pixel can cover a significant area of the surface, usually several square meters.  Because of this, it is not always advantageous to use spatial similarity to aid in classification and clustering, since neighboring pixels could easily have different labels \cite{bachmann2005exploiting}.  
Instead, we will consider only the spectral similarity among the pixels   \cite{bachmann2005exploiting,cloninger2014operator}.

Hyperspectral pixels are difficult to collect labels for, as it requires physically inspecting the surface to determine its label.  Because of this, active learning has become very popular in the remote sensing community \cite{tuia2009active,murphy2018unsupervised}.  

Another issue with labeling pixels is that clusters are inherently hierarchical in nature.  For example, in agricultural settings, one not only has to distinguish stone from trees (large separation between classes), but also distinguish a particular crop after 4 weeks of growth from crops after 5 or 6 weeks of growth (small separation between classes).  Because of this, there exists a hierarchical relationship to the labels, and the level of specificity desired can change the number of clusters and queried labels  that are necessary.

Mathematically, each pixel can be thought of as being a data point in $\mathbb{R}^q$, so that an image with $M$ pixels  can be organized as   $q\times M$ matrix, where the $j$-th column represents the $q$-band spectral observation on the $j$-th pixel. 
For all of the examples below, we begin by taking the top principal components of this matrix resulting in a  choice of dimension that captures $80\%$ of the variance of the data in place of $q$.

We also note that for these experiments, we only plan to use spectral similarity between HSI pixels.  In certain HSI algorithms, the spatial relationship between the pixels is also incorporated. 
This could be done using $\Phi_n$ by considering data $\x = (\x_{spec}, \x_{spat})$ for spectral information $\x_{spec}$ and spatial information $\x_{spat}$, and taking a product kernel $\Phi_n(\x_{spec}, \y_{spec})\cdot \Phi_m(\x_{spat},\y_{spat})$.  Such spectral-spatial product kernels have been considered on HSI \cite{cloninger2014operator, benedetto2016spatial}, and spatial-spectral kernels have been considered in the context of HSI active learning \cite{murphy2020spatially,murphy2019spectral}.

\begin{figure}[!h]
\centering
\begin{tabular}{ccc}
\includegraphics[width=.3\textwidth]{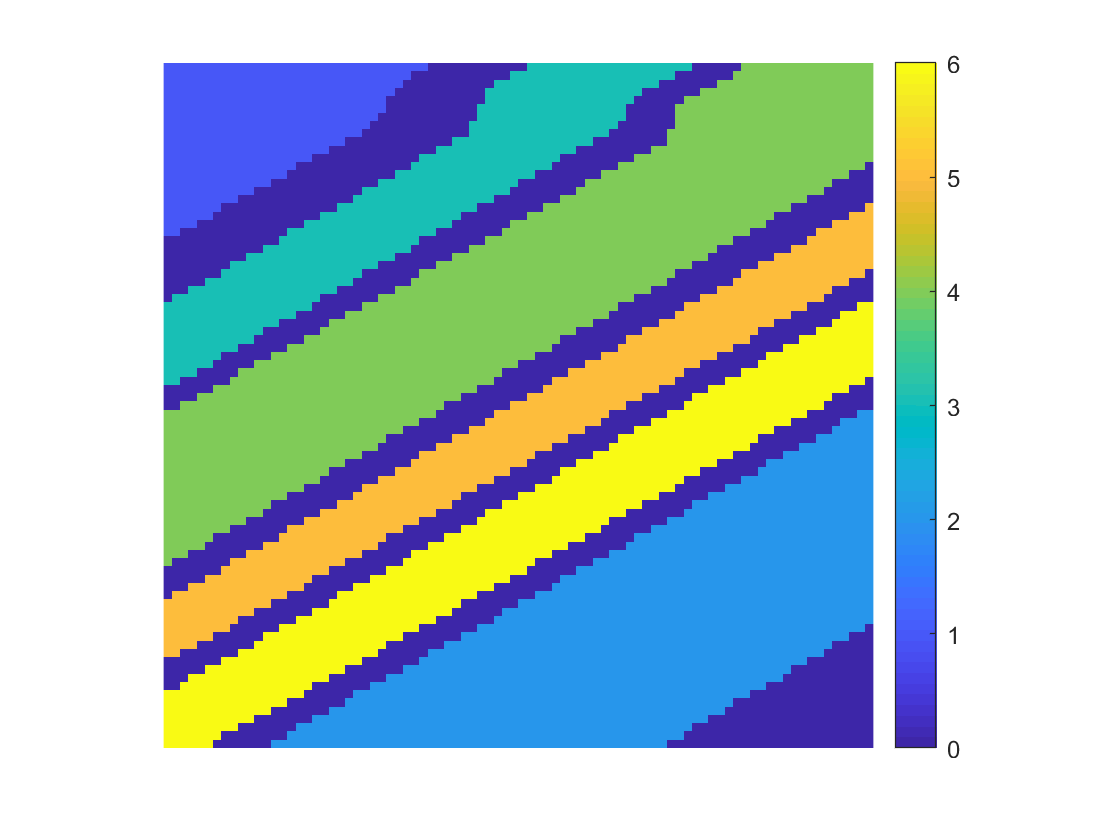} & 
\includegraphics[width=.3\textwidth]{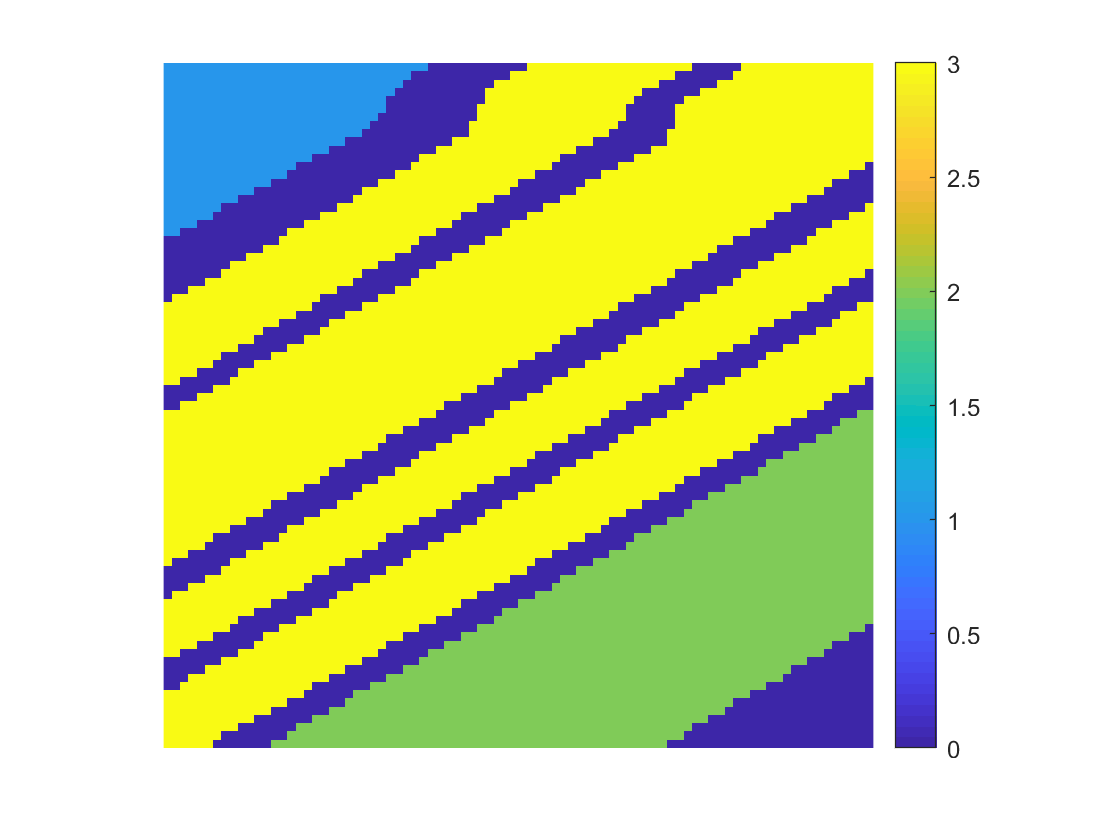} & 
\includegraphics[width=.3\textwidth]{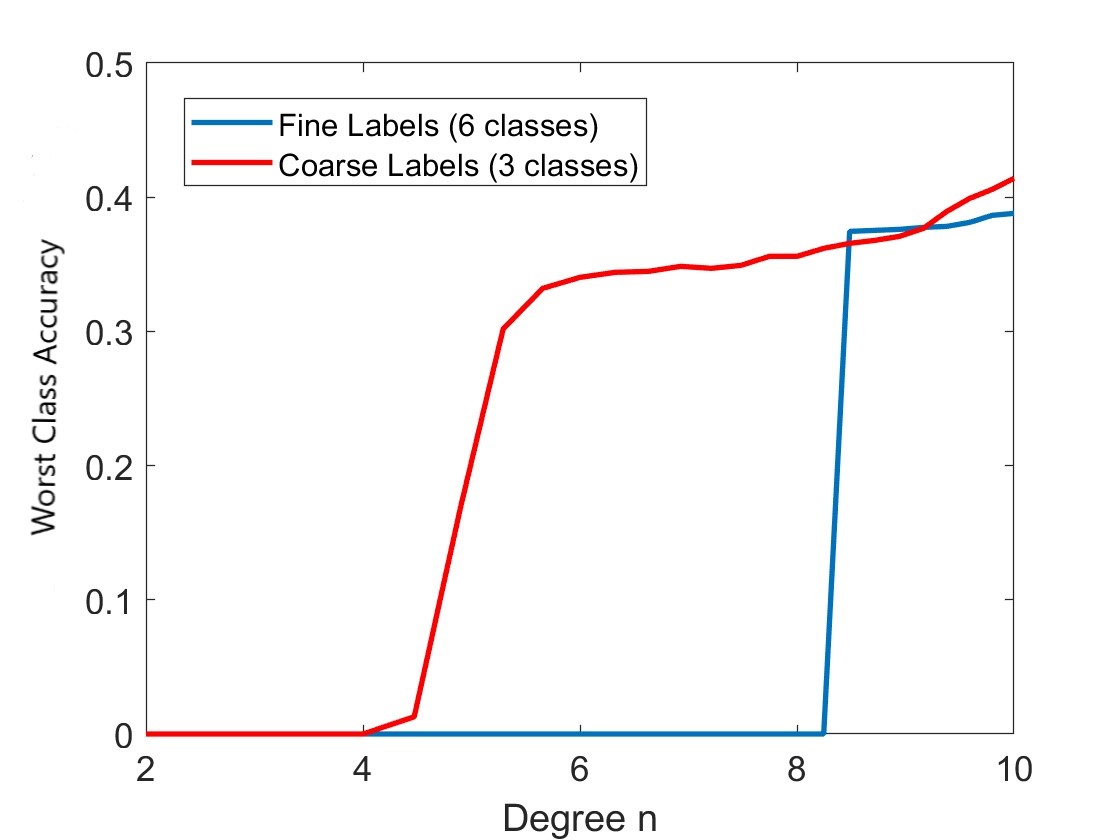} 
\end{tabular}
\includegraphics[width=.8\textwidth,height=.2\textwidth]{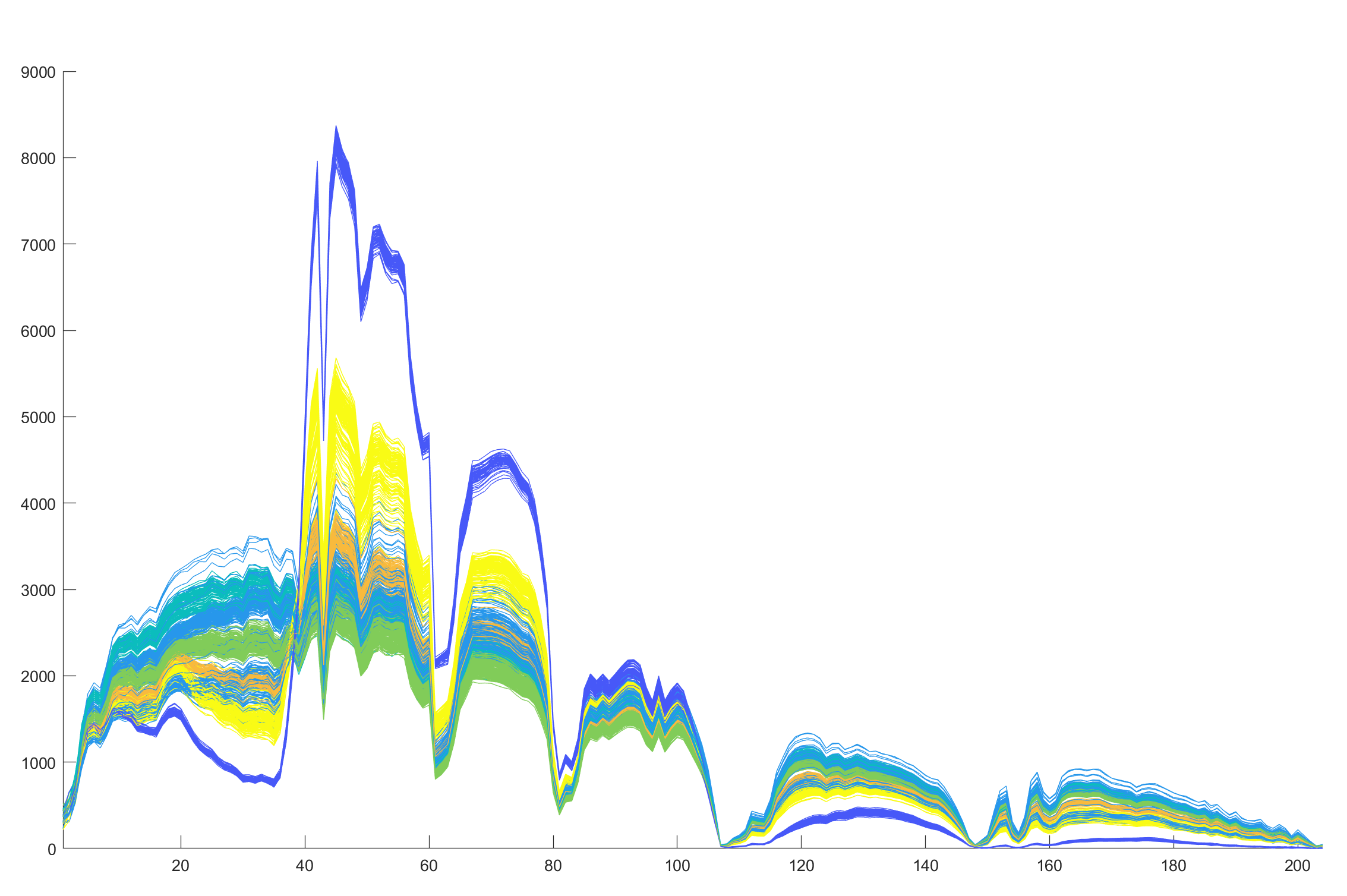}
\caption{Salinas-A HSI data.  (Top Left) Ground Truth of pixels with color corresponding to one of six labels (fine labels).  (Top middle) Same scene with all lettuce labels placed in single class (coarse labels).  (Top right) Worst classification accuracy on a class using our algorithm as a function of $n$.  Note that we clearly separate all classes much faster under coarse labeling.  (Bottom) Examples of pixels in $\mathbb{R}^{204}$, colored by label, the $x$-axis denotes the band number and the $y$-axis denotes the corresponding spectral radiance in that band.}\label{fig:salinasSmallLabels}
\end{figure}

We demonstrate this hierarchical relationship in an example using the Salinas-A data set of hyperspectral imagery \footnote{\url{http://www.ehu.eus/ccwintco/index.php/Hyperspectral_Remote_Sensing_Scenes}}.  Salinas-A is an $83\times 86$ image of three different agricultural crops being grown (broccoli, corn, lettuce). 
 Beyond this, there is a sub-classification of which week of growth the lettuce is in (4 weeks, 5 weeks, 6 weeks, 7 weeks).  
 Each pixel collects 204 bands, and we initially reduce the dimension to 10 using PCA. 
Thus, there are $7138$ points in $\RR^{204}$, which are projected to $7138$ points in $\RR^{10}$. 
 We run our cautious clustering algorithm on this data in two settings, and display summary results in Figure \ref{fig:salinasSmallLabels}.  We plot the classification accuracy on the worst class as a function of $n$.  We can see that running our cautious hierarchical clustering algorithm on the coarse labeling (broccoli, corn, lettuce) begins to yield correct classification on all classes for much smaller degree $n$ than for the same data with fine clustering (broccoli, corn, lettuce4, lettuce5, lettuce6,lettuce7).    This establishes that the effective minimal separation between all clusters is not constant, but varies depending on the desired level of specificity for the labels.

\begin{figure}[!h]
\centering
\begin{tabular}{cc}
\includegraphics[width=.3\textwidth]{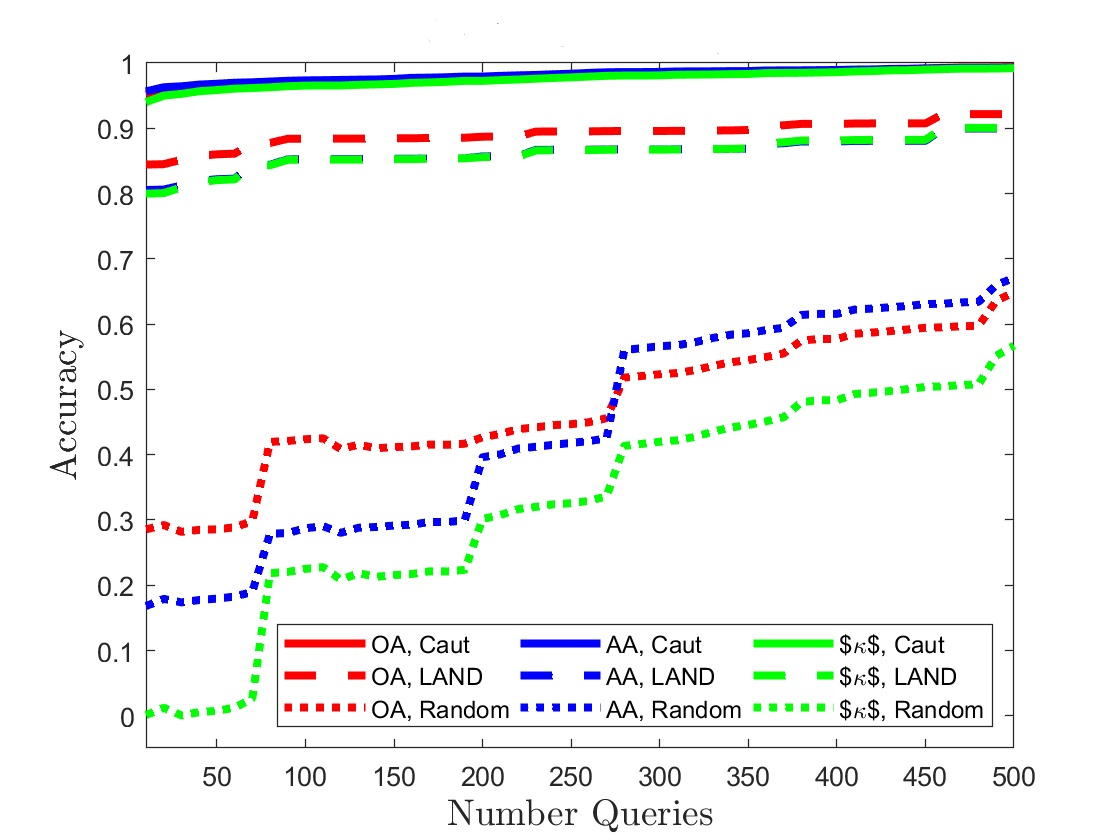} & 
\includegraphics[width=.3\textwidth]{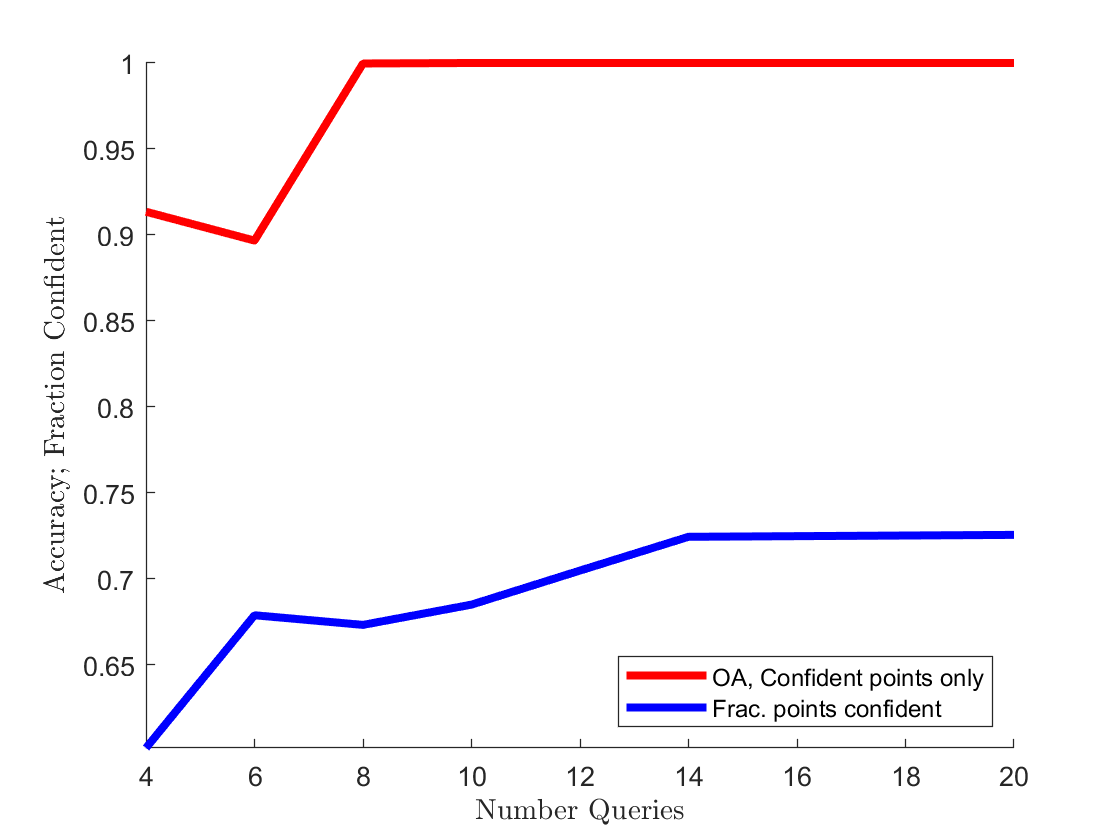} \\
\includegraphics[width=.3\textwidth]{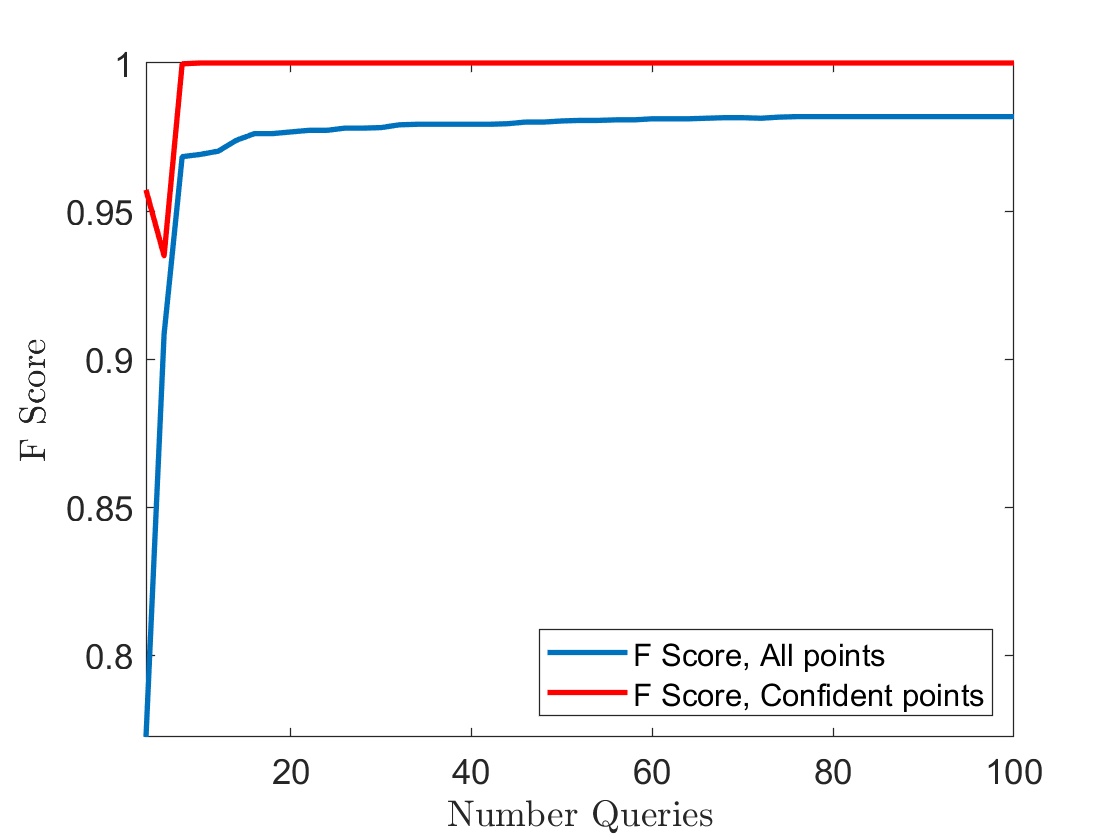}   & 
\includegraphics[width=.3\textwidth]{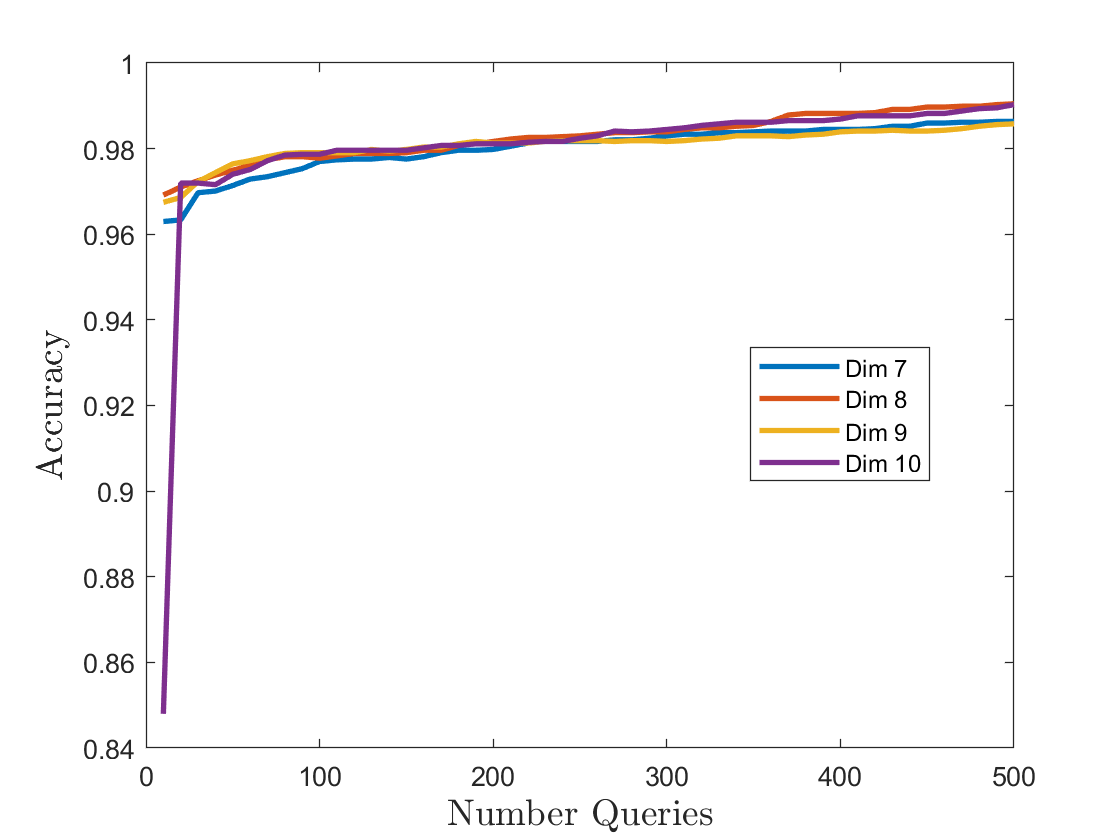}   
\end{tabular}
\caption{Salinas-A HSI data. (Top Left) Comparison of our algorithm of cautious active clustering to LAND and random sampling of labels.  (Top Right) Classification accuracy on $\mathcal{G}_{nmax}(\Theta,\C)$ only, and fraction of points in $\mathcal{G}_{nmax}(\Theta,\C)$. (Bottom Left) F score for $\mathcal{G}_{nmax}(\Theta,\C)$, and for all points $\C$.  (Bottom Right) The classification accuracy of our algorithm when varying the dimension of the PCA pre-processing.}\label{fig:salinasCompare}
\end{figure}

The main purpose of this data set is to examine the HSI active learning problem, and determine the maximum classification accuracy given a budget of only sampling $k$ labels.  We wish to emphasize that our algorithm returns an additional advantage over most active learning algorithms, namely the set $\mathcal{G}_{nmax}(\Theta,\C)$ on which we are confident in our classification.  This means we can attain near perfect accuracy on these points, as well as use them to estimate the class on $\C\setminus \mathcal{G}_{nmax}(\Theta,\C)$ using the witness function.
  We display the accuracy on $\mathcal{G}_{nmax}(\Theta,\C)$ in Figure \ref{fig:salinasCompare}, as well as the classification accuracy on the full data set after propagating labels to $\C\setminus \mathcal{G}_{nmax}(\Theta,\C)$ with the witness function. We note that there is a slight dip in the accuracy and $F$-score for approximately 6 queries.  This is due to the fact that the newest labels added begin to repeat classes (multiple samples from one label) prior to adding a first sample of the final class of points missing.  Similarly, when the minimal separation is still large (small $n$), it is possible that one cluster still contains multiple classes and thus has an artificially low classification accuracy that rebounds as $n$ increases.  Finally, in Figure \ref{fig:salinasCompare} we display the classification accuracy with a varied PCA dimension for pre-processing the data.  Reducing the dimension corresponds to removing in additional directions of variance from the HSI pixels, some of which may contain signal in determining a separation between clusters.  Despite this, our algorithm still performs well in this reduced dimension.

\begin{figure}[!h]
\centering
\begin{tabular}{ccc}
\includegraphics[width=.3\textwidth]{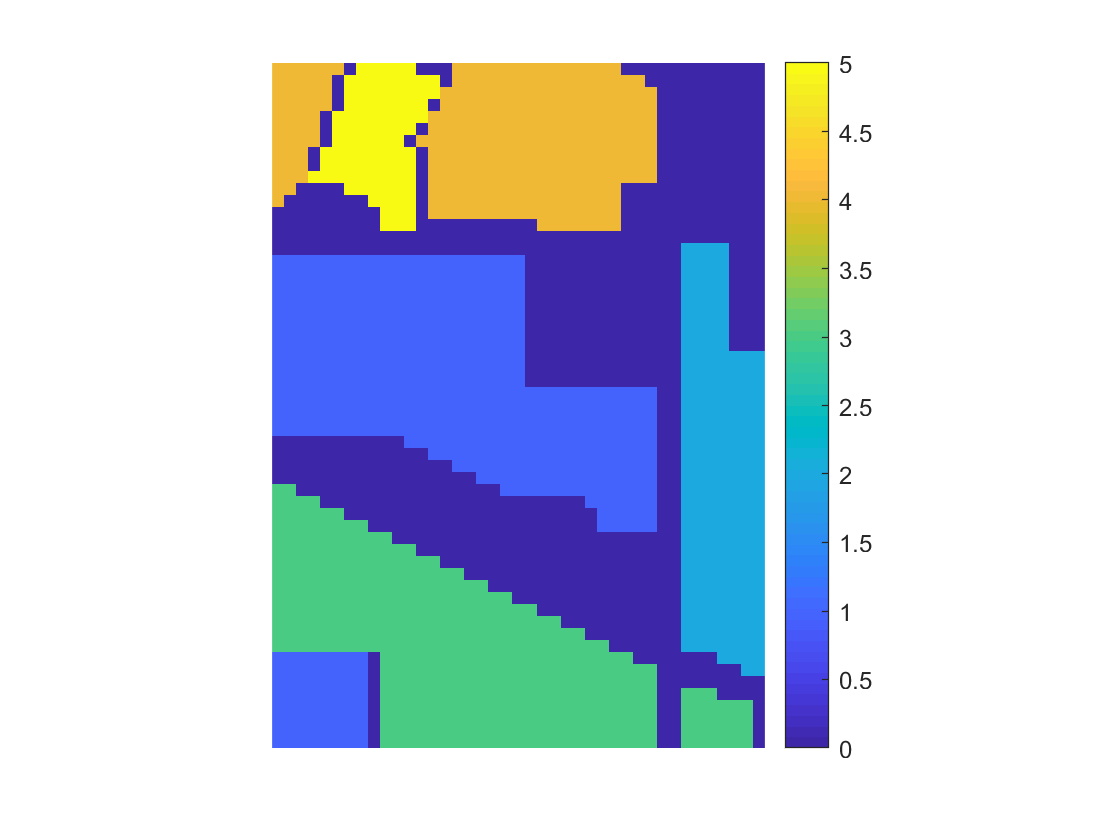} & 
\includegraphics[width=.3\textwidth]{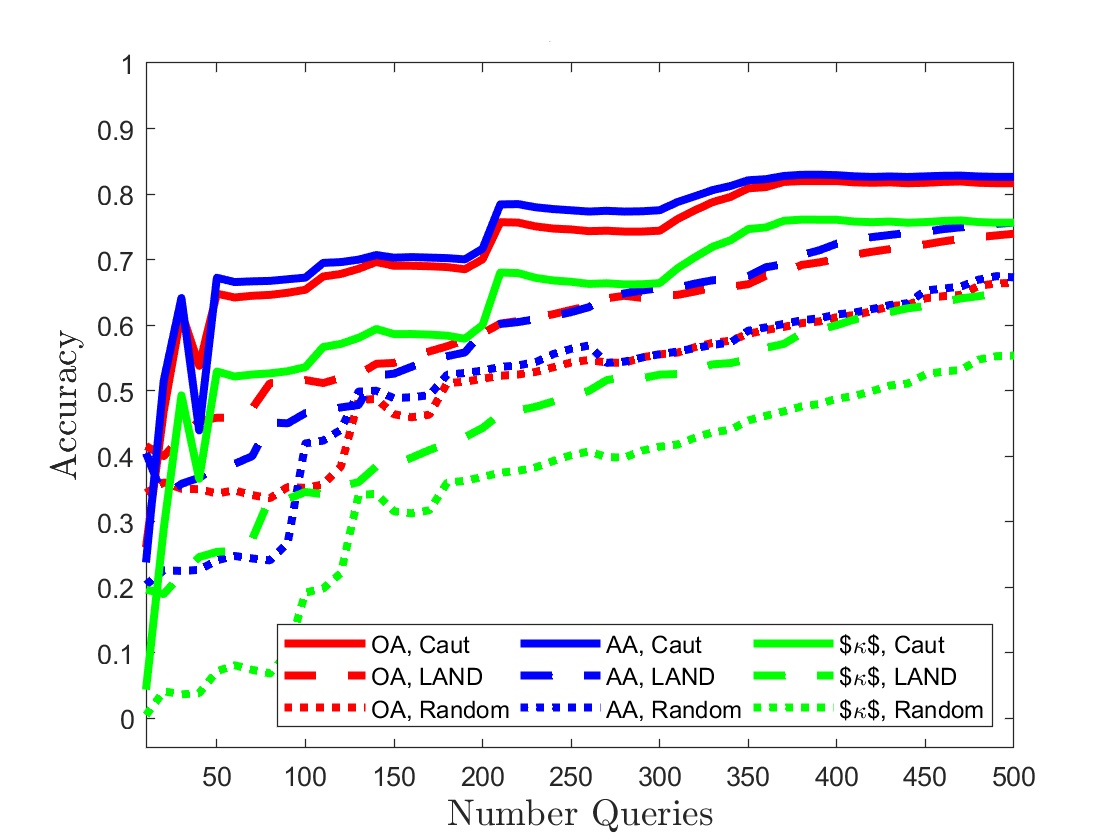} &
\includegraphics[width=.3\textwidth]{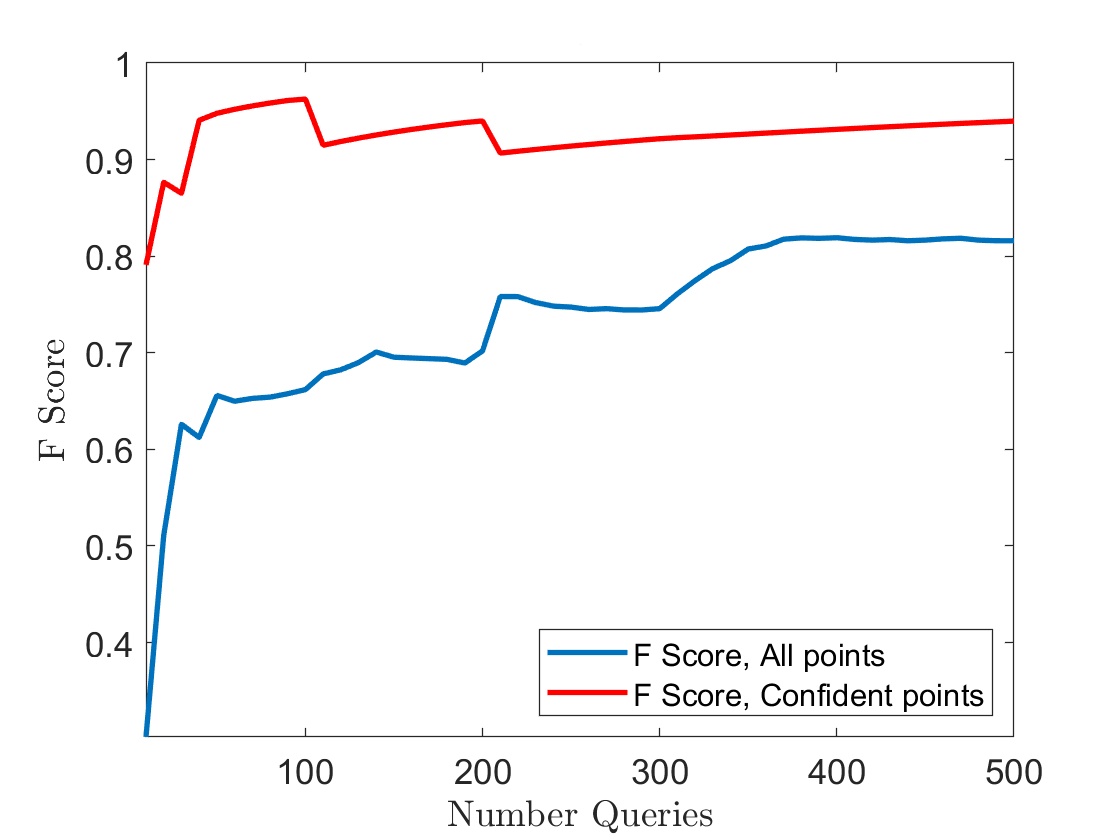} 
\end{tabular}
\caption{ Indian Pines HSI (Left) Ground truth of the small segment of Indian Pines image. (Center) Comparison of our cautious active clustering algorithm to LAND and random sampling of labels. (Right) F score for $\mathcal{G}_{nmax}(\Theta,\C)$ and for all points $\C$.}\label{fig:indianPinesCompare}
\end{figure}

We also examine a second data set, which is a $57\times 41$ subset of the Indian Pines hyperspectral data set \footnote{\url{http://www.ehu.eus/ccwintco/index.php/Hyperspectral_Remote_Sensing_Scenes}}.  Each pixel has 220 features, and we initially reduce the dimension to 20 using PCA.
The subset we focus on contains three general materials; tilled corn, stone-steel, and soybeans. 
 Furthermore, soybeans are subdivided into tilled, no till, and clean sublabels.  This leads to five labels at the finest level of label resolution.  We compare the active learning classification accuracy for our algorithm in Figure \ref{fig:indianPinesCompare}.  While this is clearly a more difficult data set than Salinas-A as evidenced by the lower classification accuracy as a function of the number of labels queried, our algorithm still compares favorably to LAND.

\bhag{Proofs}\label{bhag:proofs}
In this section, we prove all the theorems in Section~\ref{bhag:main}. 
The required background is given in Section~\ref{bhag:backsect}.
Section~\ref{bhag:prepsect} develops some preparatory results.
In Section~\ref{bhag:mainpfsect}, we prove first the deterministic analogues of Theorems~\ref{theo:main_single} and Theorems~\ref{theo:main_multi} in Theorems~\ref{theo:basic_single} and Theorem~\ref{theo:basic_multi} respectively, and then complete the proofs of the theorems presented in Section~\ref{bhag:main}.

\subsection{Background}\label{bhag:backsect}
We recall various properties of the kernel $\Phi_n$ defined in \eqref{summkerndef} (cf. \cite{hermite_recovery}). 
\begin{prop}\label{prop:kernel}
There exist $\kappa, \kappa_1, \cdots, \kappa_4>0$ depending only on $q$ and $H$ such that\\
\be\label{kern_magnitude}
\kappa_1n^{2q}\le \Phi_n(\x,\x)^2\le \kappa_2 n^{2q}, \qquad |\x|_\infty, |\y|_\infty \le \kappa n,
\ee
\be\label{kern_near_diag}
|\Phi_n(\x,\y)^2-\Phi_n(\x,\x)^2|<(1/2)\Phi_n(\x,\x)^2, \qquad |\x-\y|_\infty \le \kappa_3/n, \ |\x|_\infty \le \kappa n,
\ee
and
\be\label{phin_loc1}
|\Phi_n(\x,\y)|^2 \le \frac{\kappa_4 n^{2q}}{\max(1, (n|\x-\y|_\infty)^S)}, \qquad \x,\y\in\RR^q.
\ee
\end{prop}

For $n>0$ (not necessarily an integer), let $\Pi_{n,a}^q=\{\x\mapsto P(\x)\exp(-a|\x|_2^2) : P \mbox{ polynomial of total degree } <n^2\}$. 
We will omit the mention of $a$ when $a=1$.
Members of $\Pi_n^q$ will be called ($q$-variate) weighted polynomials.
The symbol $\|\cdot\|$ will denote the supremum norm on the space $C_0(\RR^q)$.
The following proposition states two important facts about weighted polynomials, obtained by applying corresponding univariate results in  \cite{mhasbk, mohapatrapap} one variable at a time to the multi-variate case. 
\begin{prop}\label{prop:wtpoly}
Let $n \ge 1$, $P\in\Pi_{n,a}^q$. \\
{\rm (a)} (\textbf{MRS identity}) We have
\be\label{mrsidentity}
\|P\|=\max_{\x\in [-n/a, n/a]^q} |P(\x)|.
\ee
{\rm (b)} (\textbf{Bernsetin inequality)}
There is a positive constant $\kappa_5$ depending only on $q$ such that
\be\label{bernineq}
\||\nabla P|\| \le \kappa_5 \frac{n}{a}\|P\|.
\ee
\end{prop}
The following corollary is easy to prove:
\begin{cor}\label{cor:covering}
Let $n>0$, $\C\subset [-n/a, n/a]^q$ be a finite set satisfying
\be\label{coveringmeshnorm}
\max_{\x\in [-n/a, n/a]^q}\min_{\y\in\C}|\x-\y|_\infty \le a/(2\kappa_5 n).
\ee
Then for any $P\in\Pi_{n,a}^q$,
\be\label{finitemaxnorm}
\max_{\y\in\C}|P(\y)| \le \|P\| \le 2\max_{\y\in\C}|P(\y)|.
\ee
There exists a set $\C$ as above with $|\C|\sim n^{2q}$.
\end{cor}

We will need the following facts from probability theory.
Theorem~\ref{theo:supnorm} is proved as \cite[Theorem~6.1]{witnesspaper}. 
\begin{theorem}\label{theo:supnorm}
Let $\XX$ be a topological space, $W$ be a linear subspace of $C_0(\XX)$. We assume that there is a finite  set $\C$ (norming set) satisfying
\be\label{normingset}
\sup_{x\in\XX}|f(x)|\le \mathfrak{n}(W,\C)\sup_{y\in\C}|f(y)|, \qquad f\in W.
\ee
Let $(\Omega, \mathcal{B}, \mu)$ be a probability space, and $Z : \Omega\to W$. We assume further that for any $x\in\XX$, $\omega\in\Omega$, $|Z(\omega)(x)|\le R$ for some $R>0$. Then for any $\delta>0$, integer $M\ge 1$, and independent sample $\omega_1,\cdots, \omega_M$, we have
\be\label{genprobest}
\mathsf{Prob}_\mu\left(\sup_{x\in\XX}\left|\frac{1}{M}\sum_{j=1}^M Z(\omega_j)(x)-\mathbb{E}_\mu(Z(\circ)(x))\right|\ge 4\mathfrak{n}(W,\C)R\sqrt{\frac{\log(2|\C|/\delta)}{M}}\right) \le \delta.
\ee
\end{theorem}
The following proposition summarizes the multiplicative Chernoff bounds in the form we need them (cf., e.g., \cite[Eqn~(7)]{hagerup1990guided} for an elementary proof).

\begin{prop}\label{prop:chernoff}
Let $M\ge 1$, $0\le p\le 1$, and $X_1,\cdots, X_M$ be random variables taking values in $\{0,1\}$, with $\mathsf{Prob}(X_k=1)=p$. Then for $\epsilon \in (0,1]$,
\be\label{chernoffbd}
\mathsf{Prob}\left(\sum_{k=1}^M X_k \le (1-\epsilon)Mp\right) \le \exp(-\epsilon^2 Mp/2).
\ee
\end{prop}

\subsection{Preparatory results}\label{bhag:prepsect}
In this section and the next, we will assume that $\mu^*$ is a detectable measure with parameters as described in Definition~\ref{def:mustardef}.
We denote
\be\label{supintdef}
I_n=\sup_{\z\in\supp(\mu^*)}\int_{\RR^q} \Phi_n(\z,\y)^2d\mu^*(\y).
\ee
\begin{lemma}\label{lemma:basic}
Let $d >0$, $n\ge 1$, $\x\in\RR^q$, and$\supp(\mu^*)\subseteq \BB(\bs 0, \kappa n)$. Then there exist $C_3, C_4$ such that
\be\label{awayest}
\int_{\RR^q\setminus \BB(\x,d)}\Phi_n(\x,\y)^2d\mu^*(\x) \le  C_3n^{2q-\alpha}\min\left(1, (nd)^{\alpha-S}\right),
\ee
and
\be\label{integralest}
n^{2q-\alpha}/C_4 \le \inf_{\x\in \supp(\mu^*)}\int_{\RR^q}\Phi_n(\x,\y)^2d\mu^*(\y)\le  \max_{\x\in\RR^q}\int_{\RR^q}\Phi_n(\x,\y)^2d\mu^*(\y) \le C_3n^{2q-\alpha}.
\ee
In particular, 
\be\label{full_vs_partial_support}
I_n=\sup_{\z\in\supp(\mu^*)}\int_{\RR^q} \Phi_n(\z,\y)^2d\mu^*(\y)\sim \sup_{\z\in\RR^q}\int_{\RR^q} \Phi_n(\z,\y)^2d\mu^*(\y) \sim n^{2q-\alpha}.
\ee
\end{lemma}

\begin{Proof}\ 
First, let $d\ge 1/n$, and for $k\in\ZZ_+$, $A_k=\{\y : 2^kd <|\x-\y|_\infty\le 2^{k+1}d\}$. Then  \eref{ballmeasurecond} shows that $\mu^*(A_k)\le 2^\alpha C_2 (2^k d)^\alpha$. Hence, \eref{phin_loc1} leads to
\bea\label{pf1eqn1}
\int_{\RR^q\setminus \BB(\x,d)}\Phi_n(\x,\y)^2d\mu^*(\y) &\le& \kappa_4 n^{2q-S}\int_{\RR^q\setminus \BB(\x,d)}\frac{d\mu^*(\y)}{|\x-\y|_\infty^S}=\kappa_4 n^{2q-S}\sum_{k=0}^\infty 
 \int_{A_k}\frac{d\mu^*(\y)}{|\x-\y|_\infty^S}\nonumber\\
&\le& \kappa_4 n^{2q-S}d^{-S}\sum_{k=0}^\infty 2^{-kS}\mu^*(A_k) \le 2^\alpha C_2\kappa_4 n^{2q-\alpha}(nd)^{\alpha-S}\sum_{k=0}^\infty 2^{k(\alpha-S)}
\nonumber\\
&=& \frac{2^\alpha}{1-2^{S-\alpha}} C_2\kappa_4 n^{2q-\alpha}(nd)^{\alpha-S}.
\eea
Using this estimate with $d=1/n$, and using \eref{phin_loc1} and \eref{ballmeasurecond} again, we obtain that 
$$
\int_{\RR^q} \Phi_n(\x,\y)^2d\mu^*(\y) =\int_{\BB(\x,1/n)}\Phi_n(\x,\y)^2d\mu^*(\y) +
\int_{\RR^q\setminus \BB(\x,d)}\Phi_n(\x,\y)^2d\mu^*(\y)\le C_2\kappa_4n^{2q-\alpha} + \frac{2^\alpha}{1-2^{S-\alpha}} C_2\kappa_4 n^{2q-\alpha}.
$$
This shows both the third inequality in \eref{integralest}, and  together with \eref{pf1eqn1}, also \eref{awayest} (with the same $C_3$).

Let $\x_0\in \supp(\mu^*)$. Then $\mu^*(\BB(\x_0,\kappa_3/n))\ge C_1\kappa_3^\alpha n^{-\alpha}$. In view of \eref{kern_magnitude} and \eref{kern_near_diag}, we have $\Phi_n(\x_0,\y)^2 \ge (\kappa_1/2)n^{2q}$ for all $\y\in \BB(\x_0,\kappa_3/n)$. Therefore,
$$
 \int_{\BB(\x_0,\kappa_3/n)}\Phi_n(\x_0,\y)^2d\mu^*(\y)\ge (C_1\kappa_1/2)\kappa_3^\alpha n^{2q-\alpha}.
$$
This leads to the first inequality in \eref{integralest}.
\end{Proof}

\begin{lemma}\label{lemma:hoeffding}
Let  $n\ge 1$ be large enough so that $\supp(\mu^*)\subseteq \BB(\bs 0, \kappa n)$, $\{\x_j\}_{j=1}^M$ be independent samples with $\mu^*$ as the probability distribution. 
Then 
\be\label{hoeffdingest}
\mathsf{Prob}\left(\sup_{\x\in\RR^q}\left|\frac{1}{M}\sum_{j=1}^M \Phi_n(\x,\x_j)^2 -\int_{\RR^q}\Phi_n(\x,\y)^2d\mu^*(\y)\right| \ge cn^{\alpha}\sqrt{\frac{\log n}{M}}I_n\right) \le \frac{1}{2n}.
\ee
In particular, if $\beta>0$, and with $c$ as in \eref{hoeffdingest},
\be\label{Mcond1}
M\ge (c^2/\beta^2)n^{2\alpha}\log n,
\ee
then with probability $\ge 1-1/(2n)$, for $\x\in\RR^q$,
\be\label{phitransbd}
\int_{\RR^q}\Phi_n(\x,\y)^2d\mu^*(\y)-\beta I_n \le \frac{1}{M}\sum_{j=1}^M \Phi_n(\x,\x_j)^2 \le \int_{\RR^q}\Phi_n(\x,\y)^2d\mu^*(\y)+\beta I_n,
\ee
\be\label{phitranssupbd}
(1-\beta)I_n \le \max_{\x\in\RR^q}\frac{1}{M}\sum_{j=1}^M \Phi_n(\x,\x_j)^2 \le (1+\beta)I_n.
\ee
\end{lemma}
\begin{Proof}\ 
We use Theorem~\ref{theo:supnorm} with the following choices: $\mu^*$ in place of $\mu$, $\x_j$ in place of $\omega_j$, $\delta=1/(2n)$, $Z(\circ)(\x)=\Phi_n(\x,\circ)^2$ (so that $W=\Pi_{2n}^q$, and with $\C$ as in Corollary~\ref{cor:covering}, $\mathfrak{n}(W,\C)=2$, $|\C|\sim n^{2q}$). 
This yields
$$
\mathsf{Prob}\left(\sup_{\x\in\RR^q}\left|\frac{1}{M}\sum_{j=1}^M \Phi_n(\x,\x_j)^2 -\int_{\RR^q}\Phi_n(\x,\y)^2d\mu^*(\y)\right| \ge cn^{2q}\sqrt{\frac{\log n}{M}}\right) \le \frac{1}{2n}.
$$
The proof is completed using \eref{integralest}.
\end{Proof}

\begin{lemma}\label{lemma:samplesup}
There exist $C^*, c_1, c_2>0$ with the following property. Let $n\ge c_1$, $0<\beta<1$, $\supp(\mu^*)\subseteq \BB(\bs 0, \kappa n)$, $M\ge c_2\beta^{-2}n^{2\alpha}\log n$. 
If  $\{\x_j\}_{j=1}^M$ is a random sample with $\mu^*$ as the probability distribution, then
\be\label{samplesupest}
\mathsf{Prob}\left(\max_{1\le k\le M}\frac{1}{M}\sum_{j=1}^M\Phi_n(\x_k,\x_j)^2 \le C^* I_n\right)\le 1/(2n);
\ee
i.e., with probability $\ge 1-1/n$, (cf. \eref{phitranssupbd})
\be\label{samplesupest1}
C^* I_n\le \max_{1\le k\le M}\frac{1}{M}\sum_{j=1}^M\Phi_n(\x_k,\x_j)^2 \le (1+\beta)I_n.
\ee
\end{lemma}
\begin{Proof}\ 
In this proof, let $P\in\Pi_{2n}^q$ be defined by
$$
P(\x)=\int_{\RR^q}\Phi_n(\x,\y)^2d\mu^*(\y).
$$
Let $n$ be large enough so that $\supp(\mu^*)\subset \BB(\bs 0,\kappa n)$.
Then \eref{integralest} shows that
\be\label{pf4eqn1}
\sup_{\x\in \supp(\mu^*)}|P(\x)|=P(\x^*)\ge cI_n
\ee
for some $\x^*\in \supp(\mu^*)$. 
Therefore, using the Bernstein inequality \eref{bernineq}, we obtain for $\x\in\RR^q$,
$$
|P(\x^*)-P(\x)|\le cn|\x^*-\x|_\infty I_n \le cn|\x^*-\x|_\infty\sup_{\x\in \supp(\mu^*)}|P(\x)|=cn|\x^*-\x|_\infty P(\x^*);
$$
i.e.,
\be\label{pf4eqn2}
P(\x)\ge \left(1-cn|\x^*-\x|_\infty\right)P(\x^*)\ge c_3\left(1-cn|\x^*-\x|_\infty\right)I_n, \qquad \x\in\BB(\x^*,(cn)^{-1}).
\ee
Now we consider the following random variables: for $k=1,\cdots, M$, we take $X_k=1$ if $\x_k\in \BB(\x^*, \kappa_3/n^2)$, and $0$ otherwise, so that the probability $p$ that $X_k=1$ is given by $p=\mu^*( \BB(\x^*, \kappa_3/n^2))\ge cn^{-2\alpha}$.  We then use multiplicative Chernoff bound \eref{chernoffbd}  with  $\epsilon=1$ to obtain for $M\ge cn^{2\alpha}\log n$,
$$
\mathsf{Prob}\left(\sum_{k=1}^M X_k \le 0\right)\le \exp(-Mp/2) \le 1/(2n).
$$
Thus, with probability exceeding $1-1/(2n)$, there exists $\x_\ell\in \BB(\x^*, \kappa_3/n^2)$. 
Together with \eref{pf4eqn2} this shows that with probability exceeding $1-1/(2n)$,
\be\label{pf4eqn3}
\max_{1\le k\le M}P(\x_k)\ge P(\x_\ell)\ge 2C^*I_n.
\ee
Using the first estimate in \eqref{phitransbd} with $C^*/2$ in place of $\beta$, we see that if $M\ge cn^{2\alpha}\log n$, then with $C^*$ as in \eref{pf4eqn3}, and probability exceeding $1-1/(2n)$,
$$
\max_{1\le k\le M}\frac{1}{M}\sum_{j=1}^M\Phi_n(\x_k,\x_j)^2 \ge \max_{1\le k\le M}P(\x_k)-(C^*/2)I_n \ge C^*I_n.
$$
This proves the first inequality in \eqref{samplesupest1}. The second inequality is immediate from the second inequality in \eqref{phitranssupbd}.
\end{Proof}

\subsection{Proofs of the main theorems}\label{bhag:mainpfsect}

We first state and prove some theorems in the non-noisy case.
\begin{theorem}\label{theo:basic_single}
Let $\mu^*$ be detectable, $S>\alpha$,  $\theta>0$,  and for $n\ge 1$,
\be\label{pre_threshold_set}
\mathcal{S}=\mathcal{S}_n(\theta) =\left\{\int_{\RR^q}\Phi_n(\x,\y)^2d\mu^*(y) \ge 4\theta \sup_{\z\in\supp(\mu^*)}\int_{\RR^q} \Phi_n(\z,\y)^2d\mu^*(y)\right\}.
\ee
We assume \eref{ncond1} and
\be\label{thetacond1}
0< \theta\le \min\left( (4C_3C_4)^{-1}, C_3C_4\right).
\ee
Then with
\be\label{dthetadef}
d(\theta)= \left(\frac{C_3C_4}{\theta}\right)^{1/(S-\alpha)}
\ee
\be\label{singlemeasure_det}
\supp(\mu^*)\subseteq\mathcal{S} \subseteq \left\{\x\in \RR^q : \mathsf{dist} (\x,\supp(\mu^*))\le \frac{d(\theta)}{n}\right\}.
\ee
\end{theorem}

\begin{Proof}\ 
 
The estimate \eref{integralest} shows that for $\x\in \supp(\mu^*)$,
$$
\int_{\RR^q}\Phi_n(\x,\y)^2d\mu^*(y)\ge \frac{I_n}{C_3C_4}.
$$
Since $\theta\le (4C_3C_4)^{-1}$, this shows the first inclusion in \eref{singlemeasure_det}.

Let $\mathsf{dist}(\x, \supp(\mu^*)) \ge d(\theta)/n$. The condition \eref{thetacond1} shows that $d(\theta)\ge 1$. So,  \eref{awayest} leads to  
\bea\label{pf2eqn1}
\int_{\RR^q}\Phi_n(\x,\y)^2d\mu^*(y) &=& \int_{\supp(\mu^*)}\Phi_n(\x,\y)^2d\mu^*(y)\nonumber\\
& \le& \int_{\RR^q\setminus \BB(\x,d(\theta)/n)}\Phi_n(\x,\y)^2d\mu^*(\x) \le  C_3n^{q-2\alpha}d(\theta)^{\alpha-S}\nonumber\\
& \le& C_3C_4 d(\theta)^{\alpha-S} I_n = \theta I_n.
\eea
This proves the second inclusion in \eref{singlemeasure_det}.
\end{Proof}

The next theorem shows the detection of the supports $\mathbf{S}_{k,\eta}$ of the components $\mu_k$ of $\mu^*$.

\begin{theorem}\label{theo:basic_multi}
We assume the set-up as in Theorem~\ref{theo:basic_single}. 
In addition, we assume that $\mu^*$ has a fine structure, and that
\be\label{add_n_cond}
n\ge 2d(\theta)/\eta,\ \kappa_1C_4 n^\alpha \mu^*(\mathbf{S}_{K_\eta+1,\eta}) \le \theta.
\ee
Let
\be\label{pre_partitiondef}
\mathcal{S}_{k,\eta, n}(\theta)=\mathcal{S}_n(\theta) \cap \{\x\in\RR^q : \mathsf{dist}(\x, \mathbf{S}_{k,\eta})\le d(\theta)/n\}.
\ee
Then the set $\mathcal{S}_n(\theta)$
 is a disjoint union of sets $\mathcal{S}_{k,\eta,n}(\theta)$, $k=1,\cdots, K_\eta$ such that
\be\label{pre_separation}
\mathsf{dist}(\mathcal{S}_{k,\eta,n}(\theta), \mathcal{S}_{j,\eta, n}(\theta))\ge \eta, \qquad k\not=j, \ k, j=1,\cdots, K_\eta,
\ee
and for $k=1,\cdots, K_\eta$,
\be\label{pre_comp_det}
\supp(\mu^*)\cap  \{\x: \mathsf{dist}(\x,\mathbf{S}_{k,\eta}) \le d(\theta)/n\}\subseteq \mathcal{S}_{k,\eta,n}(\theta)  \subseteq \{\x\in \RR^q : \mathsf{dist} (\x,\mathbf{S}_{k,\eta})\le d(\theta)/n\}.
\ee
\end{theorem}

\begin{Proof}\ 

The minimal separation condition and the first condition in \eref{add_n_cond} implies that the sets $\mathcal{S}_{k,\eta,n}$ are disjoint, and in fact, satisfy \eref{pre_separation}.
Also, \eref{singlemeasure_det} implies the first inclusion in \eref{pre_comp_det}.

Let $\x\in\mathcal{S}_n(\theta)$. 
Then we deduce using \eref{kern_magnitude} \eref{integralest}, and \eref{add_n_cond}  that
\be\label{pf3eqn1}
\int_{\mathbf{S}_{K_\eta+1}} \Phi_n(\x,\y)^2d\mu^*(y) \le \kappa_1n^{2q}\mu^*(\mathbf{S}_{K_\eta+1})\le \kappa_1C_4 n^\alpha \mu^*(\mathbf{S}_{K_\eta+1}) I_n \le \theta I_n.
\ee
In this proof, we will denote
$$
\mathbf{S}=\bigcup_{k=1}^{K_\eta} \mathbf{S}_{k,\eta}.
$$
If $\mathsf{dist}(\x, \mathbf{S}) \ge d(\theta)/n$ then we obtain as in the proof of Theorem~\ref{theo:basic_single}, that
$$
\int_{\mathbf{S}} \Phi_n(\x,\y)^2d\mu^*(y) \le \int_{\RR^q\setminus \BB(\x,d(\theta)/n)}\Phi_n(\x,\y)^2d\mu^*(\x) \le  C_3n^{q-2\alpha}d(\theta)^{\alpha-S}\le  C_3C_4d(\theta)^{\alpha-S} I_n = \theta I_n.
$$
Together with \eref{pf3eqn1}, this implies that $\mathcal{S}_n(\theta)\subseteq \{\x : \mathsf{dist}(\x, \mathbf{S}) \le d(\theta)/n\}$. 
Since $d(\theta)/n \le \eta/2$, the minimal separation condition shows that for any $\x$ with $\mathsf{dist}(\x, \mathbf{S}) \le d(\theta)/n$, there exists a unique $k$, $k=1,\cdots, K$, such that $\mathsf{dist}(\x, \mathbf{S}_{k,\eta}) \le d(\theta)/n$. 
Thus, $\mathcal{S}_n(\theta)=\bigcup_{k=1}^K \mathcal{S}_{k,\eta,n}(\theta)$, and the second inclusion of \eref{pre_comp_det} is proved. 
\end{Proof}

The following lemma helps us to connect Theorems~\ref{theo:basic_single} and \ref{theo:basic_multi} with 
Theorems~\ref{theo:main_single} and \ref{theo:main_multi}.

\begin{lemma}\label{lemma:almost_there}
Let $0<\Theta\le 1$, $M, n \ge 2$ be integers, $M\ge 2$ and $\C=\{\x_1,\cdots,\x_M\}$ be independently sampled from the probability distribution $\mu^*$. 
Let $\mathcal{G}_n(\Theta,\C)$ be defined by \eref{gsetdef}.
There exist constants $c, c_1, c_2$ such that
if $M\ge cn^{2\alpha}\sqrt{\log n}$ then with probability $\ge 1-c_1M^{-c_2}$,
\be\label{funda_inclusion}
\mathcal{S}_n\left(\frac{(1+C^*)\Theta}{4}\right)\subseteq \mathcal{G}_n(\Theta,\C)\subseteq \mathcal{S}_n(C^*\Theta/8).
\ee
\end{lemma}

\begin{Proof}\ 
All statements below hold with probability $\ge 1-c_1M^{-c_2}$, although the values of $c_1, c_2$ might be different at different occurrences as usual.
In applying Lemma~\ref{lemma:hoeffding}, we use $\beta=C^*\Theta/2$, 
$$
t_1=\frac{2\Theta+C^*\Theta(1+\Theta)}{8}=\frac{(1+\beta)\Theta+\beta}{4}.
$$
Let $\x\in \mathcal{S}_n(t_1)$. Using \eref{phitransbd} and \eref{samplesupest1}, we deduce that
\begin{eqnarray*}
\frac{1}{M}\sum_{j=1}^M\Phi_n(\x,\x_j)^2&\ge & \int_{\RR^q}\Phi_n(\x,\y)^2d\mu^*(\y)-\beta I_n \ge 4t_1I_n-\beta I_n =(4t_1-\beta)I_n\\
&\ge& \frac{4t_1-\beta}{1+\beta}\max_{1\le k\le M}\frac{1}{M}\sum_{j=1}^M\Phi_n(\x_k,\x_j)^2 =\Theta \max_{1\le k\le M}\frac{1}{M}\sum_{j=1}^M\Phi_n(\x_k,\x_j)^2.
\end{eqnarray*} 
Thus,
$$
 \mathcal{S}_n\left(\frac{2\Theta+C^*\Theta(1+\Theta)}{8}\right)\subseteq \mathcal{G}_n(\Theta,\C).
 $$
 Since 
 $$
 (1+C^*)\Theta/2 \ge \frac{2\Theta+C^*\Theta(1+\Theta)}{8},
 $$
this proves the first inclusion in \eref{funda_inclusion}.

Next, let $\x\in \mathcal{G}_n(\Theta,\C)$. 
Using \eref{phitransbd} and \eref{samplesupest1}, we deduce that
\begin{eqnarray*}
\int_{\RR^q}\Phi_n(\x,\y)^2d\mu^*(\y) &\ge& 
\frac{1}{M}\sum_{j=1}^M\Phi_n(\x,\x_j)^2 -\beta I_n\ge \Theta \max_{1\le k\le M}\frac{1}{M}\sum_{j=1}^M\Phi_n(\x_k,\x_j)^2\\
&\ge& (C^*\Theta -\beta) I_n= 4(C^*\Theta/8)I_n.
\end{eqnarray*}
This proves the second inclusion in \eref{funda_inclusion}.
\end{Proof}

\vskip 3ex
\noindent\textsc{Proofs of Theorems~\ref{theo:main_single} and \ref{theo:main_multi}.}\\

Theorems~\ref{theo:main_single} and \ref{theo:main_multi} follow immediately from Theorems~\ref{theo:basic_single} and \ref{theo:basic_multi} respectively using Lemma~\ref{lemma:almost_there}.\qed

\vskip 3ex
\noindent\textsc{Proof of Theorem~\ref{theo:fidelity}.}\\

In this proof, we write $\mathcal{G}_{k,\eta,n}$ in place of $\mathcal{G}_{k,\eta,n}(\Theta, \C)$.
In view of \eqref{comp_det}, we have for $k=1,\cdots,K_\eta$,
$\supp(\mu^*)\cap \mathbf{S}_{k,\eta}\subset \mathcal{G}_{k,\eta,n}.$
Hence,
\be\label{pf5eqn2}
\mu^*(\mathcal{G}_{k,\eta,n})\ge \mu^*(\mathbf{S}_{k,\eta}), \qquad k=1,\cdots,K_\eta.
\ee
With $c_1$, $c_2$ as in Theorem~\ref{theo:main_multi}, let
$$
d_n=c_2/(n\Theta^{1/(S-\alpha)}).
$$ 
Then
The estimate \eqref{comp_det}  implies again that 
\begin{eqnarray*}
\lefteqn{\mu^*\left( \{\x\in \RR^q: \mathsf{dist}(\x,\mathbf{S}_{k,\eta})\le d_n\}\right)}\\
&=&\mu^*\left(\mathsf{supp}(\mu^*)\cap  \{\x\in \RR^q: \mathsf{dist}(\x,\mathbf{S}_{k,\eta})\le d_n\}\right)
= \mu^*(\mathbf{S}_{k,\eta}\cup \mathbf{S}_{K_\eta+1,\eta})\le \mu^*(\mathbf{S}_{k,\eta})+\mu^*(\mathbf{S}_{K_\eta+1,\eta}).
\end{eqnarray*}
Hence, the second inclusion in\eqref{comp_det} implies that for $k=1,\cdots, K_\eta$,
\be\label{pf5eqn3}
\mu^*(\mathcal{G}_{k,\eta,n}) \le \mu^*(\mathbf{S}_{k,\eta})+\mu^*(\mathbf{S}_{K_\eta+1,\eta}).
\ee
In view of \eref{pf5eqn2} and \eref{pf5eqn3}, for each $k=1,\cdots, K$,
\be\label{pf5eqn1}
F_\eta(\mathcal{G}_{k,\eta,n})\ge 2\frac{\mu^*(\mathbf{S}_{k,\eta})}{2\mu^*(\mathbf{S}_{k,\eta})+\mu^*(\mathbf{S}_{K_\eta+1,\eta})}\ge 1-\frac{\mu^*(\mathbf{S}_{K_\eta+1,\eta})}{2\mu^*(\mathbf{S}_{k,\eta})+\mu^*(\mathbf{S}_{K_\eta+1,\eta})}.
\ee
Since $\{\mathcal{G}_{k,\eta,n}\}$ is a partition of $\mathcal{G}_n(\Theta,\mathcal{C})\supseteq \supp(\mu^*)$,
$$
\sum_{k=1}^{K_\eta}\mu^*(\mathcal{G}_{k,\eta,n})=1.
$$
Therefore, \eqref{pf5eqn1} and \eqref{pf5eqn3} imply that
\be\label{pf5eqn4}
\sum_{k=1}^{K_\eta} \mu^*(\mathcal{G}_{k,\eta,n})F\eta(\mathcal{G}_{k,\eta, n}) \ge 1-\mu^*(\mathbf{S}_{K_\eta+1,\eta})\sum_{k=1}^{K_\eta} \frac{\mu^*(\mathcal{G}_{k,\eta,n})}{2\mu^*(\mathbf{S}_{k,\eta})+\mu^*(\mathbf{S}_{K_\eta+1,\eta})}.
\ee
Therefore,
$$
1\ge \mathcal{F}_\eta\left(\{\mathcal{G}_{k,\eta,n}\}_{k=1}^K\right)\ge 1- \frac{\mu^*(\mathbf{S}_{K_\eta+1,\eta})}{2\min_{1\le k\le K_\eta}\mu^*(\mathbf{S}_{k,\eta})}.
$$
In view of \eqref{min_measure_cond},
this completes the proof. \qed

\bhag{Conclusions}\label{bhag:conclusions}
We have introduced the concept that the machine learning problem of classification can be considered in a manner analogous to the problem in signal processing of separating point sources. We have pointed out the various similarities and differences which makes the problem much harder than that of separation of point sources, in particular, because of overlapping class boundaries. 
We have introduced a localized kernel based on Hermite polynomials, and demonstrated its use in separating supports of the components of the data distribution corresponding to different classes. 
We have constructed a multiscale in which the number of classes can be defined hierarchically, and shown that the $F$-score for our classification scheme converges to the optimal value of $1$. 
Having separated the supports of the components, one sample per component leads in theory to a complete classification based on a minimal number of label queries.
We have given an algorithm to determine which points in the data set should be queried for labels in an optimal  and reliable manner.
The algorithm is demonstrated in several synthetic examples as well as the MNIST data set and some hyperspectral image data sets.  

\appendix
\renewcommand{\theequation}{\Alph{section}.\hindu{equation}}
\bhag{Constructing Localized Kernel $\Phi_n$}\label{bhag:construction}

With 
\be\label{projdef}
\mathsf{Proj}_m(\x,\y)=\sum_{|\k|_1=m}\psi_\k(\x)\psi_\k(\y),
\ee
we observe that
\be\label{summkerndef_bis}
\Phi_n(H;\x,\y)=\sum_{\k\in\ZZ_+^q}H\left(\frac{\sqrt{|\k|_1}}{n}\right)\psi_\k(\x)\psi_\k(\y)=\sum_{m=0}^\infty H\left(\frac{\sqrt{m}}{n}\right)\mathsf{Proj}_m(\x,\y).
\ee
In \cite{witnesspaper}, we have observed using the so-called Mehler identity that 
\be\label{reduceproj}
\mathsf{Proj}_m(\x,\y)=\sum_{j=0}^m \psi_j(|\x|)\psi_j(|\y|\cos\theta)\sum_{\ell=0}^{m-j}\psi_\ell(0)\psi_\ell(|\y|\sin\theta)D_{q-2;m-j-\ell},
\ee
where $\theta$ is the acute angle between $\x$ and $\y$, and
\be\label{reduceprojdetails1}
\psi_\ell(0)=\left\{\begin{array}{ll}
\disp \pi^{-1/4}(-1)^{\ell/2}\frac{\sqrt{\ell!}}{2^{\ell/2}(\ell/2)!},&\mbox{if $\ell$ is even},\\[1ex]
0, &\mbox{if $\ell$ is odd},
\end{array}\right.
\ee
and
\be\label{reduceprojdetails2}
D_{q-2;r}=\left\{\begin{array}{ll}
\disp \pi^{1-q/2}\frac{\Gamma(q/2+r/2-1)}{\Gamma(q/2-1) (r/2)!}, &\mbox{if $r$ is even, $q\ge 3$},\\[1ex]
0, &\mbox{if $r$ is odd, $q\ge 3$},\\[1ex]
1, &\mbox{if $q\le 2$}.
\end{array}\right.
\ee
Therefore, the procedure to compute the kernel $\Phi_n(H;\x,\y)$ is simple. 
We use the recurrence relations \eref{recurrence}
to compute the univariate Hermite functions $\psi_j$,   use these together with \eref{reduceproj} to compute $\mathsf{Proj}_m(\x,\y)$ for $|\m|_1\le n^2$, and finally compute $\Phi_n(H;\x,\y)$ using \eref{summkerndef_bis}.

\bibliographystyle{abbrv}
\bibliography{bibliography_active_clustering}
\end{document}